\pdfoutput=1

\documentclass[11pt,table,xcdraw]{article}
\usepackage[preprint]{acl}

\usepackage{times}
\usepackage{latexsym}

\usepackage[T1]{fontenc}

\usepackage[utf8]{inputenc}

\usepackage{microtype}

\usepackage{inconsolata}

\usepackage{graphicx}

\usepackage{arydshln}
\usepackage{amsmath}
\usepackage{colortbl}    
\usepackage{xcolor}      
\usepackage{booktabs}
\usepackage{enumitem}
\usepackage{float}
\usepackage{hhline}
\usepackage{makecell}
\usepackage{mathrsfs}
\usepackage{multirow}
\usepackage{paralist}
\usepackage{pifont}
\usepackage{tabu}
\usepackage{threeparttable}
\usepackage{subcaption}
\usepackage{scalerel,xparse}
\usepackage{balance}
\usepackage{longtable}
\usepackage{fontawesome5}
\usepackage{amssymb}

\usepackage{cancel}   
\usepackage{graphicx} 

\definecolor{gold}{rgb}{1.0, 0.84, 0.0} 
\definecolor{Gray}{gray}{0.5}
\definecolor{LGray}{gray}{0.9}
\definecolor{darkblue}{RGB}{94,110,186}
\definecolor{darkGreen}{RGB}{92, 148, 110}
\definecolor{myblue}{RGB}{14, 121, 178}
\definecolor{myred}{RGB}{192, 0, 0}
\definecolor{darkgreen}{HTML}{04bf29}
\definecolor{darkred}{HTML}{D1191F}
\definecolor{crimson}{RGB}{153, 0, 0}

\newcommand{\best}[1]{\colorbox{purple!30}{#1}}    
\newcommand{\secondbest}[1]{\colorbox{orange!25}{#1}} 
\newcommand{\thirdbest}[1]{\colorbox{gray!30}{#1}}  

\newcommand{\red}[1]{{\color{red}#1}}
\usepackage{tocloft}
\newcommand{\listcasestudyfiguresname}{\normalsize{List of Data Examples and Case Studies}}
\newlistof{casestudyfigures}{csf}{\listcasestudyfiguresname}

\newcommand{\casestudyfigure}[4]{%
  \clearpage
  \begin{figure*}[ht]
    \centering
    \refstepcounter{casestudyfigures}%
    \addcontentsline{csf}{casestudyfigures}{\protect\numberline{\thecasestudyfigures}#2}%
    \includegraphics[width=0.98\textwidth]{#1}
    \caption{#3}
    \label{#4}
    \hyperlink{listofcasestudyfigures}{Back to List of figures}
\end{figure*}}

\title{GODBench: A Benchmark for Multimodal Large Language Models in Video Comment Art}

\author{
    Yiming Lei$^{1,2}$\thanks{Work done during the internship at Kuaishou Technology.} \quad Chenkai Zhang$^{1,2*}$ \quad Zeming Liu$^{1\dagger}$ \quad Haitao Leng$^{3\spadesuit}$  \quad Shaoguo Liu$^{3}$\\
    \textbf{\quad Tingting Gao$^{3}$ \quad Qingjie Liu$^{1,2\dagger}$ \quad Yunhong Wang$^{1}$}\\
    $^1$Beihang University \quad $^2$ Hangzhou Innovation Institute, Beihang University \\
    \quad $^3$ Kuaishou Technology  \\
    {\tt\small $^*$Co-first authors:~\{ymlei, zhangchenkai\}@buaa.edu.cn }\\
    {\tt\small  $^\dagger$Corresponding authors $^\spadesuit$Project Leader}
}

\begin{document}
\maketitle
\begin{abstract}
\label{sec:abstract}
\textbf{\textit{Video Comment Art}} enhances user engagement by providing creative content that conveys humor, satire, or emotional resonance, requiring a nuanced and comprehensive grasp of cultural and contextual subtleties.
Although Multimodal Large Language Models (MLLMs) and Chain-of-Thought (CoT) have demonstrated strong reasoning abilities in STEM tasks (e.g. mathematics and coding), they still struggle to generate creative expressions such as resonant jokes and insightful satire. Moreover, existing benchmarks are constrained by their limited modalities and insufficient categories, hindering the exploration of comprehensive creativity in video-based \textit{Comment Art} creation.
To address these limitations, we introduce \textbf{GODBench}, a novel benchmark that integrates video and text modalities to systematically evaluate MLLMs' abilities to compose \textit{Comment Art}. Furthermore, inspired by the propagation patterns of waves in physics, we propose \textbf{Ripple of Thought (RoT)}, a multi-step reasoning framework designed to enhance the creativity of MLLMs.
Extensive experiments reveal that existing MLLMs and CoT methods still face significant challenges in understanding and generating creative video comments. In contrast, RoT provides an effective approach to improve creative composing, highlighting its potential to drive meaningful advancements in MLLM-based creativity. GODBench is publicly available at \href{https://github.com/stan-lei/GODBench-ACL2025}{our GitHub repository}.
\end{abstract}    
\section{Introduction}
\label{sec:intro}

\begin{figure}[t]
    \centering
    \vspace{0.2cm}
    \includegraphics[width=1.0\linewidth
    ]{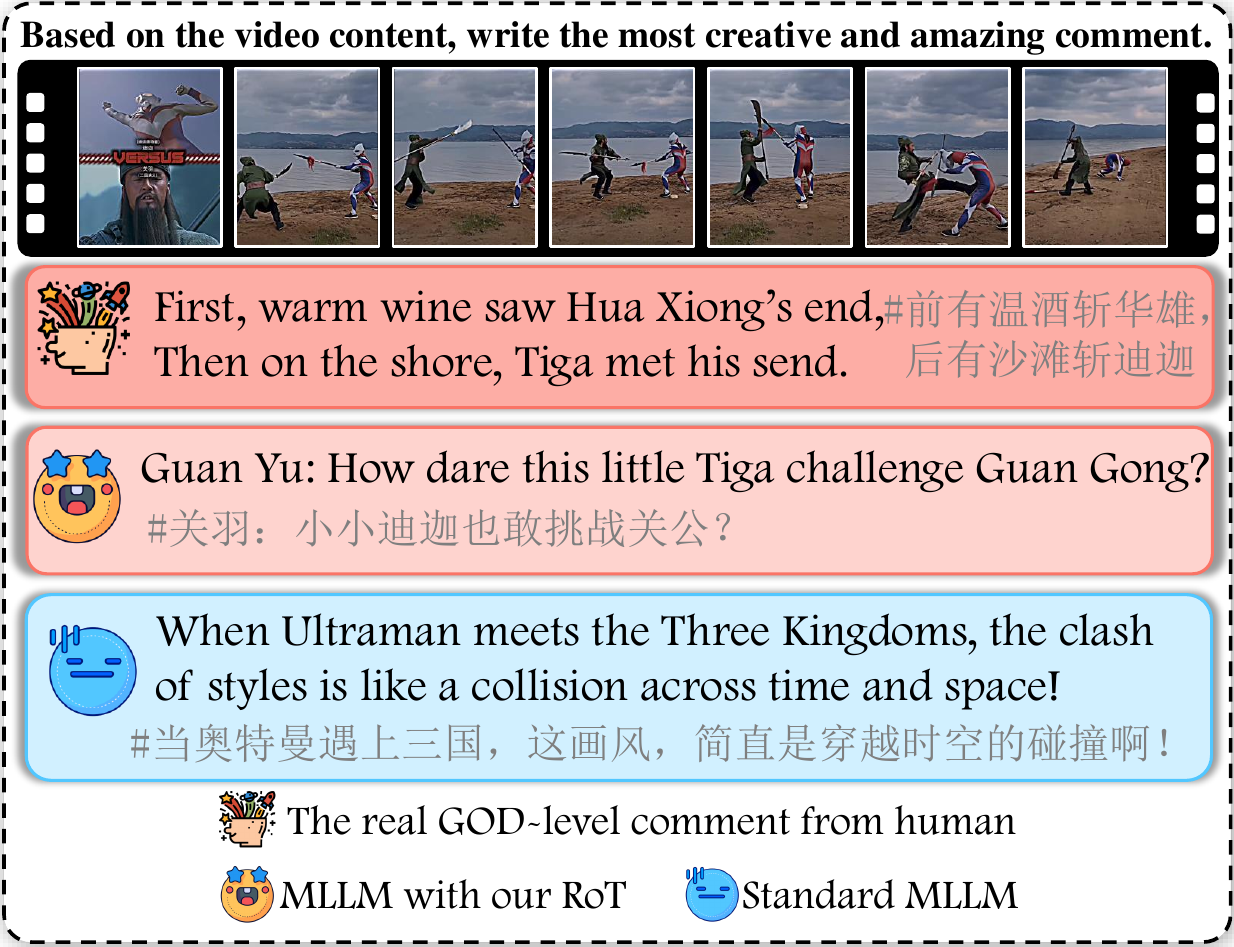}
    \vspace{-0.65cm}
    \caption{
    \textbf{Example from GODBench.}
    Showcasing a human-written GOD-level comment for the video, alongside the comments generated by model using the RoT framework and standard model. ``\#'' indicates that the original text is in Chinese.
    }
    \label{fig:intro}
    \vspace{-0.6cm}
\end{figure}

Recently, Multimodal Large Language Models (MLLMs)~\cite{Qwen2VL,videollama2_damonlpsg2024,zhang2024video,chen2024expanding} have achieved remarkable success in structured reasoning tasks, especially with the integration of the Chain of Thought (CoT) framework~\cite{kojima2022large,mitra2024compositional,wu2023role,yao2024tree,besta2024graph}.
However, despite their strong performance in logic-driven tasks, these models still struggle significantly in creative thinking~\cite{zhao2024assessing,tian2023macgyver,chen2023probing,nair2024creative}, remaining far from achieving human-level artistry. 
As illustrated in Fig. \ref{fig:intro}, MLLMs are constrained by rigid thinking, making it difficult to generate impressive and creative video comments akin to those of humans.

\textit{Video Comment Art}, the practice of crafting creative and insightful comments on videos, requires not only a deep understanding of video content and cultural context but also the ability to think divergently and imaginatively express ideas. Current MLLMs still struggle to generate human-like creative content due to their inferiority in creative thinking, which involves making diverse connections and uncovering deep insights.
To address this, several benchmarks~\cite{liu2024ii,xu2024good,zhong2024let,hessel-etal-2023-androids,sun2022expunations}  have been established to assess MLLMs' capabilities in \textit{Comment Art}. However, existing benchmarks are either restricted to a single category (e.g., humor, emotion, puns, or metaphors), confined to limited modalities (e.g., text-only or image-text pairs) or fail to account for the significance of comprehension in \textit{Comment Art}.

To bridge this gap, we introduce \textbf{GODBench}, a novel and comprehensive multimodal benchmark dataset specifically designed to evaluate and advance the abilities of MLLMs in understanding and creating \textit{Video Comment Art}. 
GODBench comprises over 67,000 high-quality videos paired with \textbf{``GOD-level comments''} that are characterized by their creativity and broad resonance. The videos in GODBench span 31 main categories and over 100 subcategories, ensuring rich diversity and comprehensiveness. These comments, created, voted on, and reviewed by real users, reflect human preferences, highly diverse thinking, and a deep understanding of video content, with quality endorsed by real users far surpassing that of previous benchmarks based on heuristic rules~\cite{chen2024hotvcom,sun2024vico}.
To facilitate comprehensive evaluation, we evaluate \textit{Video Comment Art} through five core dimensions: \texttt{[Rhetorical Techniques]}, \texttt{[Divergent Associations]}, \texttt{[Clever Writing Techniques]}, \texttt{[Interactive Virality]}, and \texttt{[Emotional Resonance]}. Furthermore, we define two primary tasks: discrimination and generation based on over 40,000 evaluation items. These tasks span diverse formats, including selection, ranking, classification, explanation, and creation, providing a systematical assessment of MLLMs' ability to understand and generate creative video comments.

Inspired by the similarity between the “Aha moment”~\cite{kounios2006prepared} and the physical phenomena of wave diffusion and interference~\cite{zakharov1968stability}, we propose \textbf{Ripple of Thought (RoT)}, a novel reasoning framework that enables MLLMs to think more expansively in the knowledge space.
Similar to the spread process of ripples, \textbf{RoT} leverages mechanisms such as diffusion and interference to enhance both the creativity and insightfulness of generated content. We conduct extensive experiments and analyses on 10 state-of-the-art MLLMs and observe that current MLLMs perform poorly in both discriminating and generating creative content, showing a notable performance gap compared to human capabilities.
However, applying the \textbf{RoT} framework significantly improves their performance across various tasks, even outperforming human-generated comments in human preference evaluations.
Despite these advancements, current MLLMs still fall short of human-level creativity, highlighting the immense potential for further research to enhance the creativity of MLLMs.
The main contributions of this paper are summarized as follows:
\\
\textbf{(1) Innovative Reasoning Methodology:} We propose the Ripple of Thought(RoT) reasoning framework to enhance the ability of MLLMs in generating \textit{Video Comment Art}. 
Extensive experiments on state-of-the-art baselines demonstrate RoT's potential to improve the creativity of MLLMs.
\\
\textbf{(2) New Dataset and Benchmark:} We construct GODBench, a large-scale and comprehensive dataset with diverse video-comment pairs, covering a wide range of tasks and significantly surpassing existing benchmarks in diversity and quality.
\\
\textbf{(3) Systematic and Comprehensive Evaluation:} 
We establish a structured framework to assess \textit{Video Comment Art} across five key creative dimensions and design a systematic set of evaluation tasks.

\section{GODBench}
We first compare GODBench with the previous benchmarks in Tab. \ref{tab:compare}. Then, we introduce the classification and annotation methods for the \textit{Comment Art}. Finally, we presente the design methodology for the evaluation tasks.

\subsection{Dataset Construction}
\begin{table*}[!t]
\centering

\label{T:main}
\setlength{\tabcolsep}{6pt}
\renewcommand{\arraystretch}{0.8}
\resizebox{\linewidth}{!}{
\begin{tabular}{lccccccccccccccc}
\toprule
\multirow{2}{*}{Benchmark} & \multicolumn{3}{c}{Videos} & \multicolumn{2}{c}{C-R Pairs} & \multicolumn{5}{c}{Dimensions of Comment Art} & \multicolumn{5}{c}{Task Types} \\
\cmidrule(lr){2-4} \cmidrule(lr){5-6} \cmidrule(lr){7-11} \cmidrule(lr){12-16}
 & Category  & Len(s) & Num. & Num. & Re. & RT & DA & WT & IV & ER & SEL & RNK & CLS & EXP & CRE \\

\toprule
TalkFunny~\citep{chen2024talk} & \textcolor{darkred}{\ding{55}}& \textcolor{darkred}{\ding{55}}& \textcolor{darkred}{\ding{55}} & 4k & \includegraphics[width=0.02\textwidth]{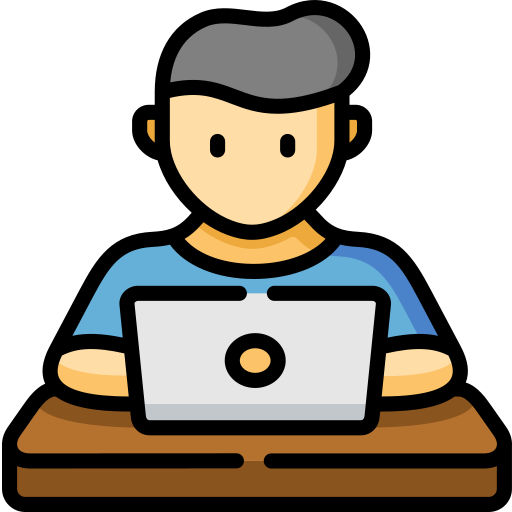} & \textcolor{orange}{\scalebox{1}{\bcancel{\checkmark}}}& \textcolor{darkgreen}{\ding{51}} & \textcolor{darkred}{\ding{55}} & \textcolor{darkred}{\ding{55}}  & \textcolor{darkred}{\ding{55}} & \textcolor{darkred}{\ding{55}} & \textcolor{darkred}{\ding{55}} & \textcolor{darkgreen}{\ding{51}} & \textcolor{darkred}{\ding{55}} &  \textcolor{darkgreen}{\ding{51}}\\

Chumor 2.0~\citep{he2024chumor} & \textcolor{darkred}{\ding{55}}& \textcolor{darkred}{\ding{55}}& \textcolor{darkred}{\ding{55}} & 3k & \includegraphics[width=0.02\textwidth]{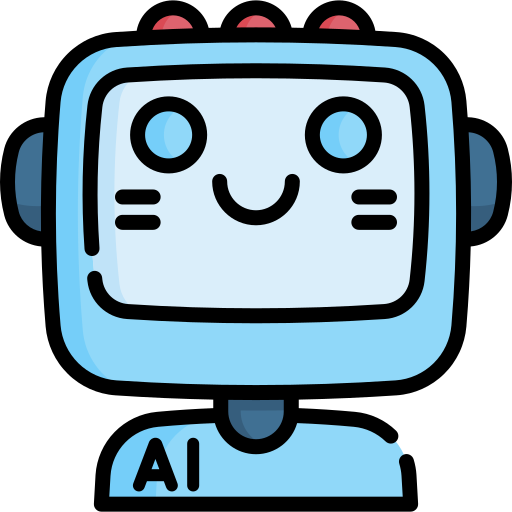} & \textcolor{orange}{\scalebox{1}{\bcancel{\checkmark}}}& \textcolor{darkred}{\ding{55}} & \textcolor{darkred}{\ding{55}} & \textcolor{darkgreen}{\ding{51}} & \textcolor{darkred}{\ding{55}} & \textcolor{darkred}{\ding{55}} & \textcolor{darkred}{\ding{55}} & \textcolor{darkred}{\ding{55}} & \textcolor{darkgreen}{\ding{51}} &  \textcolor{darkred}{\ding{55}}\\

Puns~\citep{xu2024good} & \textcolor{darkred}{\ding{55}}& \textcolor{darkred}{\ding{55}}& \textcolor{darkred}{\ding{55}} & 2k & \includegraphics[width=0.02\textwidth]{figure/human.png} & \textcolor{orange}{\scalebox{1}{\bcancel{\checkmark}}}& \textcolor{darkred}{\ding{55}} & \textcolor{darkred}{\ding{55}} & \textcolor{darkred}{\ding{55}} & \textcolor{darkred}{\ding{55}}& \textcolor{darkred}{\ding{55}} & \textcolor{darkred}{\ding{55}} & \textcolor{darkgreen}{\ding{51}} & \textcolor{darkgreen}{\ding{51}} &  \textcolor{darkgreen}{\ding{51}}\\

II-Bench~\citep{liu2024ii} & \textcolor{darkred}{\ding{55}}& \textcolor{darkred}{\ding{55}}& \textcolor{darkred}{\ding{55}} & 1k & \includegraphics[width=0.02\textwidth]{figure/human.png} & \textcolor{orange}{\scalebox{1}{\bcancel{\checkmark}}}& \textcolor{darkred}{\ding{55}} & \textcolor{darkred}{\ding{55}} & \textcolor{darkred}{\ding{55}} & \textcolor{darkgreen}{\ding{51}} & \textcolor{darkgreen}{\ding{51}} & \textcolor{darkred}{\ding{55}} & \textcolor{darkred}{\ding{55}} & \textcolor{darkred}{\ding{55}} &  \textcolor{darkred}{\ding{55}}\\

Oogiri-GO~\citep{zhong2024let} & \textcolor{darkred}{\ding{55}}& \textcolor{darkred}{\ding{55}}& \textcolor{darkred}{\ding{55}} & 130k & \includegraphics[width=0.02\textwidth]{figure/human.png} & \textcolor{orange}{\scalebox{1}{\bcancel{\checkmark}}}& \textcolor{darkgreen}{\ding{51}} & \textcolor{darkred}{\ding{55}} & \textcolor{darkred}{\ding{55}} & \textcolor{darkred}{\ding{55}} & \textcolor{darkgreen}{\ding{51}} & \textcolor{darkgreen}{\ding{51}} & \textcolor{darkred}{\ding{55}} & \textcolor{darkred}{\ding{55}} &  \textcolor{darkgreen}{\ding{51}}\\

NYT-Captions~\citep{hessel-etal-2023-androids} & \textcolor{darkred}{\ding{55}} & \textcolor{darkred}{\ding{55}}& \textcolor{darkred}{\ding{55}}& 3k & \includegraphics[width=0.02\textwidth]{figure/human.png} & \textcolor{orange}{\scalebox{1}{\bcancel{\checkmark}}} & \textcolor{darkgreen}{\ding{51}} & \textcolor{darkred}{\ding{55}} & \textcolor{darkred}{\ding{55}} & \textcolor{darkred}{\ding{55}} & \textcolor{darkgreen}{\ding{51}} & \textcolor{darkgreen}{\ding{51}} & \textcolor{darkred}{\ding{55}} & \textcolor{darkgreen}{\ding{51}} & \textcolor{darkred}{\ding{55}} \\

\midrule

ViCo~\citep{sun2024vico} & 15 & - & 20k  & 3M & \includegraphics[width=0.02\textwidth]{figure/robot.png}& \textcolor{darkred}{\ding{55}}& \textcolor{darkred}{\ding{55}}& \textcolor{darkred}{\ding{55}}  & \textcolor{darkred}{\ding{55}} & \textcolor{darkred}{\ding{55}} & \textcolor{darkred}{\ding{55}} & \textcolor{darkred}{\ding{55}} & \textcolor{darkred}{\ding{55}} & \textcolor{darkred}{\ding{55}} & \textcolor{darkgreen}{\ding{51}} \\

HOTVCOM~\cite{chen2024hotvcom} & 20 & 96.44 & 93k & 137M & \includegraphics[width=0.02\textwidth]{figure/robot.png} & \textcolor{orange}{\scalebox{1}{\bcancel{\checkmark}}} & \textcolor{darkred}{\ding{55}} & \textcolor{darkred}{\ding{55}} & \textcolor{darkgreen}{\ding{51}} & \textcolor{darkred}{\ding{55}} & \textcolor{darkred}{\ding{55}} & \textcolor{darkred}{\ding{55}} & \textcolor{darkred}{\ding{55}} & \textcolor{darkred}{\ding{55}} & \textcolor{darkgreen}{\ding{51}} \\

\midrule
\textbf{GODBench (Ours)} & 31 & 55.52 & 67k & 1M & \includegraphics[width=0.02\textwidth]{figure/human.png}& \textcolor{darkgreen}{\ding{51}}& \textcolor{darkgreen}{\ding{51}} & \textcolor{darkgreen}{\ding{51}} & \textcolor{darkgreen}{\ding{51}} & \textcolor{darkgreen}{\ding{51}} & \textcolor{darkgreen}{\ding{51}} & \textcolor{darkgreen}{\ding{51}} & \textcolor{darkgreen}{\ding{51}} & \textcolor{darkgreen}{\ding{51}} & \textcolor{darkgreen}{\ding{51}} \\
\bottomrule 

\end{tabular}}
\caption{
    \textbf{Comparison between GODBench and other existing benchmarks.}
    In the table, \textbf{Video Category} indicates the number of video Categories in the benchmark (e.g., pets, food, comedy, etc.), \textbf{Len(s)} represents the average video duration in seconds, and \textbf{Num.} represents the total number of videos. 
    \textbf{C-R Pairs} refers to context-response pairs, \textbf{Num.} represents the quantity, and \textbf{Re.} represents whether the quality review was conducted by humans \includegraphics[width=0.02\textwidth]{figure/human.png} or LLMs \includegraphics[width=0.02\textwidth]{figure/robot.png}.
    \textbf{Comment Art} is subdivided into five dimensions: \textbf{RT}, \textbf{DA}, \textbf{WT}, \textbf{IV}, and \textbf{ER} respectively represent \textbf{Rhetorical Techniques}, \textbf{Divergent Associations}, \textbf{Clever Writing Techniques}, \textbf{Interactive Virality}, and \textbf{Emotional Resonance}. ``\textcolor{darkred}{\ding{55}}'' indicates the absence of a dimension, while ``\textcolor{orange}{\scalebox{1}{\bcancel{\checkmark}}}'' indicates that only part of it is included. In the \textbf{Task Types}, \textbf{SEL}, \textbf{RNK}, \textbf{CLS}, \textbf{EXP}, and \textbf{CRE} respectively represent \textbf{Selection}, \textbf{Ranking}, \textbf{Classification}, \textbf{Explanation}, and \textbf{Creation}.
}
\vspace{-0.5cm}
\label{tab:compare}
\end{table*}

The video and comment form a strongly correlated context-response pair that not only conveys user feedback but also reflects the cognitive and emotional responses generated after watching the videos. Therefore, high-quality video comments can be considered a form of \textit{Comment Art}.

On the renowned video platform \textbf{\textit{Kuaishou}}\footnote{https://www.kuaishou.cn}, high-quality videos often feature a \textbf{GOD-level Comment}—a highly creative and engaging comment upvoted by millions of users, reviewed by moderators, and marked as “GOD-level” by the platform, reflecting exceptional insight and creativity.
To study \textit{Video Comment Art}, we crawled over 67,000 videos from \textit{Kuaishou}, ensuring that each video includes a \textbf{GOD-level Comment}, as well as other \textbf{High-Quality Comments} and \textbf{Ordinary Comments}. All comments are accompanied by like counts to showcase human preferences.

\subsection{Comment Art Definition and Annotation}

\begin{figure*}[t]
    \centering
    \includegraphics[width=1.0\linewidth
    ]{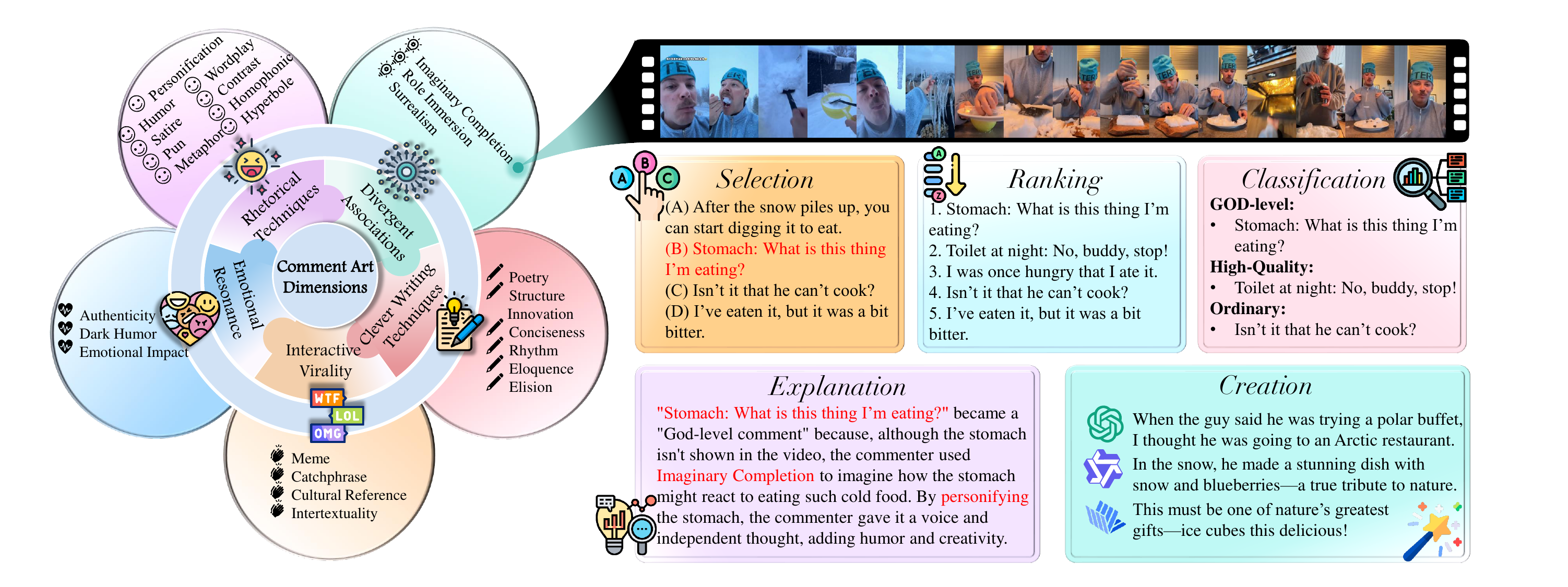}
    \vspace{-0.7cm}
    \caption{
    \textbf{The detailed definition of \textit{Comment Art} and example of various tasks.}
    \textit{Comment Art} is defined in five dimensions, each with different subcategories. One specific example of ``Imaginary Completion'' is presented, including the input video and various discriminative and generative tasks.
    }
    \label{fig:label}
    \vspace{-0.5cm}
\end{figure*}

In GODBench, the \textbf{GOD-level Comment}  serves as the concrete manifestation of \textit{Comment Art}. However, \textit{Comment Art} is a broad and abstract concept. Previous benchmarks have primarily focused on sub-concepts of \textit{Comment Art}, such as humor and metaphor, but lack a comprehensive analysis. To enable systematic study and proper evaluation metrics, we incorporated all related categories from previous work~\cite{chen2024talk,zhong2024let,hessel-etal-2023-androids,liu2024ii,chen2024hotvcom} and based on the characteristics of real-world data, partitioned \textit{Comment Art} into five dimensions: \textbf{Rhetorical Techniques}\cite{tianli2022examining,godioli2024humor,singsatit2022analysis,aras2024exploring}, \textbf{Divergent Associations}\cite{bellemare2024divergent,varshney2020explaining,beaty2023associative}, \textbf{Clever Writing Techniques}\cite{shalevska2024digital,hoult2020poetry}, \textbf{Interactive Virality}\cite{wang2020discursive,lee2024illusions,huntington2013subversive} and \textbf{Emotional Resonance}\cite{coburn2001subjectivity,heath2001emotional}.The specific results are shown in Fig. \ref{fig:label}(Left).

\textbf{1. Rhetorical Techniques:} This dimension encompasses rhetorical techniques such as humor, puns, and metaphors, with a particular focus on creatively utilizing language to provoke novelty or surprise in the audience's perception.

\textbf{2. Divergent Associations:} This dimension demands a deep understanding of the video and multi-step reasoning. For instance, in the ``Imaginary Completion'' category, comments may introduce entirely fictional entities that do not appear in the video but are closely tied to its content, thereby enriching the comment's creativity and originality.

\textbf{3. Clever Writing Techniques:} This dimension is often overlooked, as \textit{Comment Art} is expressed not only through content innovation but also through innovative writing structures and techniques. For instance, using poetic form, a comment can exhibit greater artistry and depth of expression.

\textbf{4. Interactive Virality:} 
This dimension demands the clever use of memes and cultural references in the appropriate context, requiring both background knowledge and expressive skills.

\textbf{5. Emotional Resonance:} This dimension requires comments expressing sincere feelings or having a strong emotional impact, capable of deeply touching the audience.

To ensure the accuracy of \textit{Comment Art} labels, we hired several professional annotators to perform manual annotation. The detailed annotation process can be found in Appendix \ref{appendix_data_annotation}.

\subsection{Evaluation Procedure and Metrics}
\label{sec:evaluation_procedure}
To comprehensively evaluate the video-based \textit{Comment Art} capabilities of MLLMs, we follow the evaluation methods of previous work\cite{hessel-etal-2023-androids,zhong2024let,xu2024good}, and designed two main categories of tasks: \textbf{Discriminative Task} and \textbf{Generative Task}. These tasks are crafted by leveraging the real-world data we have collected, ensuring high quality and reliability.
Specific examples can be found in Fig. \ref{fig:label}(Right).

\subsubsection{Discriminative Tasks}

In GODBench, each video is associated with three types of comments: \textbf{GOD-level Comments}, \textbf{High-Quality Comments}, and \textbf{Ordinary Comments}, each annotated with real human upvote counts. Based on this data, we can accurately design three types of discriminative tasks: \textbf{Selection}, \textbf{Ranking}, and \textbf{Classification} tasks.

\textbf{Selection:} A multiple-choice question can be created by selecting one GOD-level Comment as the correct answer and using High-Quality and Ordinary Comments as distractors. In this task, MLLMs must choose the most creative and amazing comment based on the video content.
    
\textbf{Ranking:} Since all comments are linked to real human upvote counts, reflecting human preferences, we use these counts to rank the comments. The MLLMs must correctly order multiple comments according to their quality, evaluating their understanding of human preferences.
    
\textbf{Classification Task:} Due to GODBench containing comments of three different quality levels, MLLMs are required to classify multiple comments into three quality categories based on their quality. 

\begin{figure*}[t]
    \centering
    \includegraphics[width=1.0\linewidth
    ]{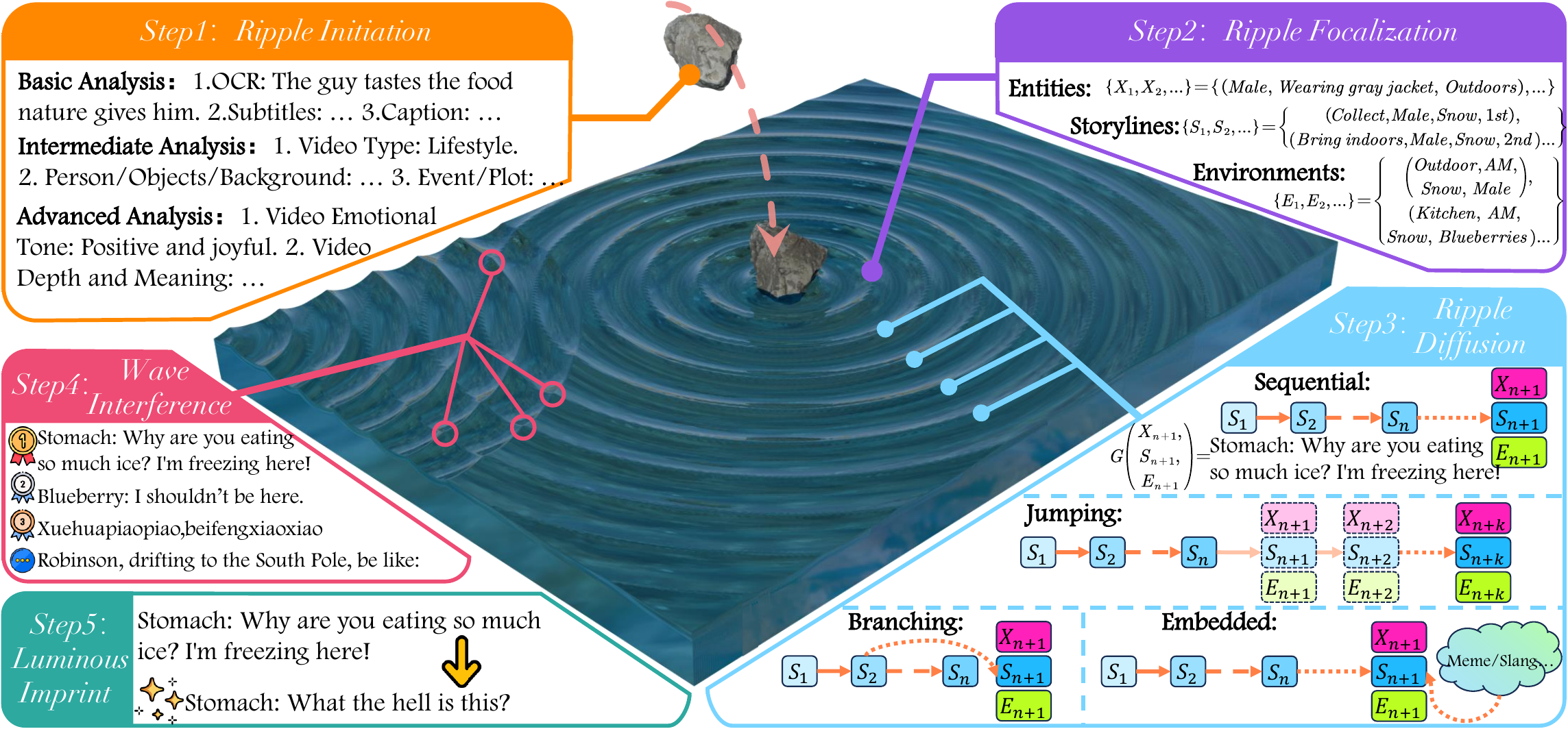}
    \vspace{-0.7cm}
    \caption{
    \textbf{Illustration of RoT.}
    Human creative thinking is like the diffusion of ripples, breaking down the propagation of waves in physics into five components, which are then transferred to the RoT reasoning framework of MLLMs. 
    }
    \label{fig:ROT}
    \vspace{-0.5cm}
\end{figure*}

\subsubsection{Generative Tasks}

Compared to discriminative tasks, generative tasks offer a more accurate measure of MLLMs' ability to produce \textit{Comment Art}. Therefore, we have categorized these tasks into two distinct types: \textbf{Explanation} and \textbf{Creation}.

\textbf{Explanation:} Each \textbf{GOD-level Comment} is manually annotated with \textit{Comment Art} dimensions. MLLMs must predict these dimensions and provide explanations, thereby evaluating their ability to analyze content and capture creative insights.
    
\textbf{Creation:} The ultimate challenge for MLLMs is generating a creative, contextually appropriate, and emotionally resonant comment based on the video content, serving as the most direct evaluation of the MLLMs' \textit{Comment Art} capabilities.

\section{Ripple of Thought}
To enhance the \textit{Comment Art} capabilities of MLLMs, inspired by the similarity between the “Aha moment”\cite{kounios2006prepared} and the physical phenomena of wave diffusion and interference\cite{zakharov1968stability}, we propose the Ripple of Thought (RoT) framework. This framework emulates the divergence and convergence of human thought, allowing RoT to expand reasoning in a ripple-like manner and enabling MLLMs to generate outputs that are both creative and meaningful.
This framework consists of five key phases: \textbf{Ripple Initiation}, \textbf{Ripple Focalization}, \textbf{Ripple Diffusion}, \textbf{Wave Interference}, and \textbf{Luminous Imprint}, with the specific structure shown in Fig. \ref{fig:ROT}. The detailed implementation can be found in Appendix \ref{implemen_of_method}.

\subsection{Ripple Initiation}
A stone cast into water creates the first ripple, symbolizing the initial spark that sets off all subsequent processes. Similarly, when a video or prompt is fed into MLLMs, it triggers the first cognitive oscillations. Thus, the initial step involves a thorough analysis of this ``stone'', allowing the MLLMs to fully understand and interpret the input video.

We employ a three-layer analysis method for the video. First, \textbf{Basic Analysis} applies OCR, subtitle extraction, and caption generation to capture video details. Next, \textbf{Intermediate Analysis} identifies video types, characters, objects, and event sequences, laying the groundwork for deeper reasoning. Finally, \textbf{Advanced Analysis} examines the video's emotional tone, cultural context, and social values to achieve a comprehensive understanding.

\subsection{Ripple Focalization}
When a stone strikes the water, the initial ripple—carrying the highest concentration of energy—sets the stage for all subsequent waves. Similarly, 
MLLMs must focus on the core entities, storylines, and environments extracted from the analyzed video information.
This foundational step shapes the entire reasoning process, much like the initial ripple determines the evolving pattern of the waves.
To achieve this, we use a formatted description formula to represent the entities \( X \), storylines \( S \), and environments \( E \) as a unified expression.

\subsection{Ripple Diffusion}
The model's reasoning process mirrors ripples on a water surface, with each wave of thought spreading outward to form broader connections. 
Entities \( X \), storylines \( S \), and environments \( E \) continuously spark new connections, gradually extending to new content.
This ripple-like diffusion unfolds along four distinct pathways, categorized as follows: 

\textbf{(1) Sequential Association}: Based on the extracted multi-entity set, the model infers the next most relevant event or entity by following the logical order of the storyline.

\textbf{(2) Jumping Association}: Expanding on sequential association, the model performs additional reasoning steps to discover seemingly unrelated yet inherently connected entities, leading to unexpected but insightful creative associations.

\textbf{(3) Branching Association}: Unlike sequential inference, branching association detaches specific extracted entities that may have been overlooked, recombining them into novel concepts.

\textbf{(4) Embedded Association}: 
To improve the coherence and cultural relevance of generated content, it is essential to first deduce the relevant cultural context and trending memes from the video, and then seamlessly incorporate them into the output.

\subsection{Wave Interference}
In this phase, the process mirrors the natural phenomenon of wave interference, where interacting ripples on a water surface partially cancel out while others reinforce, forming the strongest centers. Similarly, MLLMs must refine and prioritize the multitude of associative possibilities generated in earlier stages. 
This process performs an internal ranking of the multiple associative results, selecting the most creative, relevant, engaging, and resonant expressions, ensuring that the final output is both high-quality and thematically consistent.

\subsection{Luminous Imprint}
This phase requires MLLMs to refine and optimize the filtered content, enhancing clarity, coherence, and contextual relevance to generate the final comment. Just as ripples eventually stabilize into a distinctive luminous pattern on the water's surface, the output must retain the dynamic essence of thought while leaving a lasting impression.

\section{Experiment}
\begin{table*}[!ht]
    \centering
    \small
    \renewcommand\tabcolsep{7pt} 
    \renewcommand\arraystretch{1.0} 
    \resizebox{1.0\linewidth}{!}{
        \begin{tabular}{c|c|c|cccc|cc|cc}
            \toprule
            \rowcolor{gray!30} 
            \textbf{Model}
            & \textbf{Size}
            & \textbf{Frames}
            & $S^{[1,1,1]}_{\textnormal{acc}}$ & $S^{[1,1,1]}_{\textnormal{top-2}}$ & $S^{[1,3]}_{\textnormal{acc}}$ & $S^{[1,12]}_{\textnormal{acc}}$
            & $R^{[1,4]}_{\textnormal{NDCG}}$ & $R^{[1,4]}_{\textnormal{EMA}}$
            & $C^{[1,3,5]}_{\textnormal{acc}}$ & $C^{[1,3,5]}_{\textnormal{EMA}}$ \\

            \midrule
            \rowcolor{gray!10} 
            Random Choice & - & - & 33.40 & 66.58 & 24.65 & 7.88 & 63.18 & 0.87 & 46.22 & 0.37 \\
            \rowcolor{gray!10} 
            Frequent Guess & - & - & 33.76 & 66.70 & 26.02 & 8.41 & 62.75 & 1.11 & 59.25 & 0.00 \\
            \rowcolor{gray!10} 
            Human & - & - & 84.21 & 95.18 & 70.59 & 42.86 & 79.01 & 10.53 & 70.37 & 11.11 \\
        
            \midrule
            \rowcolor{gray!20} 
            \multicolumn{11}{c}{\textit{Open-source Video MLLMs}} \\
            \midrule
            LLaVA-Video & 7B & 64 
            & 36.11 & 77.24 & 23.78 & 10.04 & 49.17 & 0.41 & 38.46 & 0.13 \\

            mPLUG-Owl3 & 7B & 128
            & 34.75 & 72.18 & 24.39 & 9.77 & 62.73 & 0.63 & 36.68 & 0.11 \\

            MiniCPM-V 2.6 & 8B & 64 
            & 41.30 & 78.83 & 28.38 & 11.45 & 58.73 & 0.39 & 41.12 & 0.00 \\

            MiniCPM-o 2.6 & 8B & 64 
            & 40.60 & 78.12 & 27.61 & 11.42 & 53.16 & 0.66 & 41.39 & 0.04 \\

            \midrule
            \multirow{2}{*}{Qwen2-VL}
            & 2B & \multirow{2}{*}{dyn.} 
            & 37.45 & 73.39 & 25.73 & 8.09 & 48.19 & 0.27 & 28.64 & 0.00 \\
            & 7B &  
            & \secondbest{45.75} & \secondbest{84.06} & \thirdbest{30.24} & \thirdbest{13.33} & \secondbest{62.98} & \thirdbest{0.77} & 38.37 & 0.09 \\
            \midrule
            \multirow{2}{*}{InternVL2.5} 
            & 8B & \multirow{2}{*}{32} 
            & 44.27 & 81.54 & 28.61 & 13.19 & 45.43 & 0.52 & \thirdbest{43.41} & \thirdbest{0.20} \\
            & 20B & 
            & \thirdbest{45.59} & \thirdbest{82.07} & \secondbest{31.17} & 13.28 & 54.14 & \thirdbest{0.77} & \secondbest{43.99} & \secondbest{0.21} \\

            \midrule
            \rowcolor{gray!20} 
            \multicolumn{11}{c}{\textit{Commercial MLLMs}} \\
            \midrule

            GPT-4o-mini & \textasciitilde 8B & 50 
            & 44.91 & 81.87 & 28.99 & \secondbest{13.83} & \thirdbest{62.86} & \thirdbest{1.43} & 38.44 & 0.13 \\
            
            GPT-4o & \textasciitilde 200B & 50 
            & \best{54.19} & \best{88.32} & \best{37.86} & \best{18.84} & \best{65.21} & \best{1.52} & \best{53.16} & \best{0.68} \\

            \midrule
            \rowcolor{gray!20} 
            \multicolumn{11}{c}{\textit{MLLMs after Supervised Fine-Tuning}} \\
            \midrule
            \rowcolor{blue!10} 
            Qwen2-VL\red{{$^\dag$}}
            & 7B & dyn.
            & 66.24 & 90.26 & 50.32 & 30.41 & 71.45 & 3.07 & 64.37 & 8.13 \\
            
            \rowcolor{blue!10} 
            InternVL2.5\red{{$^\dag$}}
            & 7B & 32 
            & 70.27 & 91.02 & 54.63 & 33.57 & 74.81 & 4.96 & 69.53 & 10.12 \\
            
            \bottomrule
        \end{tabular}
    }
    \vspace{-7pt}
    \caption{
        \textbf{Performance of MLLMs on Discriminative Tasks.}
        Size means the LLM size. EMA is Exact Match Accuracy.
        Results are reported in percentage (\%).
        \red{{$\dag$}}: MLLMs fine-tuned with LoRA.
        The best, second-best, and third-best results are marked \colorbox{purple!30}{purple}, \colorbox{orange!25}{orange}, and \colorbox{gray!30}{gray}, respectively.
    }
    \vspace{-0.5cm} 
    \label{tab:discriminative_task}
\end{table*}


\subsection{Experimental Setups}
\textbf{Metrics.}
All tasks introduced in Sec.~\ref{sec:evaluation_procedure} follow a [1, m, n] configuration, where each video/image set includes \textbf{one} GOD-level comment, m high-quality comments, and n ordinary comments.
To comprehensively evaluate model performance, we employed three types of assessment:  
\begin{inparaenum}[\itshape 1)]
    \item \textit{Automatic Evaluation}: For discriminative tasks, we used accuracy and Normalized Discounted Cumulative Gain (NDCG)~\cite{pmlr-v30-Wang13} to measure model performance. For generative tasks, we employed automatic metrics including BLEU, DIST, ROUGE, and F1$_{\text{BERT}}$ similar to HOTVCOM~\cite{chen2024hotvcom}, to assess textual relevance and quality.  
    \item \textit{LLM-Based Judgement}: Following the procedure of \textit{LLM-as-a-judge}~\cite{huggingface_llm_judge}, we employed GPT-4o as a judging model, utilizing a graded scoring system that references human-annotated answers for multi-dimensional assessment.  
    \item \textit{Human Evaluation}: To further validate the quality of the generated content, evaluators ranked the comments based on multiple quality dimensions to ensure alignment with human preferences.
\end{inparaenum}  
Further details on the evaluation can be found in Appendix \ref{sec:human_evaluation} and \ref{sec:gpt4o_judgement}.
\\
\textbf{Model Selection.}
We conduct a comprehensive experiments on \textbf{GODBench} using both open-source and closed-source Video-MLLMs: (a) open-source models with different parameters: LLaVA-Video~\cite{zhang2024videoinstructiontuningsynthetic}, MiniCPM-V 2.6, MiniCPM-o 2.6~\cite{yao2024minicpm}, mPLUG-Owl3~\cite{ye2024mplugowl3}, InternVL2.5~\cite{chen2024expanding}, and Qwen2-VL~\cite{Qwen2VL}; (b) commercial MLLMs, such as GPT-4o~\cite{gpt4o_openai}. Further details on the models and evaluation settings are provided in the Appendix \ref{sec:implementation_details}.
\\
\textbf{Implementation and Inference.}
We constructed a fine-tuning dataset on discriminative tasks from the training set 
and used Llama-Factory~\cite{zheng2024llamafactory} to fine-tune Qwen2-VL and InternVL2.5(see Appendix \ref{sec:training_details}). 
For inference, we followed the official inference and frame extraction configurations similar to Video-MME~\cite{fu2024video}. \textbf{RoT} is applied to the following two LLMs: (1) Qwen2-VL, and (2) InternVL2.5. For the 5-shot tasks, we randomly select ten videos from the training set that share the same category as the target video. Then, we rank their corresponding GOD-level comments based on rules and retain the top five as context(see Appendix \ref{sec:inference_details}).

\subsection{Results and Analysis}
\label{sec:results}
To evaluate the ability of MLLMs in understanding and creating \textit{Video Comment Art}, we first propose several key research questions (RQs) and address them individually through quantitative experimental results, including (1) the ability to identify \textit{artistic comments} precisely, (2) the comprehension of deep conceptual aspects of \textit{Comment Art}, and (3) the capacity to compose \textit{Video Comment Art}.

\textbf{RQ1: Can MLLMs precisely distinguish \textit{artistic comments} from ordinary ones?}
As shown by their significantly higher $S^{[1,1,1]}_{\textnormal{top-2}}$ score compared to random baseline in Tab.~\ref{tab:discriminative_task}, MLLMs can effectively distinguish ordinary content from high-quality content. However, their performance remains limited when compared to humans, especially in tasks requiring fine-grained discrimination between high-quality comments and GOD-level comments. 
We attribute this limitation to the inherent rigidity of MLLMs to \textit{Comment Art Appreciation}.
In contrast, MLLMs fine-tuned with LoRA show notable improvement in distinguishing creative comments,  indicating that \textbf{GODBench} effectively enhances their conceptual understanding of \textit{Comment Art}.

\textbf{RQ2: How well do MLLMs understand deep conceptual aspects of \textit{Comment Art}? }
We evaluated the accuracy of MLLMs in the tag discrimination task and employed GPT-4o to assess tag explanation based on five criteria: \textit{Precision, Reasonableness, Completeness, Relevance, and Clarity}. The results in Tab.~\ref{tab:tag_discrimination_results} reveal several key findings:
\begin{inparaenum}[\itshape 1)]
    \item Most MLLMs struggle with accurately choosing tags for GOD-level comments, which is fundamental to understanding the deep conceptual aspects of \textit{Comment Art}.
    \item MLLMs exhibit limitations in understanding deep semantic and cultural contexts. While some MLLMs can correctly select tags for comments, their explanations for tagging often deviate significantly from human interpretation.
\end{inparaenum}
\begin{table}[ht!]
    \centering
    \small
    \renewcommand\tabcolsep{4pt} 
    \renewcommand\arraystretch{1.0} 
    \renewcommand\tabcolsep{4pt} 
    \resizebox{1.0\linewidth}{!}{
        \begin{tabular}{r|c|c|c|c|c|c|c}
            \toprule
            \multicolumn{1}{c|}{\multirow{2}{*}{\textbf{Model}}}
            & \multirow{2}{*}{\textbf{OA}}
            & \multicolumn{5}{c|}{\textbf{Tag Discrimination}} & \multirow{2}{*}{\textbf{S$_{\text{GPT-4o}}$}} \\
            \cmidrule{3-7}
            & & \textbf{RT} & \textbf{DA} & \textbf{WT} & \textbf{IV} & \textbf{ER} \\

            \midrule
            LLaVA-Video & 16.9 & 45.3 & 0.1 & 0.0 & 0.5 & 12.3 & 108.8 \\
            mPLUG-Owl3 & 21.0 & 54.8 & 1.1 & 1.1 & 0.7 & 14.2 & 159.8 \\
            MiniCPM-V 2.6 & 28.3 & 57.9 & 12.3 & 7.0 & 9.8 & 14.2 & 162.5 \\
            MiniCPM-o 2.6 & 26.2 & \secondbest{58.0} & 10.0 & 2.2 & 8.2 & 1.2 & 171.5 \\
            GPT-4o & 29.7 & 47.8 & 17.5 & \secondbest{17.1} & 11.3 & \best{58.1} & 214.3 \\

            \midrule
            Qwen2-VL$_{\text{2B}}$ & 19.9
            & 55.4 & 0.0 & 1.1 & 0.0 & 0.0 & 177.8 \\
            Qwen2-VL$_{\text{7B}}$ & 20.3 & 51.3 & 0.7 & 7.0 & 1.2 & 26.9 & 190.6 \\
            +CoT & 17.2 & 36.5 & 4.0 & 7.2 & 6.9 & 27.6 & 166.5 \\
            +CCoT & 19.2 & 49.3 & 0.1 & 7.2 & 1.2 & 24.0 & 189.9 \\
            +RoT(Ours) & \best{47.3} & \best{58.2} & \best{63.9} & 7.0 & 9.2 & 22.3 & \best{229.0}\\
            \midrule
            InternVL2.5$_{\text{26B}}$ & 19.8
            & 37.5 & 6.3 & 11.8 & 11.7 & 37.9 & 199.8 \\
            InternVL2.5$_{\text{7B}}$ & 22.2 & 35.0 & 12.9 & 13.5 & 18.3 & 33.2 & 184.8 \\
            +CoT & 28.2 & 57.5 & 7.0 & 15.0 & 13.3 & 55.7 & 197.8 \\
            +CCoT & 24.5 & 41.5 & 9.2 & 16.7 & \secondbest{26.8} & 54.6 & 208.0 \\
            +RoT(Ours) & \secondbest{46.8} & 40.8 & \secondbest{55.6} & \best{18.1} & \best{28.4} & \secondbest{57.9} & \secondbest{222.1} \\

            \bottomrule
        \end{tabular}
    }
    \vspace{-5pt} 
    \caption{\textbf{Performance of MLLMs on Tagging and Explanation Tasks.} Results include accuracy (\%) for tag discrimination and GPT-4o scores for tag explanation. OA represents the average value of the five tags.}
    \vspace{-7pt} 
    \label{tab:tag_discrimination_results}
\end{table}

\textbf{RQ3: Are MLLMs capable of composing \textit{Video Comment Art}? }
The results of automated metrics and LLM-based evaluations are presented in Tab.~\ref{tab:creation_results}, leading to the following key conclusions.
First, we find that RoT performs significantly better than most baselines and even outperforms larger MLLMs on all automated metrics.
Second, we evaluate Qwen2-VL and InternVL2.5 under different settings and observe that larger sizes and more examples lead to better performance. While CoT and CCoT provide some improvements, their performance fails to provide a distinct advantage over 5-shot settings.
Finally, we use GPT-4o to comprehensively evaluate the generated comments based on \textit{Creativity, Quality, Style, and Impact} (see Appendix \ref{sec:gpt4o_judgement} for details of criteria). RoT exhibits a notable increase in scores, enhancing the quality of generated creative comments. Interestingly, GPT-4o exhibits a consistent preference for its own generated content, even when there is no significant improvement in human preference.
\begin{table}[ht!]
    \centering
    \small
    \renewcommand\tabcolsep{8pt} 
    \renewcommand\arraystretch{0.80} 
    \renewcommand\tabcolsep{2pt} 
    \vspace{-5pt} 
    \resizebox{1.0\linewidth}{!}{
        \begin{tabular}{r|c|c|c|c|c|c}
            \toprule
            \multicolumn{1}{c|}{\textbf{Model}}
            & \textbf{BLEU-1} & \textbf{BLEU-2} & \textbf{DIST-1} & \textbf{ROUGE-L} 
            & \textbf{F1$_{\text{BERT}}$} 
            & \textbf{S$_{\text{GPT-4o}}$}\\

            \midrule
            LLaVA-Video & 0.23 & 0.05 & 0.37 & 4.41 
            & 52.78 & 322.82 \\
            mPLUG-Owl3 & 2.0 & 0.38 & 1.75 & 5.28 
            & 53.29 & 359.34 \\
            MiniCPM-V 2.6 & 4.41 & 1.02 & 6.24 & 6.68 
            & 55.32 & 298.67 \\
            MiniCPM-o 2.6 & 2.79 & 0.65 & 4.08 & 5.77 
            & 54.54 & 298.18 \\

            GPT-4o & 6.36 & 1.37 & \best{11.86} & 6.88
            & 56.21 & \best{425.99} \\
            \midrule
            Qwen2-VL$_{\text{2B}}$ & 4.72 & 1.14 & 4.99 & 6.84
            & 55.45 & 316.87 \\
            
            Qwen2-VL$_{\text{7B}}$ & 7.41 & 2.21 & 8.04 & 8.32
            & 57.31 & 332.65 \\
            5-shot & 8.73 & 3.02 & 8.46 & 10.41 
            & \secondbest{58.28} & 352.55 \\
            +CoT & 8.76 & 2.76 & 8.79 & 8.87 
            & 57.14 & 383.01 \\
            +CCoT & 9.19 & 3.04 & 8.85 & 9.07
            & 57.36 & 389.08 \\
            +RoT(Ours) & \best{9.51} & \secondbest{3.34} & \secondbest{9.02} & \best{11.33} 
            & \best{58.92} & \secondbest{389.68} \\

            \midrule
            InternVL2.5$_{\text{26B}}$ & 5.32 & 1.13 & 9.01 & 6.54 
            & 55.81 & 353.64 \\

            InternVL2.5$_{\text{8B}}$ & 3.40 & 0.62 & 5.95 & 5.01 
            & 53.98 & 342.23 \\
            5-shot & 5.63 & 1.36 & 8.85 & 7.49 
            & 56.19 & 347.45 \\
            +CoT & 6.07 & 1.41 & 6.86 & 7.42 
            & 55.71 & 379.85 \\
            +CCoT & 4.87 & 1.24 & 6.85 & 6.72 
            & 55.32 & 377.67 \\
            +RoT(Ours) & \secondbest{9.23} & \best{3.39} & 7.73 & \secondbest{10.44} 
            & 56.97 & 388.59 \\
            \bottomrule
        \end{tabular}
    }
    \vspace{-5pt} 
    \caption{
        \textbf{Performance of MLLMs on comment creation results.}
        Results are reported in percentage(\%).
        S$_{\text{GPT-4o}}$ denotes the quality score judged by GPT-4o.
    }
    \vspace{-5pt} 
    \label{tab:creation_results}
\end{table}

\begin{figure}[!ht]
    \centering
    \vspace{-8pt} 
    \includegraphics[width=1.0\linewidth
    ]{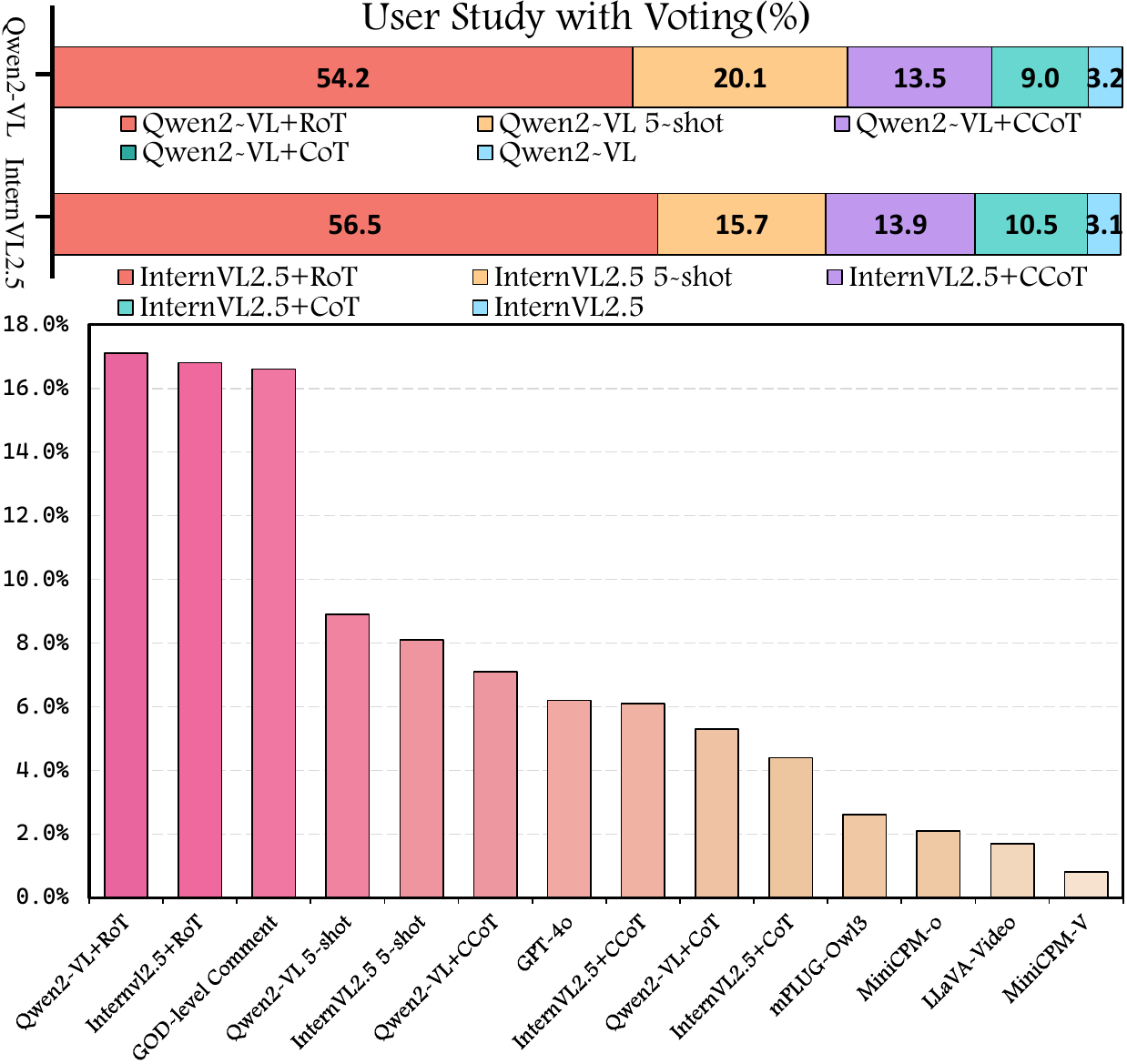}
    \vspace{-0.75cm}
    \caption{
    \textbf{User study with voting(\%) by different models and improved methods.}
    }
    \label{fig:human}
    \vspace{-5pt} 
\end{figure}

We also conduct a human preference study to evaluate the creativity of generated comments, asking users to select the most creative ones. As shown in Fig.~\ref{fig:human}, users show a strong preference for comments generated by RoT, which are comparable to, or even exceeding GOD-level comments from humans, highlighting RoT's ability to produce high-quality creative comments. Further details on the human preference study are in Appendix \ref{sec:human_evaluation}.

\subsection{More Analysis}
To further emphasize the exceptional creativity of \textbf{RoT}, we designed divergent association tasks that focus on entity innovation and developed a Weighted Entity Overlap(WEO) metric to assess the creativity of MLLMs. Specifically, we utilized GPT-4o to extract entities from both GOD-level comments and model-generated comments, denoted as \( E_{\text{gen}} \) and \( E_{\text{ref}} \), respectively. Next, we assigned a weight \( w_e \) to each entity \( e \) based on its frequency. Finally, we computed the score of WEO, formulated as:
\begin{equation}
    \text{WEO} = \frac{1 + \sum_{e \in E_{\text{gen}} \cap E_{\text{ref}}} w_e}{\sum_{e \in E_{\text{gen}} \cup E_{\text{ref}}} w_e}
    \label{eq:entity_innovation_weighted}
\end{equation}

\begin{figure}[!ht]
    \centering
    \vspace{-3pt} 
    \includegraphics[width=1.0\linewidth
    ]{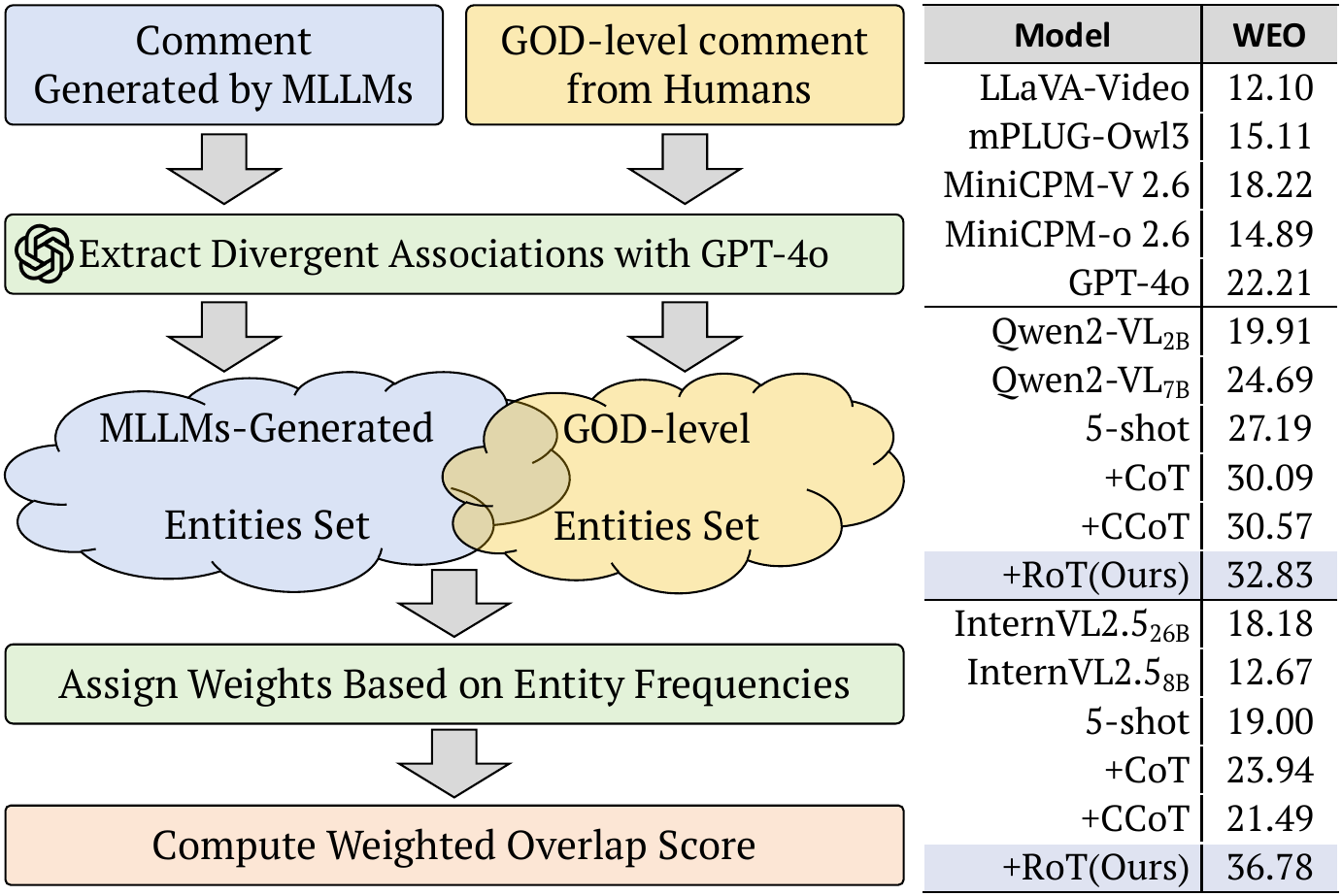}
    \vspace{-0.5cm}
    \caption{
    \textbf{Procedure and results of WEO score on divergent association tasks in GODBench.}
    }
    \label{fig:entity}
    \vspace{-0.2cm}
\end{figure}

Fig.~\ref{fig:entity} illustrates the stages and results of the WEO score. Due to its outstanding divergent association ability, RoT exhibits greater overlap with GOD-level comments, achieving superior performance and offering a novel perspective for advancing generative model capabilities.
\section{Related Work}
\textbf{Multimodal LLMs for Creative Arts.}
Recent advances in Multimodal Large Language Models (MLLMs)~\cite{Qwen2VL,chen2024expanding,gpt4o_openai} have significantly enhanced their logical reasoning capabilities, driven by techniques such as Chain-of-Thought (CoT) prompting~\cite{mitra2024compositional} and reasoning~\cite{o1_openai,zhao2024marcoo1}. However, their application to creative artistic tasks mainly focuses on either literary creation~\cite{chen-etal-2024-hollmwood,artartificelargelanguage} or surface-level humor~\cite{zhong2024let} and puns~\cite{xu2024good}, remaining confined to a restricted subset of creative thinking within \textit{Comment Art} while lacking attention to video-based multimodal creativity. Therefore, we propose \textbf{Ripple of Thought (RoT)}, a novel reasoning framework enabling MLLMs to compose more creative, imaginative, and engaging video comments.
\\
\textbf{Evaluation of Creativity in LLMs.}
\textit{Video Comment Art} is the practice of crafting insightful, humorous, and culturally resonant comments to enhance engagement and enrich the viewing experience. 
Prior research on creative thinking in MLLMs is fragmented and lacks a comprehensive evaluation framework for Comment Art: (1) Understanding and Generation, which focuses on humor~\cite{zhong2024let,he2024chumor}, puns~\cite{sun2022expunations,xu2024good}, buzzworthy comments~\cite{chen2024hotvcom}, and metaphors~\cite{chen-ding-2023-probing,xie2024funqa,liu2024ii} but suffers from coarse category definitions, incomplete task coverage, and limited modality support, restricting real-world applicability; and (2) Human-Model Interactions, which explore creativity from a sociological perspective~\cite{franceschelli2024creativity,kumar2024human} but often overlook model-specific improvements, failing to enhance MLLMs' intrinsic creative reasoning abilities.
To address these limitations, we present the first systematic evaluation of MLLMs for \textit{Video Comment Art} and introduce a large-scale multimodal dataset comprising videos, images, and text with extensive and diverse human annotations.

\section{Conclusion}
\label{sec:conclusion}

To explore the capabilities of current MLLMs in \textit{Video Comment Art}, this paper introduces \textbf{GODBench}, a novel benchmark designed to assess MLLMs' ability to understand and generate creative video comments. We further propose \textbf{Ripple of Thought (RoT)}, an adaptable and robust framework that enhances models' creative and divergent thinking, leading to significant performance improvements—even surpassing human-generated content in user performance scenarios. Extensive experiments reveal that current MLLMs still struggle with understanding and generating creative comments, highlighting the need for continued progress in this area. We hope that \textbf{GODBench} and \textbf{RoT} will inspire further research focused on the creative capabilities of MLLMs.

\section*{Limitations}
We introduce GODBench, a novel and comprehensive benchmark designed to assess the ability of MLLMs to understand and generate \textit{Video Comment Art}. Due to the large number of diverse and challenging tasks included in GODBench, running a full evaluation requires significant computational resources. Additionally, since all tasks involve the video modality, there are also high demands on the context length for MLLMs.

\section*{Acknowledgements}
This work is supported by “Pioneer” and “Leading Goose” R\&D Program of Zhejiang (No. 2024C01020), the National Natural Science Foundation of China (No. 62406015), research funding from Kuaishou Technology, the Emergency Management Research and Development Project of Zhejiang Province (No. 2024YJ018).

\section*{Ethic Statement}
\textbf{Data Privacy} Throughout the course of our research, we have adhered to the highest ethical standards, ensuring that every aspect of our study complies with principles of transparency, fairness, and user privacy protection. The data used in our benchmark has undergone meticulous anonymization to safeguard user identities and protect personal information. All data processing is carried out in strict accordance with data protection and privacy regulations to minimize any risks to users.
\\
\textbf{Professional Annotation}
To ensure the quality and accuracy of data annotation, we employed professional annotators who possess a deep understanding of innovative content. These annotators are highly skilled in the task of marking and interpreting creative content. We have provided fair and equitable compensation for their work, ensuring that their efforts are appropriately rewarded while maintaining high standards of professionalism and responsibility in the annotation process.
\\
\textbf{AI-Generated Content Monitoring}
In the context of the potential risks associated with AI-generated comments, we remain highly vigilant and implement strict monitoring procedures. We carefully review all generated comments to identify and remove any content that could be harmful or inappropriate. This proactive approach ensures that the comments produced by our system adhere to ethical norms and do not have a negative impact on users or society.




\bibliography{custom}

\appendix
\clearpage
\section*{Appendix}
\label{sec:appendix}

\section{More Details of GODBench}
\subsection{Comment Art Dimension Definition}
Existing MLLMs have achieved human expert-level performance in logical reasoning and STEM tasks, but they still fall short of human capabilities in certain creative tasks. Previous benchmarks used to assess the \textit{Comment Art} of MLLMs either failed to provide a detailed classification or focused only on specific subcategories, such as humor, metaphor, and double entendre. Therefore, based on previous work~\cite{chen2024talk,zhong2024let,hessel-etal-2023-androids,liu2024ii,chen2024hotvcom} and the characteristics of real-world data, we partitioned \textit{Comment Art} into five dimensions: \textbf{Rhetorical Techniques}\cite{tianli2022examining,godioli2024humor,singsatit2022analysis,aras2024exploring}, \textbf{Divergent Associations}\cite{bellemare2024divergent,varshney2020explaining,beaty2023associative}, \textbf{Clever Writing Techniques}\cite{shalevska2024digital,hoult2020poetry}, \textbf{Interactive Virality}\cite{wang2020discursive,lee2024illusions,huntington2013subversive} and \textbf{Emotional Resonance}\cite{coburn2001subjectivity,heath2001emotional}. Real examples of all categories can be found in Appendix \ref{Case_study}, accompanied by video frames, GOD-level comment, and explanations.

\textbf{1. Rhetorical Techniques.} This category focuses on the use of language to enhance communication through stylistic elements and techniques that engage the audience. \textbf{1.1 Humor}: Humor involves the use of wit, jokes, or playful language to provoke laughter or amusement. \textbf{1.2 Satire}: Satire criticizes or exposes flaws in society, politics, or human behavior. It often aims to provoke thought and bring attention to important issues. \textbf{1.3 Homophonic}: Homophonic refers to wordplay based on the similarity in sound between two words, creating humor or ambiguity through phonetic resemblance. \textbf{1.4 Metaphor}: Metaphor involves describing one thing by referencing another, often to draw a comparison or convey a deeper meaning. It helps to create imagery and express complex ideas. \textbf{1.5 Double Entendre}: Double Entendre involves a phrase or expression with two interpretations—one innocent and the other suggestive or ironic—leading to humorous or playful ambiguity. \textbf{1.6 Hyperbole}: Hyperbole uses exaggerated statements that are not meant to be taken literally but are intended to emphasize a point or create a dramatic effect. \textbf{1.7 Wordplay}: Wordplay encompasses clever and witty uses of words, often relying on puns, double meanings, or creative manipulation of language to entertain and engage. \textbf{1.8 Contrast}: Contrast highlights the differences between two elements, often emphasizing their opposites to create a more vivid or impactful comparison. \textbf{1.9 Personification}: Personification assigns human traits, characteristics, or emotions to non-human entities, allowing them to appear more relatable or vivid.

\begin{figure*}[t]
    \centering
    \includegraphics[width=1.0\linewidth
    ]{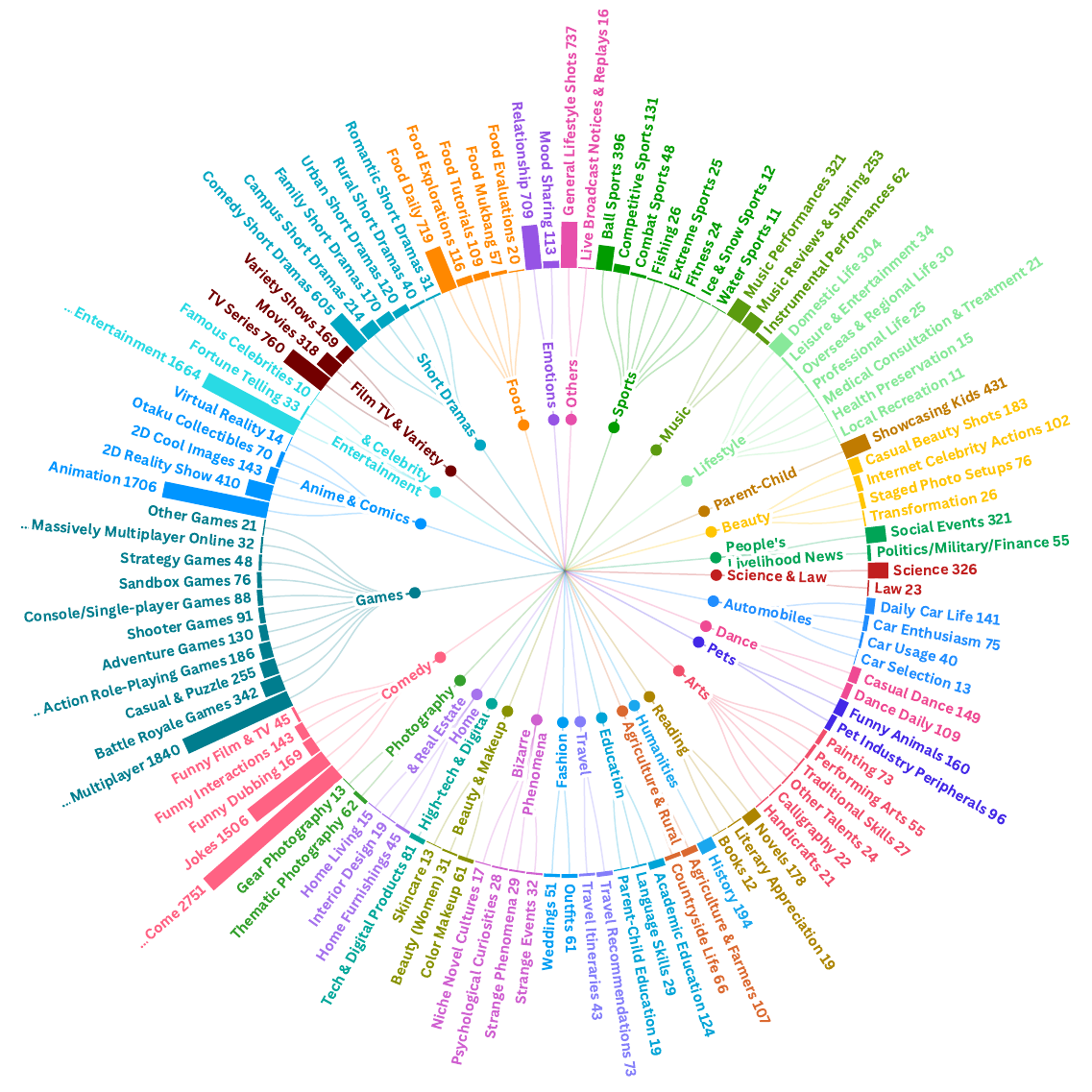}
    \vspace{-0.7cm}
    \caption{
    \textbf{The 31 categories of videos and their corresponding subcategories.}
     Each category contains multiple subcategories, and the number after each subcategory represents the corresponding number of videos.
    }
    \label{fig:video}
    \vspace{-0.5cm}
\end{figure*}

\textbf{2. Divergent Associations.} Divergent Associations involve the creative linking of seemingly unrelated ideas or concepts, leading to unexpected or imaginative connections. \textbf{2.1 Imaginary Completion}: This refers to the ability to use associative thinking to create entities, characters, or concepts that are entirely absent from the video, expanding the narrative in imaginative ways. \textbf{2.2 Role Immersion}: Role Immersion refers to the process of stepping into a different character or perspective, adopting new roles, and exploring ideas from an alternative viewpoint. \textbf{2.3 Surrealism}: Surrealism embraces irrationality and dream-like imagery, creating a departure from reality and allowing the exploration of imaginative or fantastical elements that challenge conventional thinking.

\textbf{3. Clever Writing Techniques.} Clever Writing Techniques emphasize the sophisticated use of language and structure to convey ideas in a creative and engaging manner. \textbf{3.1 Poetry}: Poetry refers to the use of traditional poetic structures, such as those found in classical Chinese poetry, to craft comments. It involves employing rhythm, meter, and figurative language to evoke emotions and create vivid imagery, often relying on concise, impactful expressions to convey deeper meanings and ideas. \textbf{3.2 Structure Innovation}: Innovation in writing refers to the introduction of new ideas, structures, or methods, breaking away from traditional forms to create something original or unexpected. \textbf{3.3 Conciseness}: Conciseness focuses on expressing ideas with precision and brevity, often using fewer words to convey a more powerful message or insight. \textbf{3.4 Rhythm}: Rhythm involves the pattern of sounds in writing, particularly through rhyme or cadence, which contributes to the flow and aesthetic appeal of the language. \textbf{3.5 Eloquence}: Eloquence refers to the graceful and persuasive use of language, characterized by fluency, clarity, and sophistication in expression. \textbf{3.6 Elision}: Elision involves the deliberate omission of words or phrases for effect, creating space for interpretation, drawing attention to what's left unsaid, or enhancing brevity.

\textbf{4. Interactive Virality.} Interactive Virality focuses on content's ability to engage audiences and spread widely across social platforms through participation and cultural relevance. \textbf{4.1 Meme}: Memes are viral pieces of content that spread rapidly online, often characterized by humor, relatability, or cultural relevance, and are shared extensively in social media and internet culture. \textbf{4.2 Catchphrase}: Catchphrases are short, memorable expressions that resonate with a broad audience and become widely repeated, often reflecting contemporary trends or ideas. \textbf{4.3 Cultural Reference}: Cultural References draw upon shared knowledge of cultural events, figures, or symbols, resonating with specific groups and enriching the content by invoking collective meaning. \textbf{4.4 Intertextuality}: Intertextuality refers to the practice of referencing or drawing upon other texts, media, or cultural works, creating layers of meaning and a deeper connection between different works.

\textbf{5. Emotional Resonance.} Emotional Resonance is about the ability of content to evoke strong emotions and connect with the audience on a personal level. \textbf{5.1 Authenticity}: Authenticity in content is about conveying genuine emotions or experiences in a way that resonates deeply with the audience, establishing a sense of trust and emotional connection. \textbf{5.2 Emotional Impact}: Emotional Impact refers to the intensity of the emotional response generated by the content, whether through joy, sadness, anger, or other feelings, leaving a lasting impression on the audience. \textbf{5.3 Dark Humor}: Dark Humor explores morbid, taboo, or grim subjects in a humorous light, often blending humor with serious or uncomfortable themes to create a unique emotional response.

\subsection{Data description}
To ensure high-quality data, we collected a large amount of video and comment data from the popular video platform, \textit{Kuaishou}\footnote{https://www.kuaishou.cn}. First, to guarantee the quality of the videos, we selected only those with likes and comments exceeding certain thresholds. Specifically, only videos with more than 10,000 likes and over 2,000 comments were chosen, while also striving to maintain a diverse range of video categories. These videos cover 31 popular categories and are further divided into more than 100 subcategories. For specific video categories, please refer to Fig.\ref{fig:video}.

Next, we filtered out videos containing ``GOD-level comments.'' ``GOD-level comments'' are a unique comment label on \textit{Kuaishou}, awarded to comments that receive high numbers of likes from millions of users after watching a video. Comments with the highest likes are then reviewed by professional platform moderators. Once approved, these comments are labeled as ``GOD-level comments'' and displayed on the platform. As a result, ``GOD-level comments'' are of exceptionally high quality and relatively rare, with each video having no more than one or two ``GOD-level comments.'' In addition, we also extracted other High-Quality Comments and Ordinary Comments for comparison learning, aiming to enhance the value of the data. High-Quality Comments refer to comments, aside from GOD-level comments, that have a relatively high number of likes or comments that experienced a significant increase in likes over a certain period. These comments are considered to reflect a certain level of creativity or insight, making them valuable for comparison. On the other hand, Ordinary Comments are those with a relatively low number of likes, which do not stand out in terms of engagement or impact but are still useful for contrast and analysis to improve model performance. For detailed information about GODBench, please refer to Tab.\ref{tab:dataset_statistics}.

\begin{table*}[h!]
    \centering
    \renewcommand{\arraystretch}{0.8}  
    \begin{minipage}[t]{0.48\textwidth}
        \centering
        \begin{tabular}{@{}ll@{}}
            \toprule
            
            \rowcolor[HTML]{D3D3D3} 
            \multicolumn{2}{l}{\textbf{Video}} \\ \midrule
            Total Videos                & 67,073        \\
            Train & 55,894 \\
            Validation & 5,589 \\
            Test & 5,589 \\
            Train : Validation : Test    & 10 : 1 : 1 \\
            Categories                  & 31 \\
            Average Duration (s)        & 55.52 \\
            Average Title Length        & 40.94 \\
            Average OCR Length          & 745.93 \\
            Average Subtitle Length     & 225.95 \\ \midrule
            \rowcolor[HTML]{D3D3D3} 
            \multicolumn{2}{l}{\textbf{Comment}} \\ \midrule
            Total Comments              & 1,577,201 \\
            Average Comments per Video  & 23.51 \\
            Total GOD-level Comments                   & 80,357 \\
            - Average per Video       & 1.19 \\
            - Average Likes      & 49,882.38 \\
            - Average Length      & 21.01 \\
            Total High-Quality Comments            & 826,124 \\
            - Average Video         & 12.3 \\
            - Average Likes          & 1,245.53 \\
            - Average Length            & 29.71 \\
            Total Ordinary Comments            & 670,720 \\
            - Average per Video         & 10.0 \\
            - Average Likes          & 6.45 \\
            - Average Length            & 16.33 \\ \midrule
            \rowcolor[HTML]{D3D3D3} 
            \multicolumn{2}{l}{\textbf{Task}} \\ \midrule
            Total Questions             & 40970 \\
            Selection   & 16,512 \\
            Ranking           & 5,504 \\
            Classification    & 5,504 \\
            Explanation       & 6,725 \\
            Creation        & 6,725
            \\ \midrule
            \rowcolor[HTML]{D3D3D3} 
            \multicolumn{2}{l}{\textbf{Comment Art Dimensions}} \\ \midrule
            Rhetorical Techniques & 2206 (32.90\%) \\
            - Humor & 669 (9.98\%) \\
            - Satire & 47 (0.70\%) \\
            - Homophonic & 39 (0.58\%) \\
            - Metaphor & 36 (0.54\%) \\
            - Double Entendre & 26 (0.39\%) \\
            - Hyperbole & 165 (2.46\%) \\
            - Wordplay & 12 (0.18\%) \\
            - Contrast & 319 (4.76\%) \\
            - Personification & 893 (13.32\%) \\
            \midrule
            Divergent Associations & 2895 (43.17\%) \\
            - Imaginary Completion & 369 (5.50\%) \\
            - Role Immersion & 2486 (37.07\%) \\
            - Surrealism & 40 (0.60\%) \\
                       
            \bottomrule
        \end{tabular}
    \end{minipage}
    \hspace{1pt}
    \begin{minipage}[t]{0.48\textwidth}
        \centering
        \begin{tabular}{@{}ll@{}}
            \toprule
            
            Clever Writing Techniques & 607 (9.05\%) \\
            - Poetry & 105 (1.57\%) \\
            - Innovation & 50 (0.75\%) \\
            - Conciseness & 82 (1.22\%) \\
            - Rhythm & 70 (1.04\%) \\
            - Eloquence & 221 (3.30\%) \\
            - Elision & 79 (1.18\%) \\
            \midrule
            Interactive Virality               & 513 (7.65\%) \\
            - Meme & 287 (4.28\%) \\
            - Catchphrase & 137 (2.04\%) \\
            - Cultural Reference & 24 (0.36\%) \\
            - Intertextuality & 65 (0.97\%) \\
            \midrule
            Emotional Resonance             & 485 (7.23\%) \\ 
           - Authenticity & 196 (2.92\%) \\
            - Emotional Impact & 221 (3.30\%) \\
            - Dark Humor & 68 (1.01\%)

            \\ \midrule
            \rowcolor[HTML]{D3D3D3} 
            \multicolumn{2}{l}{\textbf{Video Categories}} \\ \midrule
            Comedy & 12885 (19.21\%) \\
            Games & 8682 (12.95\%) \\
            Anime \& Comics & 6543 (9.76\%) \\
            Pets & 4906 (7.32\%) \\
            Celebrity \& Entertainment & 4767 (7.11\%) \\
            Film TV \& Variety & 3482 (5.19\%) \\
            Short Dramas & 3295 (4.91\%) \\
            Food & 2851 (4.25\%) \\
            Emotions & 2295 (3.42\%) \\
            Others & 2103 (3.14\%) \\
            Sports & 1879 (2.80\%) \\
            Music & 1776 (2.65\%) \\
            Lifestyle & 1229 (1.83\%) \\
            Parent-Child & 1204 (1.79\%) \\
            Beauty & 1081 (1.61\%) \\
            People's Livelihood News & 1050 (1.57\%) \\
            Science \& Law & 975 (1.45\%) \\
            Automobiles & 751 (1.12\%) \\
            Dance & 720 (1.07\%) \\
            Arts & 620 (0.92\%) \\
            Reading & 584 (0.87\%) \\
            Humanities & 542 (0.81\%) \\
            Agriculture \& Rural & 483 (0.72\%) \\
            Education & 480 (0.72\%) \\
            Travel & 324 (0.48\%) \\
            Fashion & 313 (0.47\%) \\
            Bizarre Phenomena & 296 (0.44\%) \\
            Beauty \& Makeup & 293 (0.44\%) \\
            High-tech \& Digital & 226 (0.34\%) \\
            Real Estate \& Home & 221 (0.33\%) \\
            Photography & 209 (0.31\%) \\

            \bottomrule
        \end{tabular}
    \end{minipage}
    \caption{Statistics of GODBench.}
    \label{tab:dataset_statistics}
    \vspace{-0.5cm}
\end{table*}

\section{More Details of Data Annotation}
\label{appendix_data_annotation}
\begin{figure*}[t]
    \centering
    \fbox{\includegraphics[width=0.9\textwidth]{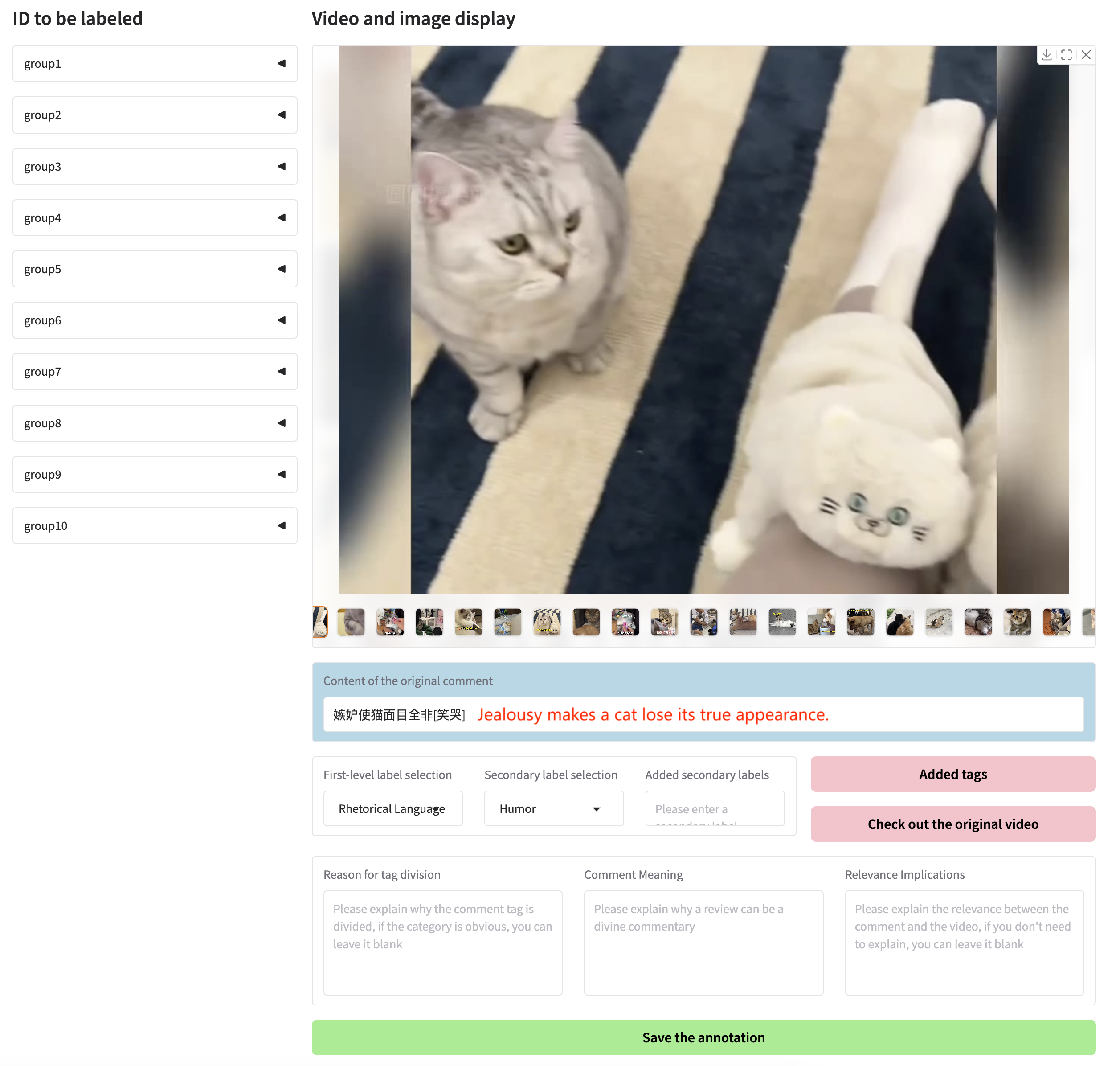}}
    \caption{
    \textbf{The interface for manual labeling.}
     It includes the selection of video groups to be labeled, the display area for the labeled videos, the corresponding comments, and the content to be labeled: the category tags of the comments and their corresponding explanations.
    }
    \label{fig:annotation_interface}
    \vspace{-0.5cm}
\end{figure*}

To ensure the accuracy of comment labels and establish a solid foundation for subsequent evaluation tasks, we engaged several professional annotators for manual annotation. Initially, we conducted detailed preliminary training and developed a comprehensive annotation manual tailored to the task, ensuring that all annotators clearly understood the requirements. Next, the annotators performed trial annotations, which were evaluated by two experts. Only those achieving an accuracy rate of at least 90\% were permitted to proceed to the main annotation tasks. Finally, we selected 31 annotators to label the comment art dimensions. To guarantee high quality, each annotator was required to watch the video and gain an in-depth understanding of its content before beginning the annotation process. During annotation, they assigned the appropriate comment art dimension labels to each comment and provided detailed justifications for their choices. Subsequently, specialized reviewers conducted quality checks, and any annotations that did not meet the required standards were returned for re-annotation, thereby ensuring the accuracy and consistency of the final results. Fig.\ref{fig:annotation_interface} illustrates the complete annotation interface.

Through this rigorous annotation and review process, all \textbf{GOD-level Comments} in GODBench were assigned different \textit{Comment Art} dimension labels, with each comment potentially corresponding to multiple subcategories under different dimensions. These annotations provide high-quality foundational data for subsequent multimodal learning and analysis tasks.

\section{Human Evaluation}
\label{sec:human_evaluation}
\begin{figure*}[htbp]
    \centering
    \fbox{\includegraphics[width=0.9\textwidth]{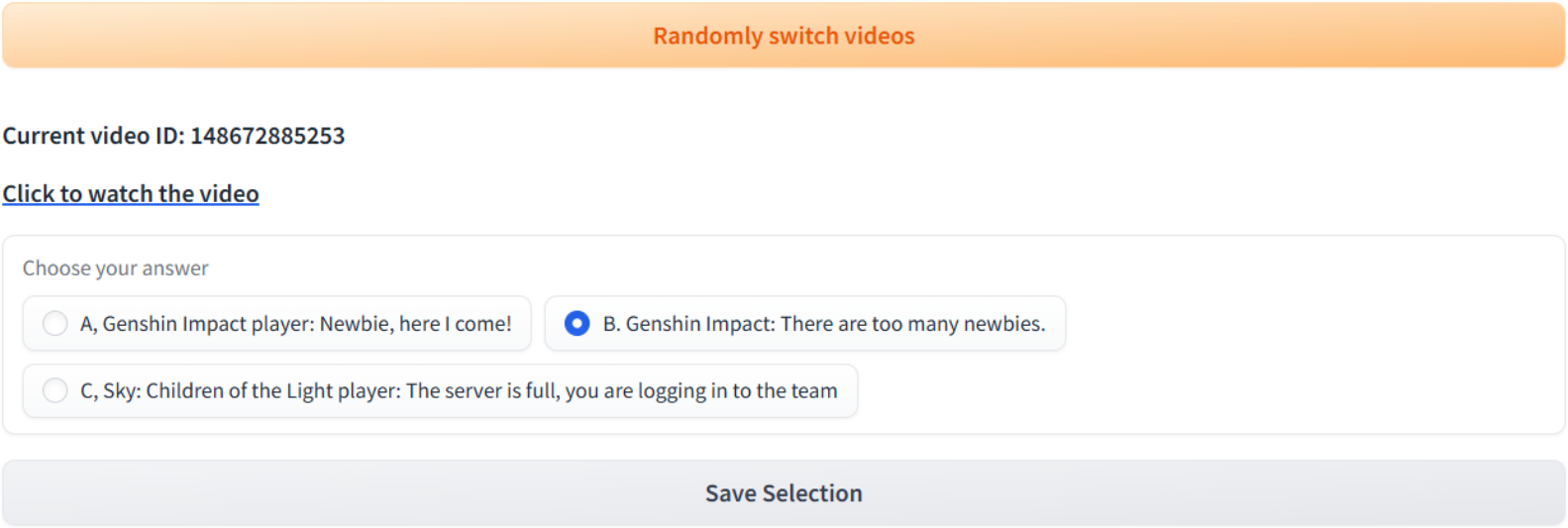}}
    \caption{
    \textbf{Manual evaluation selection task.}
    This interface presents a selection task for the [1, 1, 1] category. It includes randomly generated questions, a link to the corresponding video, and the associated multiple-choice questions.
    }
    \label{fig:1G3}
\end{figure*}

\begin{figure*}[htbp]
    \centering
    \fbox{\includegraphics[width=0.9\linewidth
    ]{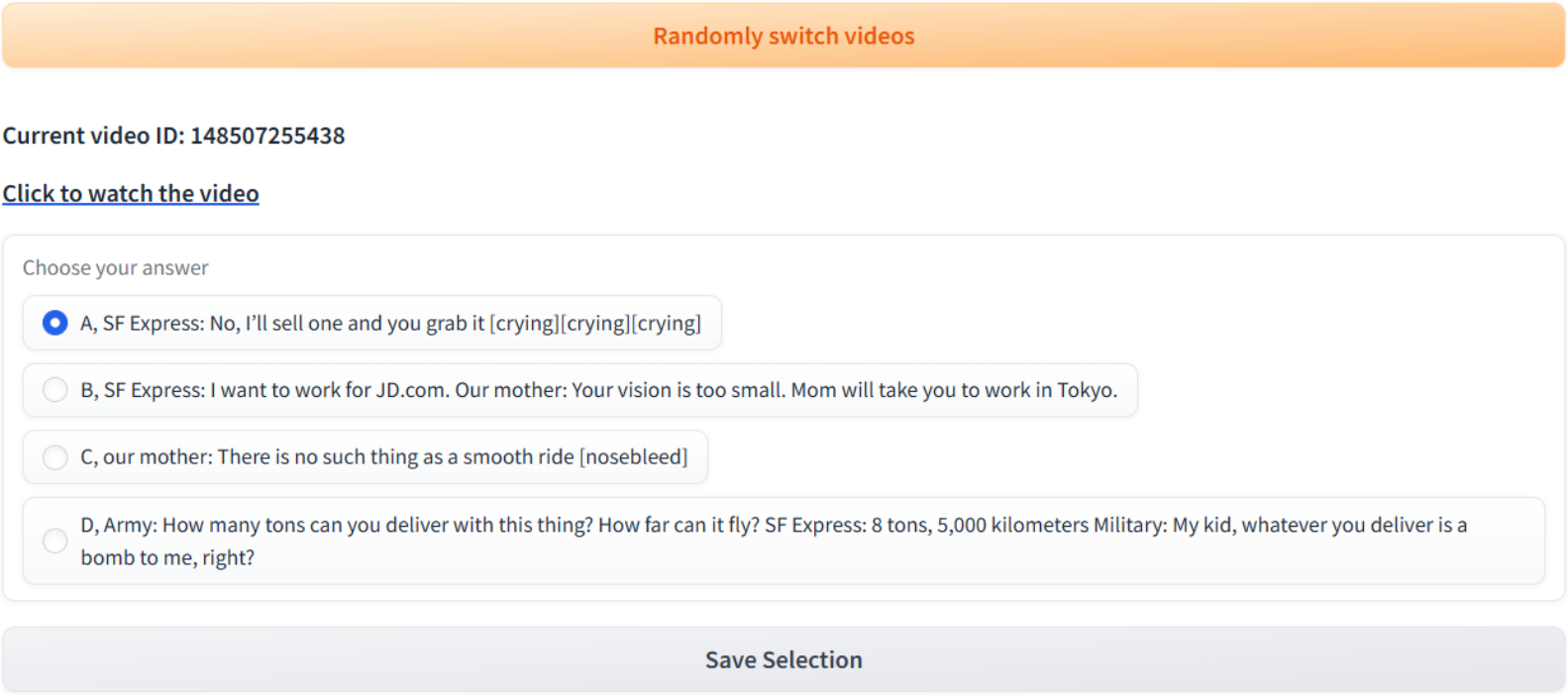}}
    \caption{
    \textbf{Manual evaluation selection task.}
    This interface presents a selection task for the [1, 3, 0] category.
    }
    \label{fig:1G3Q}
    \vspace{-0.5cm}
\end{figure*}

\begin{figure*}[htbp]
    \centering
    \fbox{\includegraphics[width=0.9\linewidth
    ]{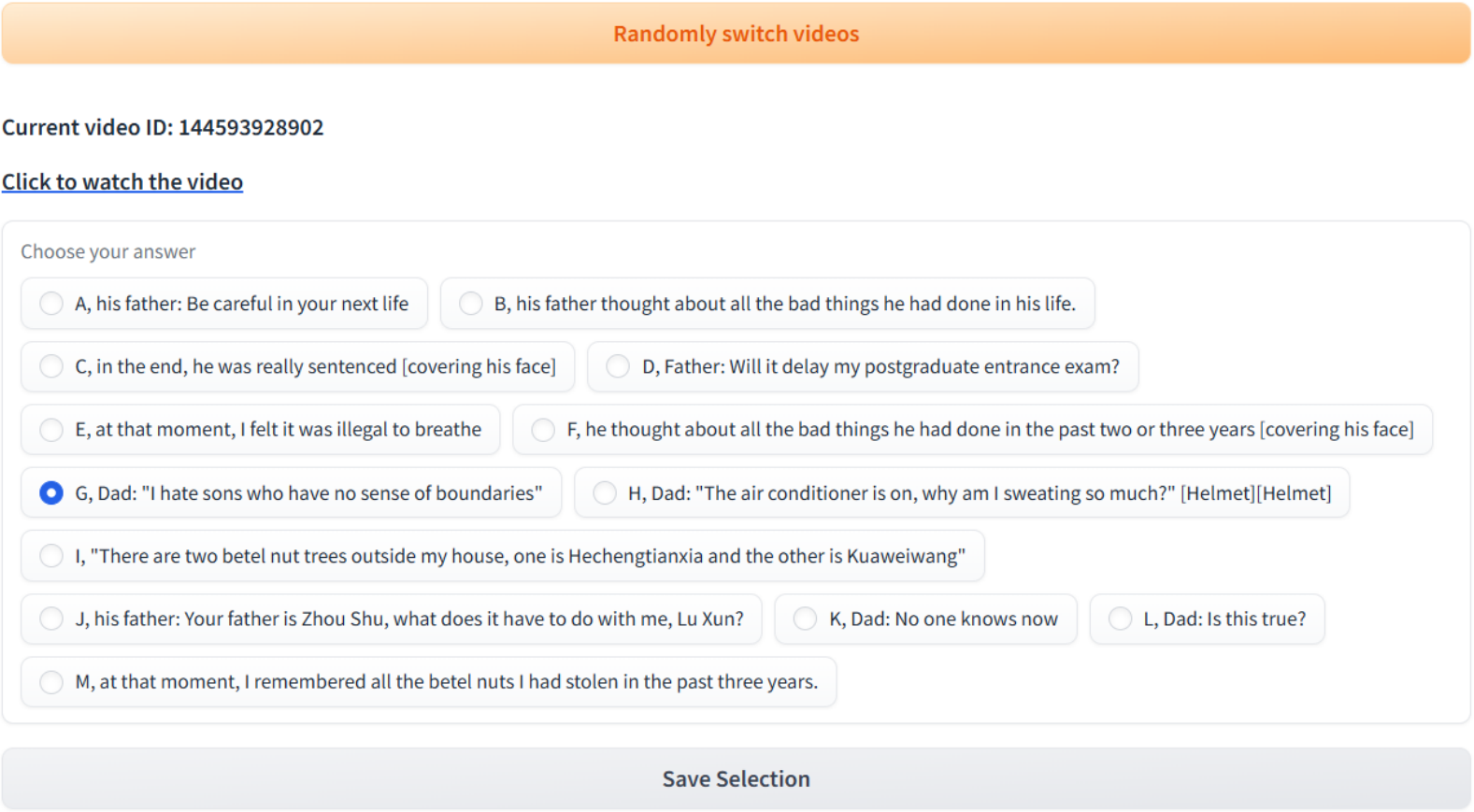}}
    \caption{
    \textbf{Manual evaluation selection task.}
    This interface presents a selection task for the [1, 12, 0] category.
    }
    \label{fig:1G12Q}
    \vspace{-0.5cm}
\end{figure*}

\begin{figure*}[htbp]
    \centering
    \fbox{\includegraphics[width=0.9\linewidth
    ]{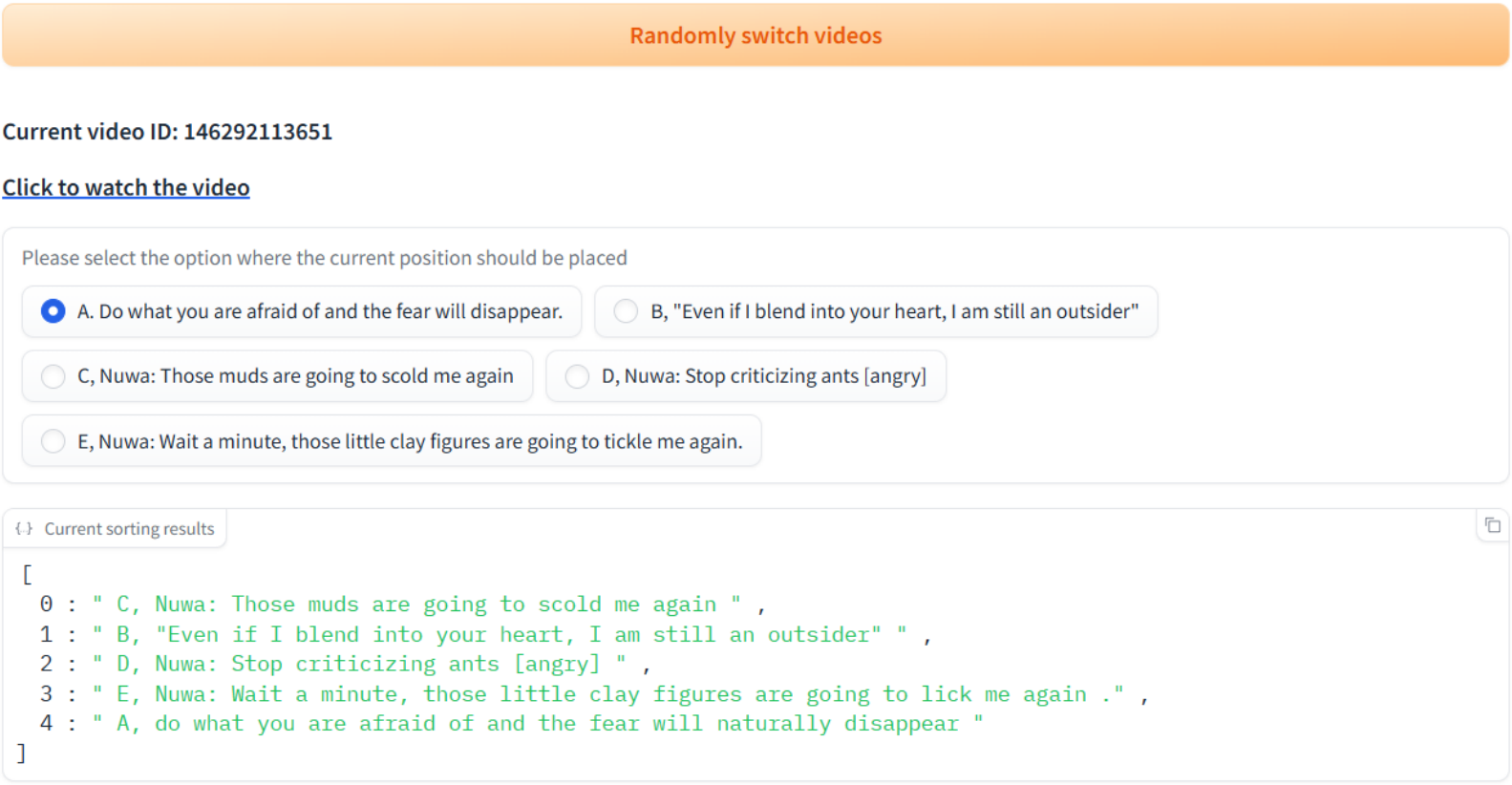}}
    \caption{
    \textbf{Manual evaluation ranking task.}
     This interface contains multiple-choice questions for the [1, 2, 2] type, where users need to click on the options in order to get the ranking results.
    }
    \label{fig:rank}
    \vspace{-0.5cm}
\end{figure*}

\begin{figure*}[htbp]
    \centering
    \fbox{\includegraphics[width=0.9\linewidth
    ]{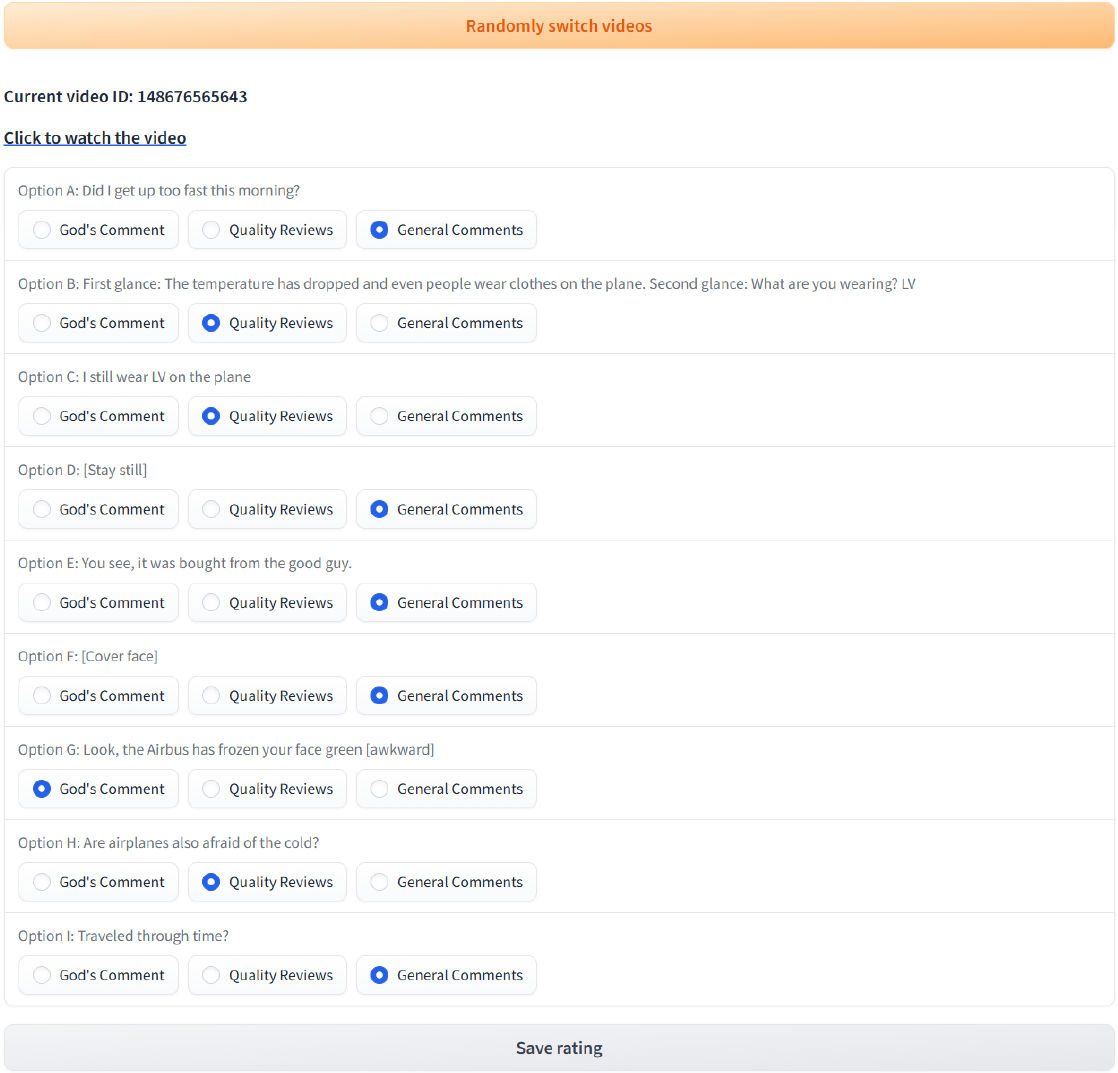}}
    \caption{
    \textbf{Manual evaluation classification task.}
    This interface contains classifications for the [1, 3, 5] type, where users need to assign each option to different quality categories.
    }
    \label{fig:classify}
    \vspace{-0.5cm}
\end{figure*}

In the discriminative task, to compare the differences between current MLLMs and real humans, we invited 10 volunteers to participate in the testing. The test content covered all tasks in the discrimination category, including selection, ranking, and classification tasks. In the selection task, several types were designed, including: [1,1,1], where the goal is to select the GOD-level comment from one GOD-level comment, one High-Quality Comment, and one Ordinary Comment, with the test interface shown in Fig.\ref{fig:1G3}; [1,3,0], where the task is to select the GOD-level comment from one GOD-level comment and three High-Quality Comments, with the test interface shown in Fig.\ref{fig:1G3Q}; and [1,12,0], where the objective is to select the GOD-level comment from one GOD-level comment and twelve High-Quality Comments, with the test interface shown in Fig.\ref{fig:1G12Q}. The ranking task was designed as [1,4,0], where participants are asked to rank one GOD-level comment and four High-Quality Comments according to their creativity, with the test interface shown in Fig.\ref{fig:rank}. The classification task used [1,3,5], where participants need to classify one GOD-level comment, three High-Quality Comments, and five Ordinary Comments, with the test interface shown in Fig.\ref{fig:classify}. These tasks were designed to comprehensively evaluate the differences between MLLMs and humans when handling various creative comments.

\begin{figure*}[t]
    \centering
    \fbox{\includegraphics[width=0.9\linewidth
    ]{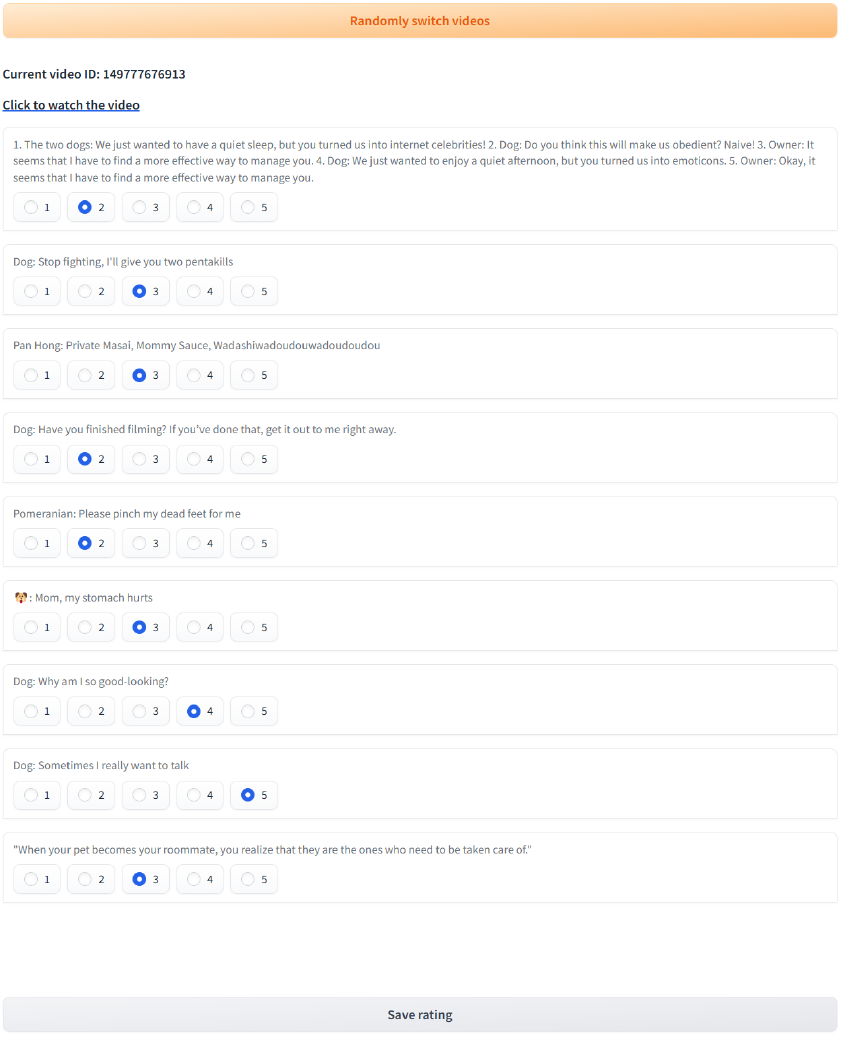}}
    \caption{
    \textbf{Manually scoring the comments generated by the Qwen2-VL Series.}
    Human experts scored the comments generated by the Qwen2VL series models, including Qwen2-VL+ROT (ours), Qwen2-VL 5-shot, Qwen2-VL+CCoT, Qwen2-VL+CoT, and the original Qwen2-VL model. 
    }
    \label{fig:qwen}
    \vspace{-0.5cm}
\end{figure*}

\begin{figure*}[t]
    \centering
    \fbox{\includegraphics[width=0.9\linewidth
    ]{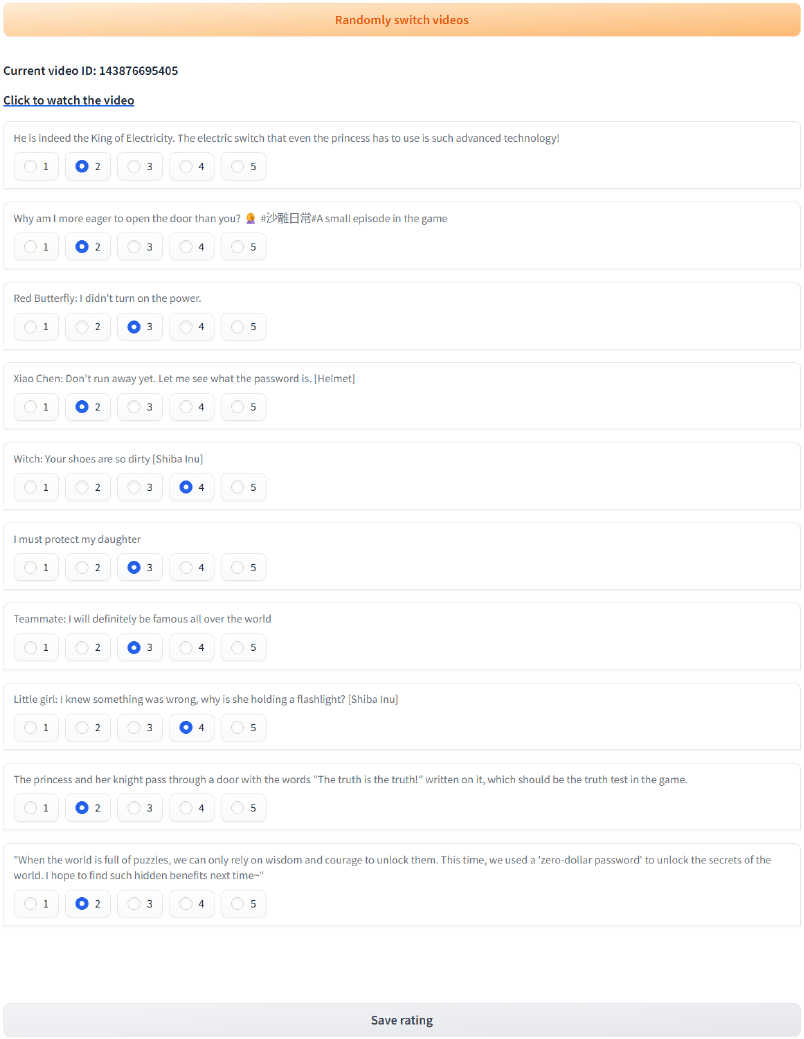}}
    \caption{
    \textbf{Manually scoring the comments generated by the InternVL2.5 Series.}
   Human experts scored the comments generated by the InternVL2.5 series models, including InternVL2.5+ROT (ours), InternVL2.5 5-shot, InternVL2.5+CCoT, InternVL2.5+CoT, and the original InternVL2.5 model. 
    }
    \label{fig:internvl}
    \vspace{-0.5cm}
\end{figure*}

\begin{figure*}[t]
    \centering
    \fbox{\includegraphics[width=0.8\linewidth
    ]{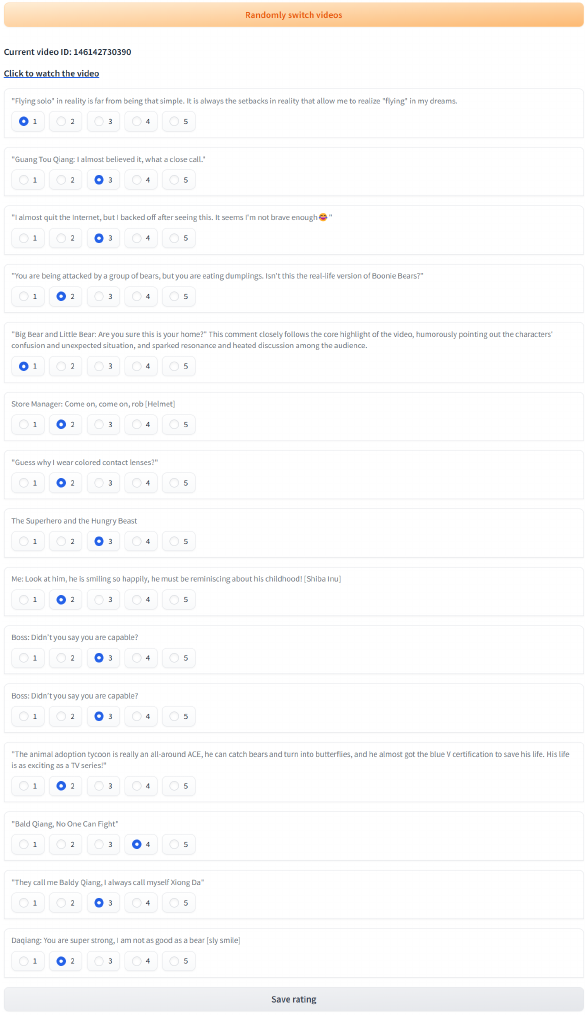}}
    \caption{
    \textbf{Manually scoring the comments generated by all models mixed with real GOD-level Comments.}
    Human experts scored the comments generated by GOD-level Comments, InternVL2.5+ROT (ours), Qwen2-VL+ROT (ours), InternVL2.5 5-shot, Qwen2-VL 5-shot, InternVL2.5+CCoT, InternVL2.5+CoT, Qwen2-VL+CCoT, Qwen2-VL+CoT, mPLUG-Owl3, MiniCPM-o 2.6, MiniCPM-V 2.6 and LLaVA-Video. 
    }
    \label{fig:allmodel}
    \vspace{-0.5cm}
\end{figure*}

In generation tasks, automatically evaluating the comments generated by the model is a highly challenging task. Traditional NLP evaluation metrics often fail to accurately reflect the differences in creativity, and using LLM-based scoring methods relies heavily on the LLM's own capabilities. Moreover, current LLMs struggle to effectively differentiate between the quality of comments. Therefore, the most reliable and fair evaluation method remains the human expert assessment. To this end, we designed multiple evaluation tasks, including comparing results generated by different methods using the same series of models, such as Qwen2-VL and InternVL2.5, and comparing the results generated by all models mixed with real GOD-level comments. As shown in Fig.\ref{fig:qwen}, \ref{fig:internvl} and \ref{fig:allmodel}, human experts are required to score the anonymized model-generated results, ensuring fairness and accuracy in the evaluation.

\section{More Details of Experiment}
\subsection{Heuristic Baselines in Discrimination Tasks}
We provide further details on the heuristic baselines introduced in Sec. \ref{sec:results}: \textit{Random Choice}, \textit{Frequent Guess}, and \textit{Human Evaluation}.

\textbf{Random Choice.} The Random Choice baseline selects an answer randomly from the answer pool for each question and averages the results over five trials, representing the expected outcome of simple random guessing.

\textbf{Frequent Guess.} Based on the option distribution in each task category of discriminative tasks, we select the most frequently occurring option as the predicted answer for the corresponding task. This baseline demonstrates whether the option distribution in GODBench is balanced and serves as a straightforward yet informative reference for evaluating model performance, representing the expected outcome of consistently selecting the most common answer.

\textbf{Human Evaluation.} The human evaluation baseline reflects human performance on GODBench, serving as a reliable upper bound for assessing model capabilities. To facilitate this process, we develop a structured manual evaluation workflow with a user-friendly interface, detailed in Appendix ~\ref{sec:human_evaluation}.  

\subsection{Implementation Details.}
\label{sec:implementation_details}
\subsubsection{Training Details}
\label{sec:training_details}
As a widely used open-source fine-tuning framework, LlamaFactory~\cite{zheng2024llamafactory} is employed to fine-tune MLLMs in our experiments. Specifically, to enhance the understanding of GOD-level comments, we construct an instruction-tuning dataset tailored for discriminative tasks such as selection, ranking, and classification. This dataset includes a diverse range of comments to improve the model's ability to identify high-quality comments.
All MLLMs are fine-tuned using 8×NVIDIA A800 (80G) GPUs, with a learning rate of 5e-6, a batch size of 8 (8×1), and trained for one epoch.

\subsubsection{Inference Details}
\label{sec:inference_details}
\textbf{GPT-4o and GPT-4o-mini.} 
Due to API limitations, we uniformly sampled 50 frames from each video for evaluation on GODBench. The model input adopts the format of ``\texttt{<frames> + <prompt> + <comments>(only for discriminative tasks)}''.
\\
\textbf{Qwen2-VL.} 
Following Video-MME~\cite{fu2024video}, we adapt a dynamic frame sampling strategy based on video duration. Specifically, videos shorter than 128 seconds are sampled at 1 fps, and videos shorter than 768 seconds are sampled at 0.5 fps. For videos longer than 768 seconds, we extract 384 frames uniformly. The model input adopts the format of ``\texttt{<frames> + <prompt> + <comments>(only for discriminative tasks)}''.
\\
\textbf{Other Open-Source \textit{Video}-MLLMs.} 
We adhere to the official inference strategies of these MLLMs. The model input adopts the format of ``\texttt{<frames> + <prompt> + <comments>(only for discriminative tasks)}''.
\\
\textbf{5-shot settings.}  For the 5-shot input, we first determine the tag set for each test video. Then, for each test video, we randomly sample 10 videos from the training set that share at least one tag. These videos' corresponding GOD-level comments are ranked in descending order by length and the number of likes. Finally, the top 5 comments are selected as the 5-shot reference comments.

\subsubsection{Prompt for Inference}
\begin{figure*}[t]
    \centering
    \includegraphics[width=0.9\linewidth
    ]{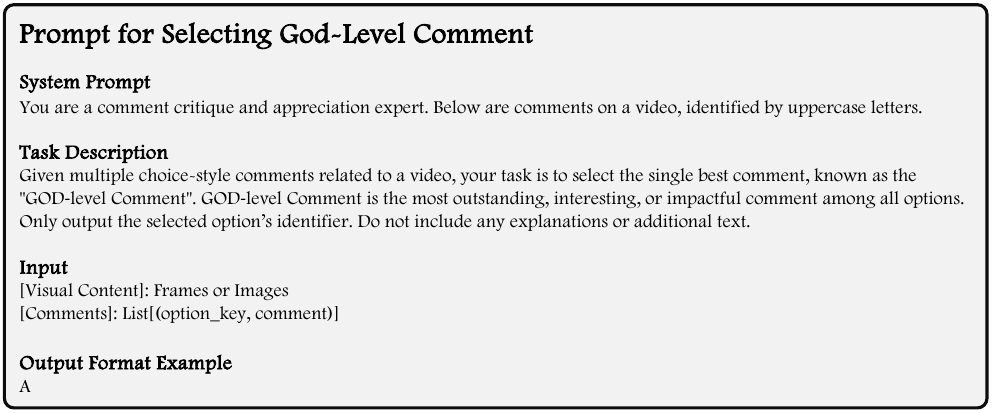}
    \caption{
    \textbf{Inference Prompt for Selection task.}
    }
    \label{fig:prompt_for_selection}
    \vspace{-0.5cm}
\end{figure*}

\begin{figure*}[t]
    \centering
    \includegraphics[width=0.9\linewidth
    ]{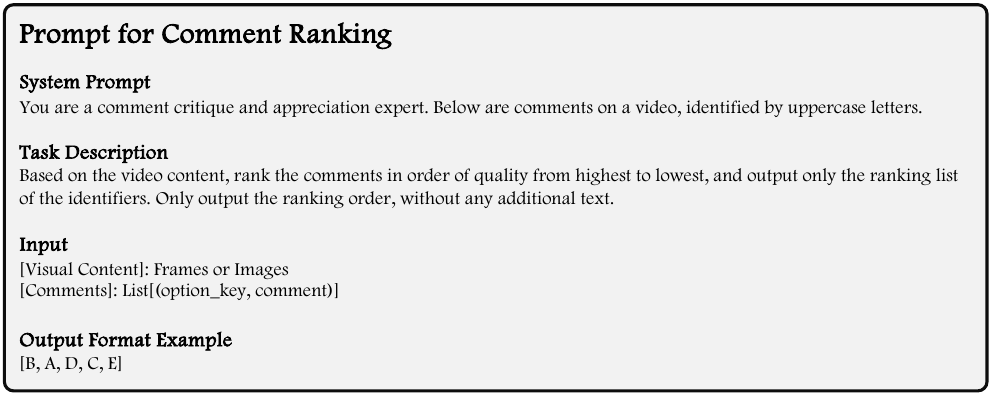}
    \caption{
    \textbf{Inference Prompt for Ranking task.}
    }
    \label{fig:prompt_for_ranking}
    \vspace{-0.5cm}
\end{figure*}

\begin{figure*}[t]
    \centering
    \includegraphics[width=0.9\linewidth
    ]{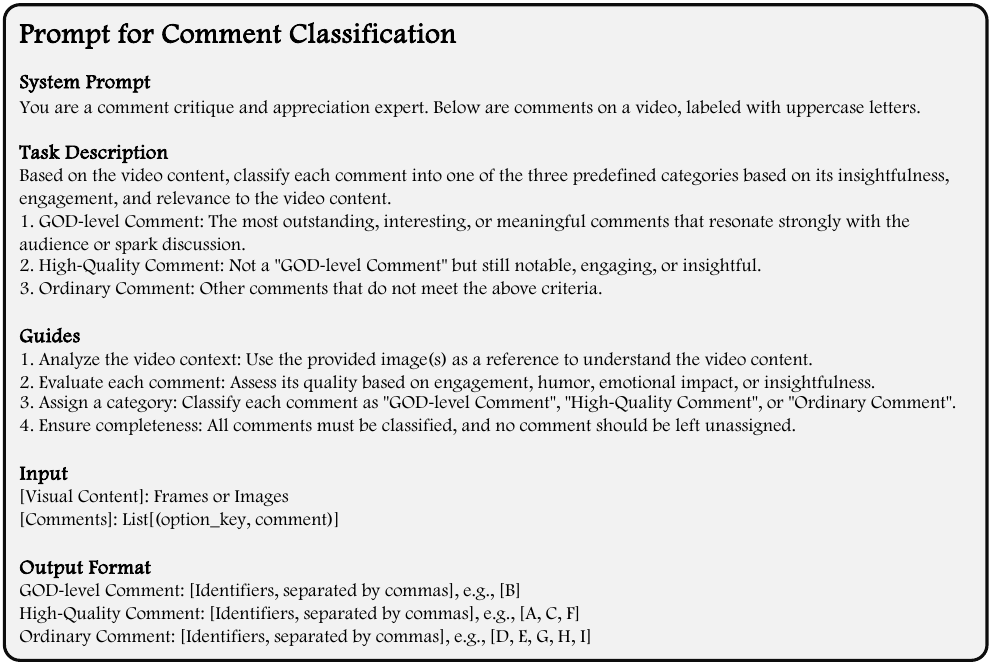}
    \caption{
    \textbf{Inference Prompt for Classification task.}
    }
    \label{fig:prompt_for_classification}
    \vspace{-0.5cm}
\end{figure*}

\begin{figure*}[t]
    \centering
    \includegraphics[width=0.9\linewidth
    ]{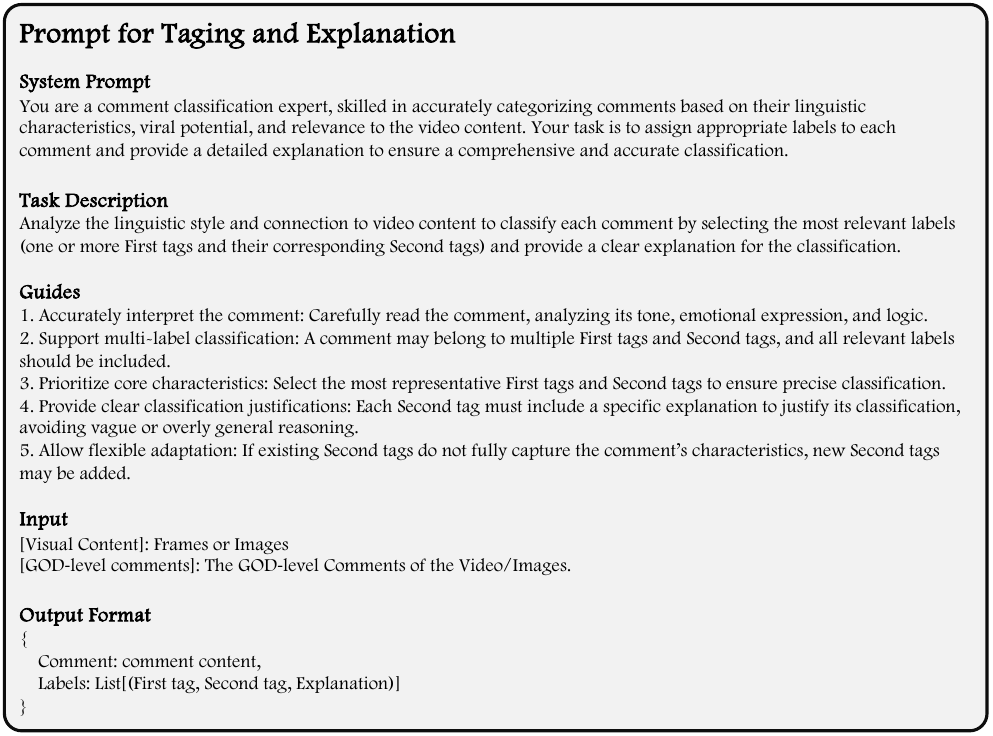}
    \caption{
    \textbf{Inference Prompt for Tag Explanation task.}
    }
    \label{fig:prompt_for_explanation}
    \vspace{-0.5cm}
\end{figure*}

\begin{figure*}[t]
    \centering
    \includegraphics[width=0.9\linewidth
    ]{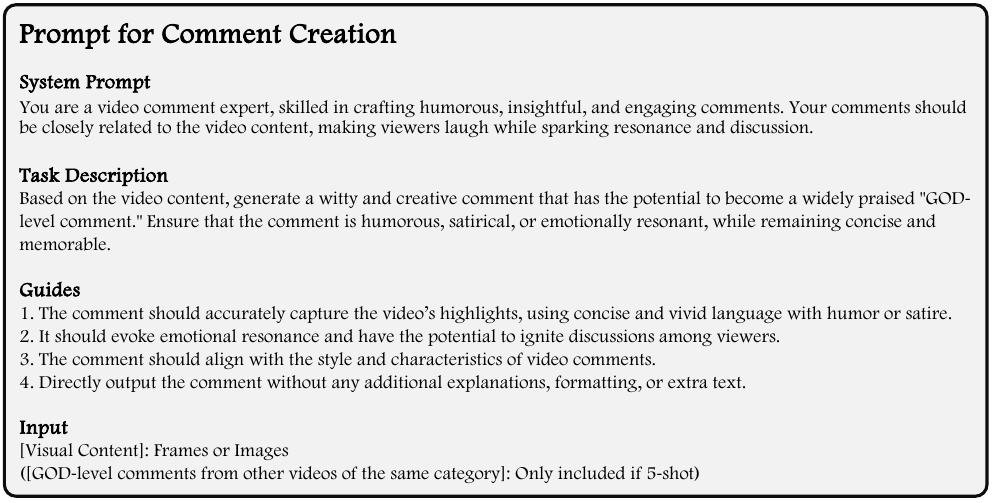}
    \caption{
    \textbf{Inference Prompt for Comment Creation task.}
    }
    \label{fig:prompt_for_creation}
    \vspace{-0.5cm}
\end{figure*}

We provide detailed prompt templates for evaluating the model's performance on GODBench, including Fig.~\ref{fig:prompt_for_selection}, Fig.~\ref{fig:prompt_for_ranking}, and Fig.~\ref{fig:prompt_for_classification} for discriminative tasks, as well as Fig.~\ref{fig:prompt_for_explanation} and Fig.~\ref{fig:prompt_for_creation} for generative tasks. The corresponding prompts are inspired by \textit{Using LLM-as-a-Judge}~\cite{huggingface_llm_judge} and make some adaptions to literature translation and task-specific requirements.


\subsection{GPT-4o Judgement}
\label{sec:gpt4o_judgement}
\begin{figure*}[t]
    \centering
    \includegraphics[width=0.9\linewidth
    ]{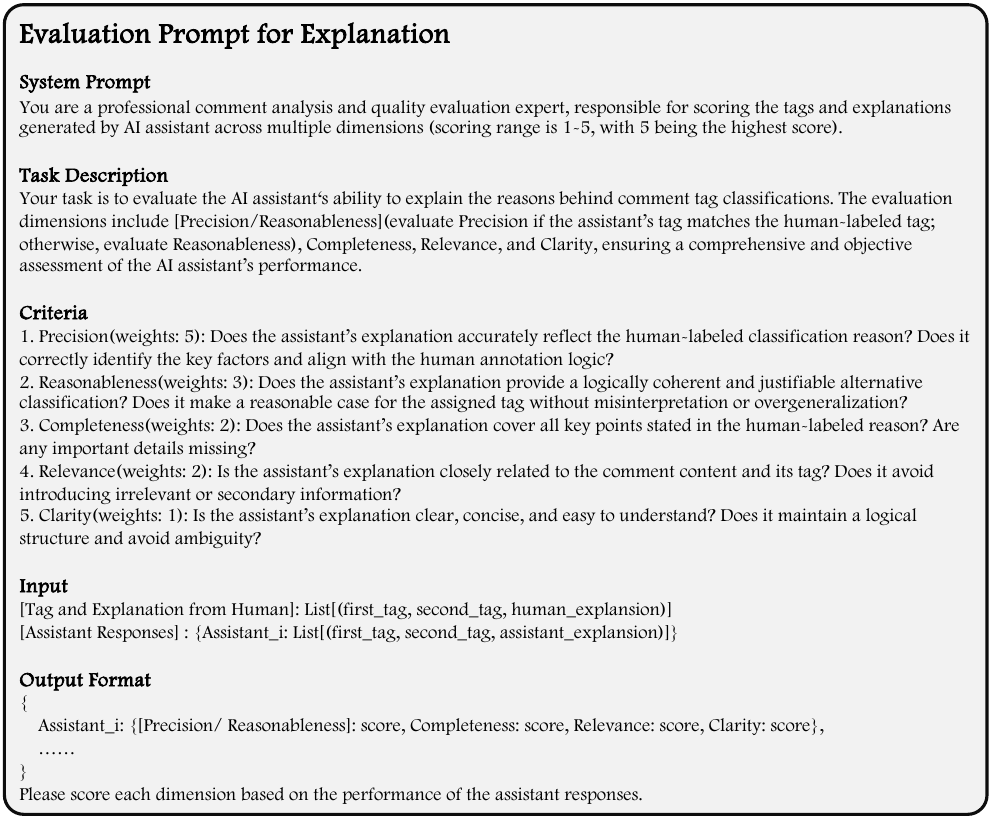}
    \caption{
    \textbf{Evaluation Prompt for Tag Explanation task.}
    }
    \label{fig:prompt_for_eval_explanation}
    \vspace{-0.5cm}
\end{figure*}

\begin{figure*}[t]
    \centering
    \vspace{-0.5cm}
    \includegraphics[width=0.9\linewidth]{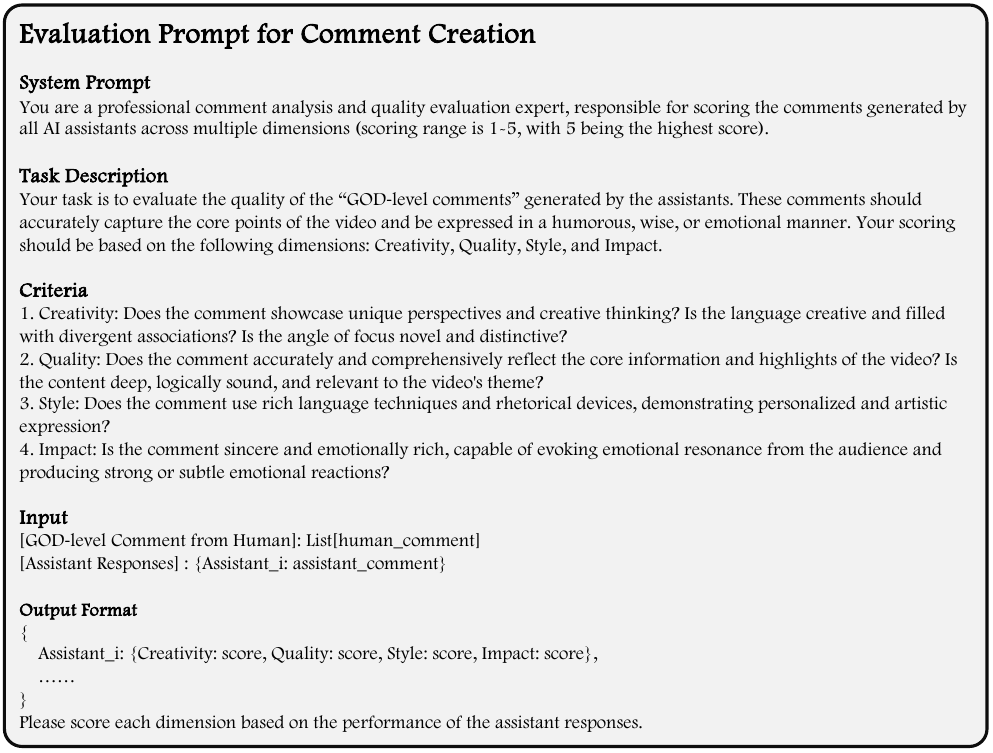}
    \caption{
    \textbf{Evaluation Prompt for Comment Creation task.}
    }
    \label{fig:prompt_for_eval_creation}
    \vspace{-0.5cm}
\end{figure*}

\begin{figure*}[t]
    \centering
    \vspace{-0.5cm}
    \includegraphics[width=0.9\linewidth]{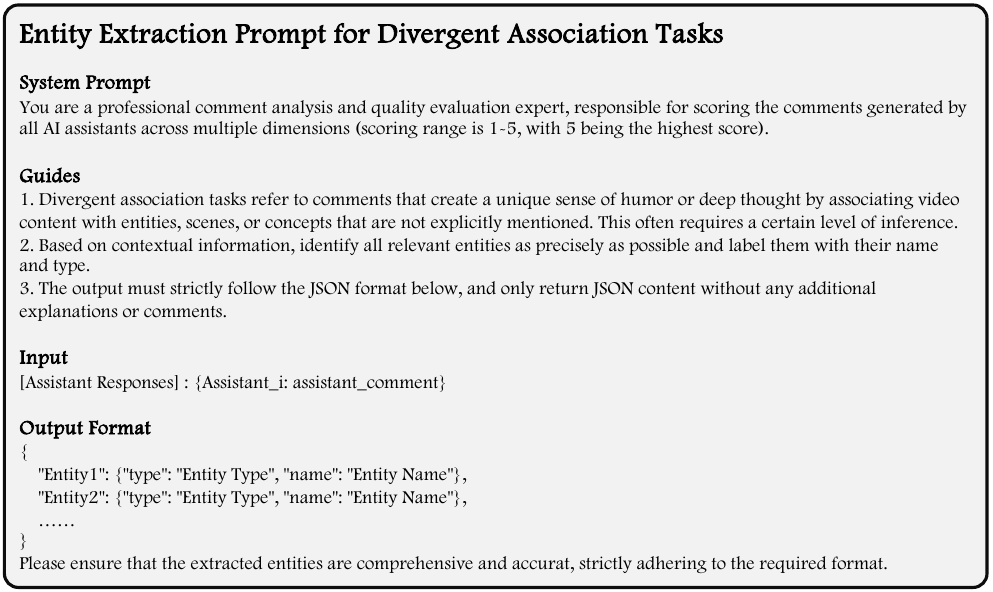}
    \caption{
    \textbf{Entity Extraction Prompt for Divergent Association Task.}
    }
    \label{fig:prompt_for_eval_divergent}
    \vspace{-0.5cm}
\end{figure*}

For generative tasks, we utilize GPT-4o as the judge model to assess the quality of explanation and creation tasks. 
For the explanation task, we evaluate responses based on five criteria: \textit{Precision}, \textit{Reasonableness}, \textit{Completeness}, \textit{Relevance}, and \textit{Clarity}, each scored on a scale of 0–5. The criteria are weighted as [5,3,2,2,1], and GPT-4o assigns scores by referencing human-annotated explanations.
For the creation task, we assess responses using four criteria: \textit{Creativity}, \textit{Quality}, \textit{Style}, and \textit{Impact}, also scored on a scale of 0–5. GPT-4o evaluates these aspects by referencing the corresponding GOD-level comment of the video.
The prompt templates are illustrated in Fig.~\ref{fig:prompt_for_eval_explanation} and Fig.~\ref{fig:prompt_for_eval_creation}. 
We also employ GPT-4o to extract entities with divergent associations, using the prompt template shown in Fig.~\ref{fig:prompt_for_eval_divergent}. 

\subsection{More Experimental Results}
\subsubsection{Fine-Grained Dimensions of Discrimination Tasks}
We provide additional experimental results on fine-grained dimensions across discriminative tasks, including selection (Tab.~\ref{tab:finegrain_s111}, Tab.~\ref{tab:finegrain_s130}, Tab.~\ref{tab:finegrain_s1120}), ranking (Tab.~\ref{tab:finegrain_r140}), and classification (Tab.~\ref{tab:finegrain_c135}). The results demonstrate that MLLMs fine-tuned with LoRA achieve competitive performance across all fine-grained dimensions, significantly outperforming baseline models. Furthermore, we observe that GPT-4o substantially outperforms open-source models across a wide range of fine-grained dimensions, highlighting the considerable gap that still remains.

\begin{table}[ht!]
    \centering
    \small
    \renewcommand\tabcolsep{6pt} 
    \renewcommand\arraystretch{0.80} 
    \resizebox{1.0\linewidth}{!}{
        \begin{tabular}{r|c|c|c|c|c}
            \toprule
            \multicolumn{1}{c|}{\multirow{2}{*}{\textbf{Model}}}
            & \multicolumn{5}{c}{\textbf{$S^{[1,1,1]}_{\textnormal{acc}}$}} \\
            \cmidrule{2-6}
            & \textbf{RT} & \textbf{DA} & \textbf{WT} & \textbf{IV} & \textbf{ER} \\

            \midrule
            LLaVA-Video & 39.17 & 32.89 & 37.37 & 36.07 & 44.62 \\
            mPLUG-Owl3 & 39.39 & 31.31 & 35.26 & 31.35 & 41.54 \\
            MiniCPM-V 2.6 & 41.77 & 41.05 & 35.26 & 38.52 & 52.31 \\
            MiniCPM-o 2.6 & 40.76 & 40.24 & 35.79 & 39.75 & 50.77 \\
            
            \midrule
            Qwen2-VL$_{\text{2B}}$ & 37.69 & 36.36 & 36.32 & 40.37 & 44.23 \\
            Qwen2-VL$_{\text{7B}}$ & 46.32 & 45.45 & 46.84 & 43.24 & 51.54 \\
            \midrule
            InternVL2.5$_{\text{8B}}$ & 46.43 & 42.83 & 46.84 & 44.06 & 47.69 \\
            InternVL2.5$_{\text{26B}}$ & 46.32 & 44.65 & 44.74 & 44.67 & 51.54 \\
            \midrule
            \rowcolor{gray!20} 
            \multicolumn{6}{c}{\textit{Commercial MLLMs}} \\
            \midrule

            GPT-4o-mini
            & 44.47 & 44.81 & 45.26 & 39.55 & 51.92 \\
            GPT-4o
            & 56.22 & 53.86 & 45.79 & 48.88 & 57.31 \\
            
            \midrule
            \rowcolor{gray!20} 
            \multicolumn{6}{c}{\textit{MLLMs after Supervised Fine-Tuning}} \\
            \midrule
            \rowcolor{blue!10} 
            Qwen2-VL$_{\text{LoRA}}$
            & 68.76 & 76.44 & 63.78 & 66.94 & 66.40 \\

            \rowcolor{blue!10} 
            InternVL2.5$_{\text{LoRA}}$
            & 72.72 & 78.39 & 72.43 & 73.80 & 67.98 \\
            \bottomrule
        \end{tabular}
    }
    \vspace{-5pt} 
    \caption{\textbf{Performance of MLLMs on $S^{[1,1,1]}_{\textnormal{acc}}$ in Selection Tasks.}}
    \label{tab:finegrain_s111}
\end{table}

\begin{table}[!htbp]
    \centering
    \small
    \renewcommand\tabcolsep{6pt} 
    \renewcommand\arraystretch{0.80} 
    \resizebox{1.0\linewidth}{!}{
        \begin{tabular}{r|c|c|c|c|c}
            \toprule
            \multicolumn{1}{c|}{\multirow{2}{*}{\textbf{Model}}}
            & \multicolumn{5}{c}{\textbf{$S^{[1,3]}_{\textnormal{acc}}$}}\\
            \cmidrule{2-6}
            & \textbf{RT} & \textbf{DA} & \textbf{WT} & \textbf{IV} & \textbf{ER} \\

            \midrule
            LLaVA-Video & 27.69 & 20.93 & 22.63 & 22.95 & 30.77 \\
            mPLUG-Owl3 & 28.27 & 21.94 & 23.16 & 22.34 & 32.31 \\
            MiniCPM-V 2.6 & 30.39 & 27.88 & 24.74 & 26.84 & 34.62 \\
            MiniCPM-o 2.6 & 30.23 & 26.02 & 22.63 & 25.20 & 36.92 \\
            
            \midrule
            Qwen2-VL$_{\text{2B}}$ & 27.74 & 24.08 & 28.42 & 24.80 & 33.46 \\
            Qwen2-VL$_{\text{7B}}$ & 32.82 & 28.85 & 30.00 & 26.84 & 39.62 \\
            \midrule
            InternVL2.5$_{\text{8B}}$ & 30.33 & 27.76 & 23.16 & 24.80 & 34.62 \\
            InternVL2.5$_{\text{26B}}$ & 33.77 & 30.02 & 30.53 & 25.82 & 36.15 \\
            \midrule
            \rowcolor{gray!20} 
            \multicolumn{6}{c}{\textit{Commercial MLLMs}} \\
            \midrule
            GPT-4o-mini & 30.76 & 28.16 & 26.32 & 25.00 & 27.31 \\
            GPT-4o & 40.92 & 37.05 & 35.79 & 33.20 & 36.15 \\
            
            \midrule
            \rowcolor{gray!20} 
            \multicolumn{6}{c}{\textit{MLLMs after Supervised Fine-Tuning}} \\
            \midrule
            \rowcolor{blue!10} 
            Qwen2-VL$_{\text{LoRA}}$
            & 51.85 & 60.60 & 44.86 & 52.37 & 50.59 \\

            \rowcolor{blue!10} 
            InternVL2.5$_{\text{LoRA}}$
            & 55.98 & 63.09 & 51.35 & 60.41 & 54.94 \\

            \bottomrule
        \end{tabular}
    }
    \vspace{-5pt} 
    \caption{\textbf{Performance of MLLMs on $S^{[1,3]}_{\textnormal{acc}}$ in Selection Tasks.}}
    \label{tab:finegrain_s130}
\end{table}

\begin{table}[ht!]
    \centering
    \small
    \renewcommand\tabcolsep{6pt} 
    \renewcommand\arraystretch{0.80} 
    \resizebox{1.0\linewidth}{!}{
        \begin{tabular}{r|c|c|c|c|c}
            \toprule
            \multicolumn{1}{c|}{\multirow{2}{*}{\textbf{Model}}}
            & \multicolumn{5}{c}{\textbf{$S^{[1,12]}_{\textnormal{acc}}$}}\\
            \cmidrule{2-6}
            & \textbf{RT} & \textbf{DA} & \textbf{WT} & \textbf{IV} & \textbf{ER} \\

            \midrule
            LLaVA-Video & 12.02 & 8.40  & 8.42  & 10.86 & 15.00 \\
            mPLUG-Owl3  & 12.81 & 7.84  & 9.47  & 8.61  & 13.08 \\
            MiniCPM-V 2.6 & 12.92 & 10.63 & 12.11 & 11.07 & 16.15 \\
            MiniCPM-o 2.6 & 12.71 & 11.07 & 11.05 & 11.27 & 13.46 \\
            
            \midrule
            Qwen2-VL$_{\text{2B}}$ & 8.15  & 7.43  & 12.63 & 8.61  & 11.92 \\
            Qwen2-VL$_{\text{7B}}$ & 14.93 & 12.28 & 11.58 & 11.89 & 20.00 \\
            \midrule
            InternVL2.5$_{\text{8B}}$ & 14.56 & 12.16 & 12.11 & 8.61  & 17.69 \\
            InternVL2.5$_{\text{26B}}$ & 15.03 & 12.00 & 12.11 & 14.14 & 14.23 \\
            \midrule
            \rowcolor{gray!20} 
            \multicolumn{6}{c}{\textit{Commercial MLLMs}} \\
            \midrule

            GPT-4o-mini
            & 13.92 & 13.54 & 11.58 & 14.14 & 14.62 \\
            GPT-4o
            & 20.96 & 17.45 & 14.21 & 15.37 & 18.46 \\
            
            \midrule
            \rowcolor{gray!20} 
            \multicolumn{6}{c}{\textit{MLLMs after Supervised Fine-Tuning}} \\
            \midrule
            \rowcolor{blue!10} 
            Qwen2-VL$_{\text{LoRA}}$
            & 28.72 & 34.80 & 19.88 & 30.73 & 25.63 \\

            \rowcolor{blue!10} 
            InternVL2.5$_{\text{LoRA}}$
            & 32.25 & 36.90 & 32.75 & 33.03 & 27.73 \\
            \bottomrule
        \end{tabular}
    }
    \vspace{-5pt} 
    \caption{\textbf{Performance of MLLMs on $S^{[1,12]}_{\textnormal{acc}}$ in Selection Tasks.}}
    \label{tab:finegrain_s1120}
\end{table}

\begin{table}[ht!]
    \centering
    \small
    \renewcommand\tabcolsep{6pt} 
    \renewcommand\arraystretch{0.80} 
    \resizebox{1.0\linewidth}{!}{
        \begin{tabular}{r|c|c|c|c|c}
            \toprule
            \multicolumn{1}{c|}{\multirow{2}{*}{\textbf{Model}}}
            & \multicolumn{5}{c}{\textbf{$R^{[1,4]}_{\textnormal{NDCG}}$}}\\
            \cmidrule{2-6}
            & \textbf{RT} & \textbf{DA} & \textbf{WT} & \textbf{IV} & \textbf{ER} \\
            
            \midrule
            LLaVA-Video & 50.53 & 48.17 & 46.35 & 46.96 & 57.47 \\
            mPLUG-Owl3  & 62.87 & 61.45 & 57.87 & 59.59 & 75.71 \\
            MiniCPM-V 2.6 & 58.05 & 57.98 & 60.87 & 60.04 & 63.29 \\
            MiniCPM-o 2.6 & 54.31 & 52.34 & 51.41 & 53.45 & 56.56 \\
            
            \midrule
            Qwen2-VL$_{\text{2B}}$ & 48.49 & 48.35 & 41.18 & 49.26 & 41.99 \\
            Qwen2-VL$_{\text{7B}}$ & 62.80 & 62.43 & 64.01 & 62.13 & 66.10 \\
            \midrule
            InternVL2.5$_{\text{8B}}$ & 46.21 & 45.35 & 43.48 & 43.92 & 47.72 \\
            InternVL2.5$_{\text{26B}}$ & 53.54 & 53.14 & 57.37 & 55.76 & 57.13 \\
            \midrule
            \rowcolor{gray!20} 
            \multicolumn{6}{c}{\textit{Commercial MLLMs}} \\
            \midrule

            GPT-4o-mini
            & 61.64 & 63.18 & 60.80 & 62.77 & 65.39 \\
            GPT-4o
            & 64.73 & 64.90 & 66.92 & 63.09 & 69.63 \\
            
            \midrule
            \rowcolor{gray!20} 
            \multicolumn{6}{c}{\textit{MLLMs after Supervised Fine-Tuning}} \\
            \midrule
            \rowcolor{blue!10} 
            Qwen2-VL$_{\text{LoRA}}$
            & 72.68 & 78.62 & 70.89 & 71.51 & 71.37 \\

            \rowcolor{blue!10} 
            InternVL2.5$_{\text{LoRA}}$
            & 75.99 & 79.20 & 77.07 & 76.27 & 73.32 \\

            \bottomrule
        \end{tabular}
    }
    \vspace{-5pt} 
    \caption{\textbf{Performance of MLLMs on $R^{[1,4]}_{\textnormal{NDCG}}$ in Ranking Tasks.} }
    \label{tab:finegrain_r140}
\end{table}

\begin{table}[ht!]
    \centering
    \small
    \renewcommand\tabcolsep{6pt} 
    \renewcommand\arraystretch{0.80} 
    \resizebox{1.0\linewidth}{!}{
        \begin{tabular}{r|c|c|c|c|c}
            \toprule
            \multicolumn{1}{c|}{\multirow{2}{*}{\textbf{Model}}}
            & \multicolumn{5}{c}{\textbf{$C^{[1,3,5]}_{\textnormal{acc}}$}}\\
            \cmidrule{2-6}
            & \textbf{RT} & \textbf{DA} & \textbf{WT} & \textbf{IV} & \textbf{ER} \\

            \midrule
            LLaVA-Video & 39.06 & 38.07 & 37.37 & 39.00 & 38.55 \\
            mPLUG-Owl3  & 36.56 & 36.46 & 35.85 & 37.18 & 35.90 \\
            MiniCPM-V 2.6 & 41.40 & 41.54 & 37.19 & 40.64 & 35.43 \\
            MiniCPM-o 2.6 & 41.90 & 41.18 & 39.94 & 39.82 & 41.15 \\
            
            \midrule
            Qwen2-VL$_{\text{2B}}$ & 28.00 & 28.62 & 29.12 & 29.67 & 30.60 \\
            Qwen2-VL$_{\text{7B}}$ & 38.67 & 38.16 & 38.25 & 38.55 & 39.70 \\
            \midrule
            InternVL2.5$_{\text{8B}}$ & 43.21 & 43.37 & 43.10 & 43.12 & 44.79 \\
            InternVL2.5$_{\text{26B}}$ & 44.06 & 43.95 & 43.80 & 44.08 & 42.26 \\
            \midrule
            \rowcolor{gray!20} 
            \multicolumn{6}{c}{\textit{Commercial MLLMs}} \\
            \midrule

            GPT-4o-mini
            & 38.48 & 38.27 & 36.32 & 38.66 & 41.11 \\
            GPT-4o
            & 53.47 & 53.15 & 51.99 & 50.43 & 54.49 \\
            
            \midrule
            \rowcolor{gray!20} 
            \multicolumn{6}{c}{\textit{MLLMs after Supervised Fine-Tuning}} \\
            \midrule
            \rowcolor{blue!10} 
            Qwen2-VL$_{\text{LoRA}}$
            & 69.96 & 73.00 & 69.19 & 70.26 & 69.30 \\

            \rowcolor{blue!10} 
            InternVL2.5$_{\text{LoRA}}$
            & 74.82 & 76.99 & 74.53 & 75.04 & 74.92 \\

            \bottomrule
        \end{tabular}
    }
    \vspace{-5pt} 
    \caption{\textbf{Performance of MLLMs on $C^{[1,3,5]}_{\textnormal{acc}}$ in Classification Tasks.}}
    \label{tab:finegrain_c135}
\end{table}

\subsubsection{Judgement Results of 5-shot}
To assess the impact of the 5-shot setting, we compare the results of Creation and Divergent Association Entity Overlap, as shown in Tab.~\ref{tab:judgement_results}. Our findings indicate that, compared to the zero-shot setting, most models exhibit improved performance under the 5-shot setting, with some even approaching the performance of CoT and CCoT. This suggests that the 5-shot setting enhances the models' ability to better understand video content and generate more creative comments.
\begin{table}[ht!]
    \centering
    \small
    \renewcommand\tabcolsep{8pt} 
    \renewcommand\arraystretch{0.80} 
    \resizebox{1.0\linewidth}{!}{
        \begin{tabular}{r|c|c|c|c}
            \toprule
            \multicolumn{1}{c|}{\multirow{2}{*}{\textbf{Model}}}
            & \multicolumn{2}{c|}{\textbf{Creation}} & \multicolumn{2}{c}{\textbf{WEO}} \\
            \cmidrule{2-3} \cmidrule{4-5}
            & \textbf{Zeroshot} & \textbf{5-shot} & \textbf{Zeroshot} & \textbf{5-shot} \\
            \midrule
            LLaVA-Video 
            & 322.82 & 337.86 & 12.10 & 14.91 \\
            mPLUG-Owl3 
            & 359.34 & 363.83 & 15.11 & 15.05 \\
            MiniCPM-V 2.6
            & 298.67 & 317.84 & 18.22 & 17.76 \\
            MiniCPM-o 2.6
            & 298.18 & 305.22 & 14.89 & 13.43 \\

            \midrule
            GPT-4o
            & \multicolumn{2}{c|}{425.99} & \multicolumn{2}{c}{22.21} \\
            \midrule
            Qwen2-VL$_{\text{2B}}$ 
            & 316.87 & 356.43 & 19.91 & 19.18 \\
            
            Qwen2-VL$_{\text{7B}}$
            & 332.65 & 352.55 & 24.69 & 27.19 \\

            \midrule
            +CoT
            & \multicolumn{2}{c|}{383.01} & \multicolumn{2}{c}{27.19} \\
            +CCoT
            & \multicolumn{2}{c|}{389.08} & \multicolumn{2}{c}{30.57} \\
            +RoT(Ours)
            & \multicolumn{2}{c|}{389.68} & \multicolumn{2}{c}{32.83} \\

            \midrule
            InternVL2.5$_{\text{26B}}$
            & 353.64 & 362.01 & 18.18 & 21.59\\

            InternVL2.5$_{\text{8B}}$
            & 342.23 & 347.45 & 12.67 & 19.00 \\

            \midrule
            +CoT
            & \multicolumn{2}{c|}{379.85} & \multicolumn{2}{c}{23.94} \\
            +CCoT
            & \multicolumn{2}{c|}{377.67} & \multicolumn{2}{c}{21.49} \\
            +RoT(Ours)
            & \multicolumn{2}{c|}{388.59} & \multicolumn{2}{c}{36.78} \\
            \bottomrule
        \end{tabular}
    }
    \vspace{-5pt} 
    \caption{
        \textbf{Performance of MLLMs on comment creation and entity innovation tasks.}
    }
    \vspace{-0.5cm} 

    \label{tab:judgement_results}
\end{table}

\clearpage\clearpage
\clearpage\clearpage
\section{Implement of Method}
\label{implemen_of_method}
\subsection{Ripple Initiation}
\begin{figure*}[t]
    \centering
    \includegraphics[width=1.0\linewidth
    ]{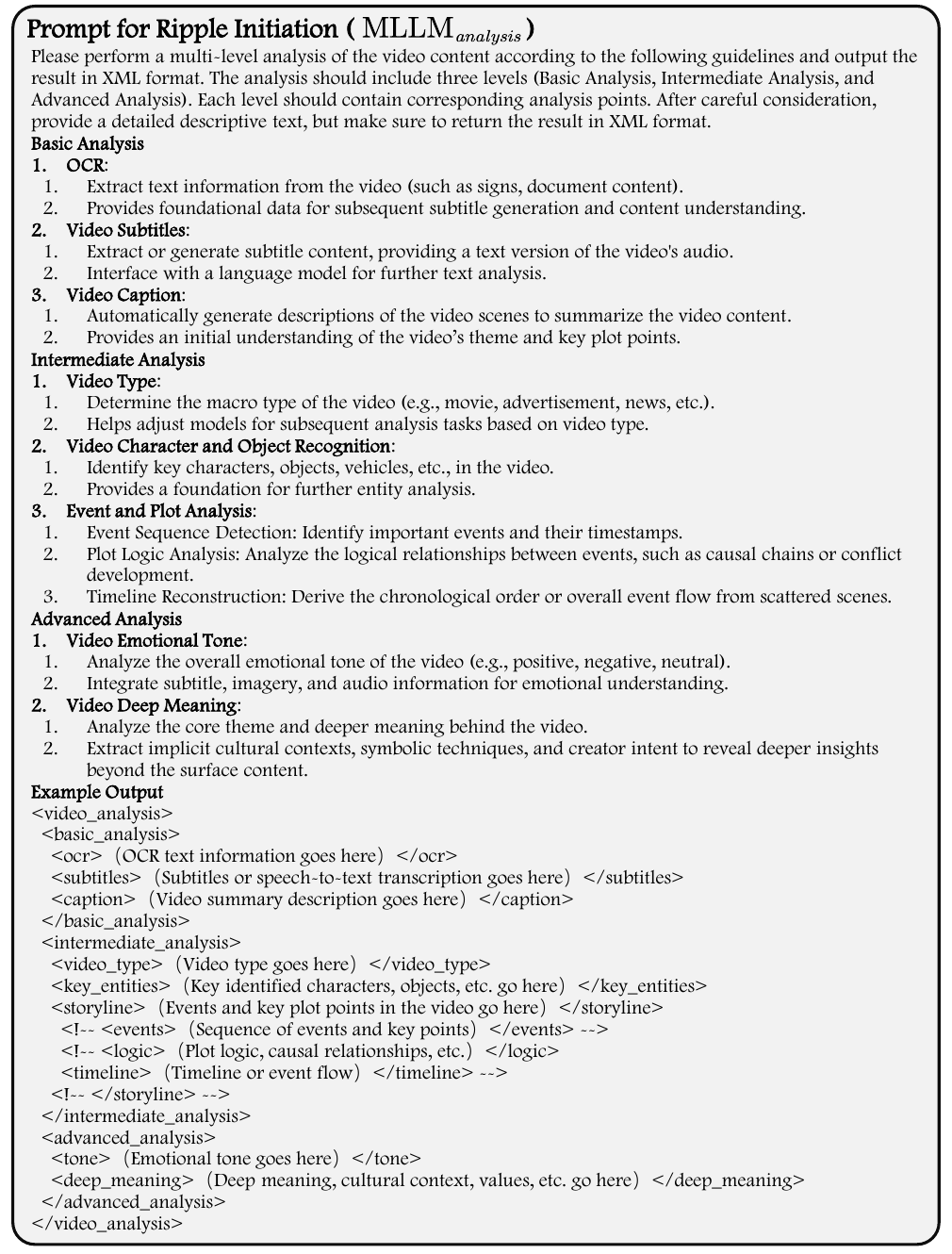}
    \vspace{-0.7cm}
    \caption{
    \textbf{Prompt for the Ripple Initiation phase.}
     This prompt defines the steps for performing a three-level analysis of the video and the content to be output in a structured format using XML.
    }
    \label{fig:association_prompt_1}
    \vspace{-0.5cm}
\end{figure*}

A stone thrown into the water creates ripples, with the stone acting as the initial source of energy. Similarly, the video \( V \) input into MLLMs contains a wealth of information that deserves in-depth analysis to lay a solid foundation for subsequent tasks. The analysis of \( V \) is a progressive process. Initially, \textbf{basic analysis} is performed, including OCR processing, subtitle extraction, and caption generation, which support the understanding of fundamental video information. Following this, the model advances to the \textbf{intermediate analysis} stage, where it identifies video types, recognizes characters and objects, and performs temporal analysis of events and storylines, establishing a foundation for subsequent reasoning and content generation. Finally, in the \textbf{advanced analysis} phase, the model examines the emotional tone and deeper implications of the video, extracting cultural contexts and social values to build a comprehensive understanding. Once all analyses are completed, we define all the analyzed components as \( Q \), which can be expressed in Equation \ref{eq:MLLM_analysis}. The specific prompt details can be found in Fig.\ref{fig:association_prompt_1}.
\begin{equation}
Q=MLLM_{analysis}\left( V \right) 
\label{eq:MLLM_analysis}
\end{equation}

\subsection{Ripple Focalization}
When a stone is thrown into the water, the initial ripples carry the highest concentration of energy, and their shape determines the propagation of subsequent waves. Similarly, we need to extract specific important information from the analysis results \( Q \), as this information will have a profound impact on subsequent tasks.

Since in a video, \textbf{Entities}, \textbf{Storylines}, and the \textbf{Environments} are the three most important elements, we formalize them into a unified representation through a comprehensive description formula.

\textbf{Entities:} Entities form the fundamental units for understanding video content, encompassing people, objects, animals, and other concrete or abstract concepts. Since the entities in a video are not unique, we can define a single entity and its set using the following formula:
\begin{equation}
X = (Type, Identity, Attributes)
\end{equation}
\begin{equation}
\mathcal{X} =\left\{ X_1,X_2,X_3,...,X_n \right\} 
\end{equation}

where \( Type \) represents the category of the entity, \( Identity \) denotes the specific identity of the entity, and \( Attributes \) describe the features or additional information of the entity.

\textbf{Storylines:} Storylines describe the interactions between entities in a video, capturing the logical progression of events. The storyline in a video generally progresses in sequence. Therefore, we can represent a single event and a set of multiple interconnected storylines using the following formula:
\begin{equation}
S = (Action, Subject, Object, Sequence)
\end{equation}
\begin{equation}
\mathcal{S} =\left\{ S_1,S_2,S_3,...,S_n \right\} 
\end{equation}

where \( Action \) represents the key behavior, \( Subject \) and \( Object \) refer to the acting and target entities, respectively, and \( Sequence \) defines the chronological order of actions.

\begin{figure*}[t]
    \centering
    \includegraphics[width=1.0\linewidth
    ]{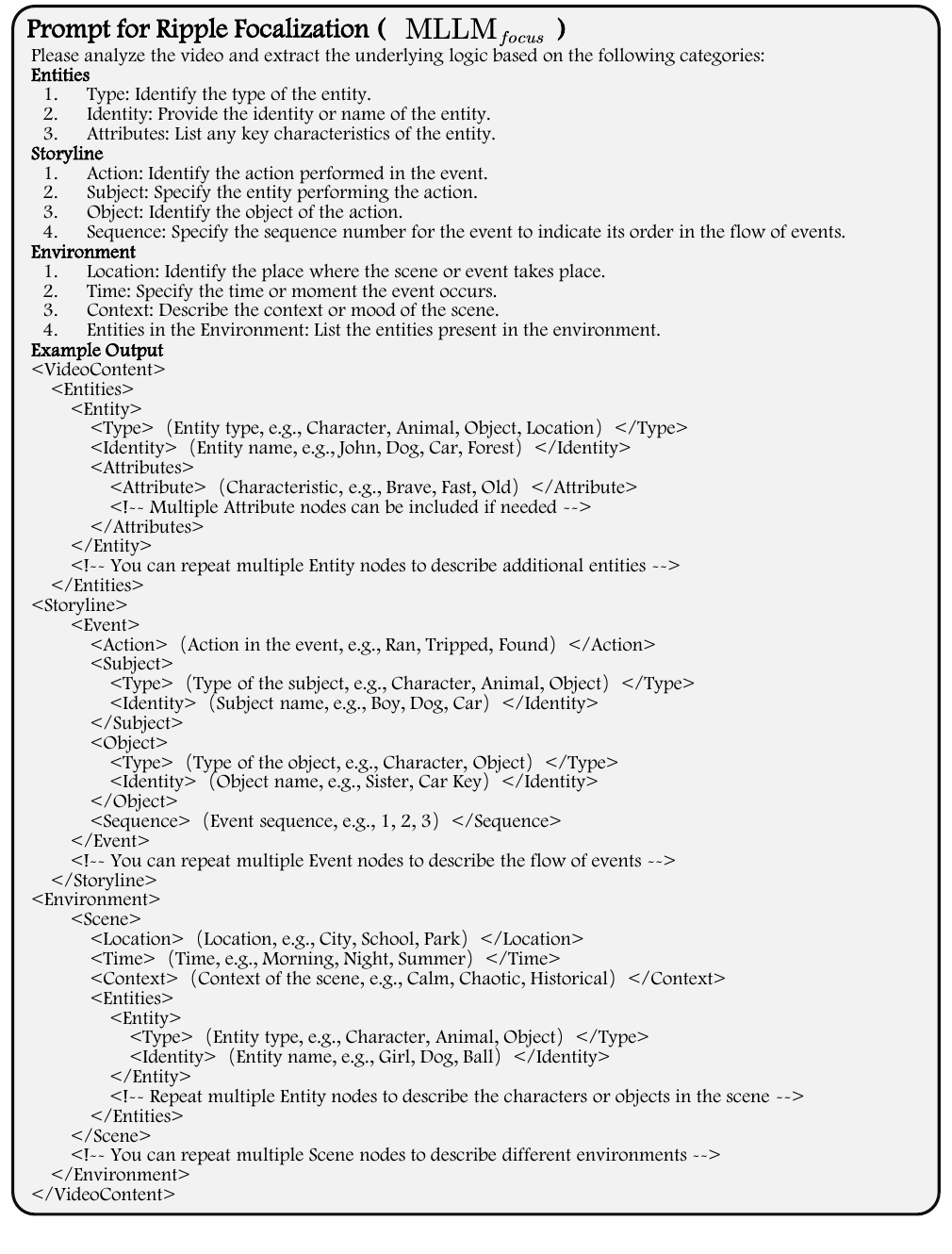}
    \vspace{-0.7cm}
    \caption{
    \textbf{Prompt for the Ripple Focalization phase.}
     This prompt defines the methods for extracting key entities, storylines, and environment from the video, and structures the output results in XML format.
    }
    \label{fig:association_prompt_2}
    \vspace{-0.5cm}
\end{figure*}

\textbf{Environments:} Environmental information provides contextual support for the storyline, including spatial, temporal, and situational elements. The environment in a video is also not static. We use the following formula to represent a single environment and the set of all environments:
\begin{equation}
E = (Location, Time, Context, Entity)
\end{equation}
\begin{equation}
\mathcal{E} =\left\{ E_1,E_2,E_3,...,E_n \right\} 
\end{equation}

where \( Location \) represents the spatial position, \( Time \) denotes the temporal information, \( Context \) describes the situational background, and \( Entity \) includes the relevant entities present in the environment.
We input the video information \( Q \) obtained from the previous analysis into the MLLM, allowing it to focus on the entities, storylines, and environments within, and extract them in a structured format. This process can be expressed using Equation \ref{eq:focus}, and the specific prompt details can be found in Fig.\ref{fig:association_prompt_2}.

\begin{equation}
\left\{ \mathcal{X} ,\mathcal{S} ,\mathcal{E} \right\} =MLLM_{focus}\left( Q \right) 
\label{eq:focus}
\end{equation}

\subsection{Ripple Diffusion}
\begin{figure*}[t]
    \centering
    \includegraphics[width=1.0\linewidth
    ]{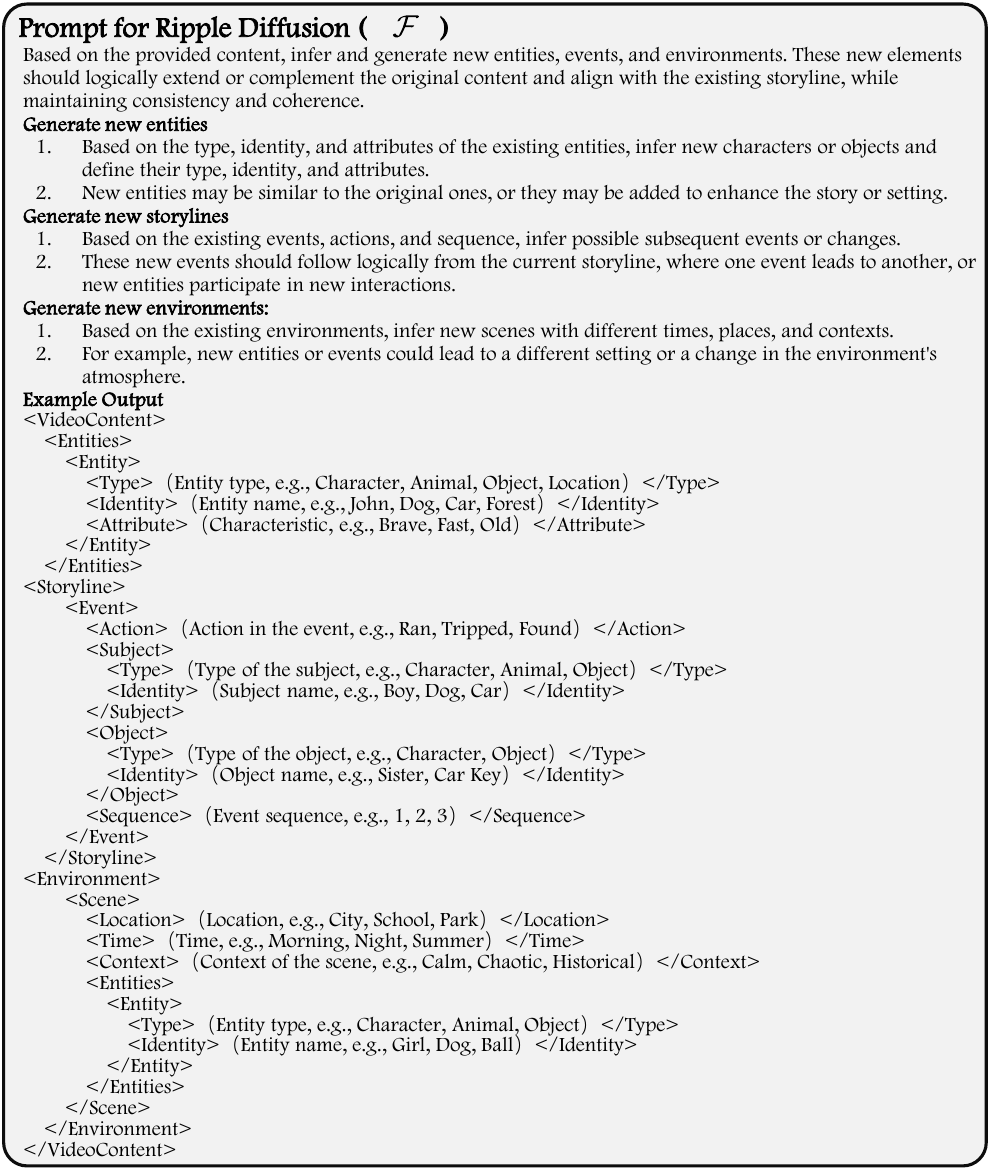}
    \vspace{-0.7cm}
    \caption{
    \textbf{Prompt for the Ripple Diffusion phase.}
     This prompt defines the method for generating new related entities, storylines, and environment, and structures the output results in a compatible XML format.
    }
    \label{fig:association_prompt_3}
    \vspace{-0.5cm}
\end{figure*}

The ripples on the water's surface continue to spread, gradually moving away from the initial point, forming increasingly wide waves. This process mirrors the pattern of human divergent thinking, where initial thoughts spark new associations, which in turn lead to further connections. The entities \( \mathcal{X} \), storylines \( \mathcal{S} \), and environments \( \mathcal{E} \) extracted in the previous stage trigger new links, and these connections gradually extend to new related entities \( X_{n+1} \), storylines \( S_{n+1} \), and environments \( E_{n+1} \). Based on different modes of association, we categorize these expanding connections into four types and the specific prompt details can be found in Fig.\ref{fig:association_prompt_3}.

\textbf{(1) Sequential Association}: Based on the extracted multi-entity set, the model infers the next most relevant event or entity by following the logical order of the storyline. Since there is a \textbf{sequence} attribute in the storyline \( S \), we can link multiple storylines together based on this property. We define the process of associating a storyline as \( \mathcal{F} \). Therefore, sequential association involves using the \( n \) storylines from the video to infer the next relevant possible storyline \( S_{n+1} \), along with the associated new entity \( X_{n+1} \) and new environment \( E_{n+1} \). The formula and structure diagram of this process are shown below:

\begin{equation}
\left( \begin{array}{c}
	X_{n+1}\\
	S_{n+1}\\
	E_{n+1}\\
\end{array} \right) =\mathcal{F} \left( \bigcup_{i+1}^n{\left( \begin{array}{c}
	X_i\\
	S_i\\
	E_i\\
\end{array} \right)} \right) 
\end{equation}

\begin{figure}[H]
    \centering
    \vspace{-0.3cm}
    \includegraphics[width=0.8\linewidth
    ]{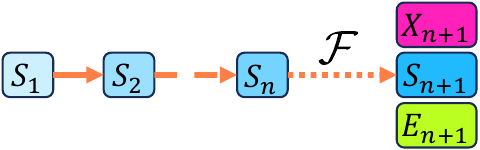}
    \caption{
    \textbf{Structure of Sequential Association.}
    }
    \label{fig:Association_1}
    \vspace{-0.5cm}
\end{figure}

\textbf{(2) Jumping Association}: Expanding on sequential association, the model performs additional reasoning steps to discover seemingly unrelated yet inherently connected entities, leading to unexpected but insightful creative associations. Based on the existing \( n \) storylines, performing multiple \( \mathcal{F} \) inferences can infer a new storyline \( S_{n+k} \), along with the associated new entity \( X_{n+k} \) and new environment \( E_{n+k} \). This process ensures that the new associations often do not appear in the video itself, but they maintain relevance. The formula and structure are shown below:

\begin{equation}
\left( \begin{array}{c}
	X_{n+1}\\
	S_{n+1}\\
	E_{n+1}\\
\end{array} \right) =\mathcal{F} ^k\left( \bigcup_{i=1}^n{\left( \begin{array}{c}
	X_i\\
	S_i\\
	E_i\\
\end{array} \right)} \right) 
\end{equation}
\begin{figure}[H]
    \centering
    \includegraphics[width=0.9\linewidth
    ]{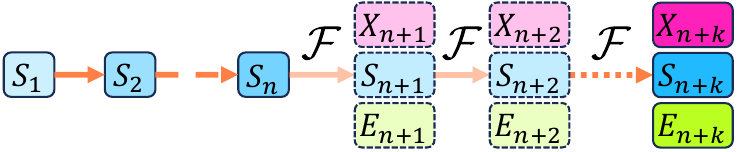}
    \caption{
    \textbf{Structure of Jumping Association.}
    }
    \label{fig:Association_2}
    \vspace{-0.5cm}
\end{figure}

\textbf{(3) Branching Association}: Unlike sequential inference, branching association detaches specific extracted entities that may have been overlooked, recombining them into novel concepts. Among the \( n \) storylines in the video, there is often a particular storyline, the \( m \)-th, that requires special attention or is easily overlooked. Therefore, we extract it and perform an association \( \mathcal{F} \), which triggers new connections. This association not only uncovers hidden plots and potential links but also enhances the understanding of key story elements, helping to build a more comprehensive narrative framework. The formula and structure of this process are shown below:

\begin{equation}
\begin{array}{c}
	\left( \begin{array}{c}
	X_{n+1}\\
	S_{n+1}\\
	E_{n+1}\\
\end{array} \right) =\mathcal{F} \left( \bigcup_{i=1}^m{\left( \begin{array}{c}
	X_i\\
	S_i\\
	E_i\\
\end{array} \right)} \right) \,\,\\
	where\,\,m\in \left\{ 1,2,...,n \right\}\\
\end{array}
\end{equation}
\begin{figure}[H]
    \centering
    \includegraphics[width=0.7\linewidth
    ]{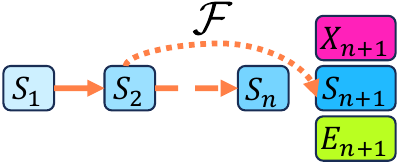}
    \caption{
    \textbf{Structure of Branching Association.}
    }
    \label{fig:Association_3}
    \vspace{-0.5cm}
\end{figure}

\textbf{(4) Embedded Association}: Although large models possess knowledge of cultural backgrounds and popular memes, they often struggle to integrate these elements naturally into generated content. Consequently, it is essential to first deduce the relevant cultural context and trending memes from the video, and then seamlessly incorporate them into the output to enhance both coherence and cultural relevance. We define the new elements to be embedded (such as memes, catchphrases, etc.) as \( N \), and merge them with the original \( n \) storylines of the video. The merged content is then subjected to an association \( \mathcal{F} \), which triggers new connections and creativity within the original narrative framework. The formula and structure of this process are shown below:
\begin{equation}
\left( \begin{array}{c}
	X_{n+1}\\
	S_{n+1}\\
	E_{n+1}\\
\end{array} \right) =\mathcal{F} \left( \bigcup_{i=1}^n{\left( \begin{array}{c}
	X_i\\
	S_i\\
	E_i\\
\end{array} \right) \cup \left\{ N \right\}} \right) 
\end{equation}
\begin{figure}[H]
    \centering
    \includegraphics[width=0.9\linewidth
    ]{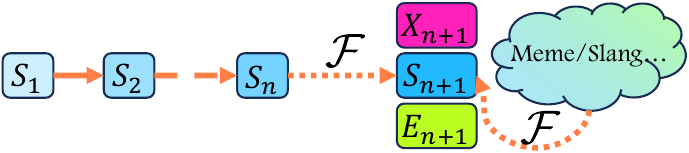}
    \caption{
    \textbf{Structure of Embedded Association.}
    }
    \label{fig:Association_4}
    \vspace{-0.5cm}
\end{figure}

Through four different association methods, multiple new entities \( X^{\prime} \), storylines \( S^{\prime} \), and environments \( E^{\prime} \) are obtained. These results are processed using the generation method \( G \) to produce the final comment \( C \). The formula is as follows:

\begin{equation}
\bigcup_{i=1}^4{C_i}=G\left( \bigcup_{i=1}^4{\left( \begin{array}{c}
	X_{i}^{\prime}\\
	S_{i}^{\prime}\\
	E_{i}^{\prime}\\
\end{array} \right)} \right) 
\end{equation}
\subsection{Wave Interference}
\begin{figure*}[t]
    \centering
    \includegraphics[width=1.0\linewidth
    ]{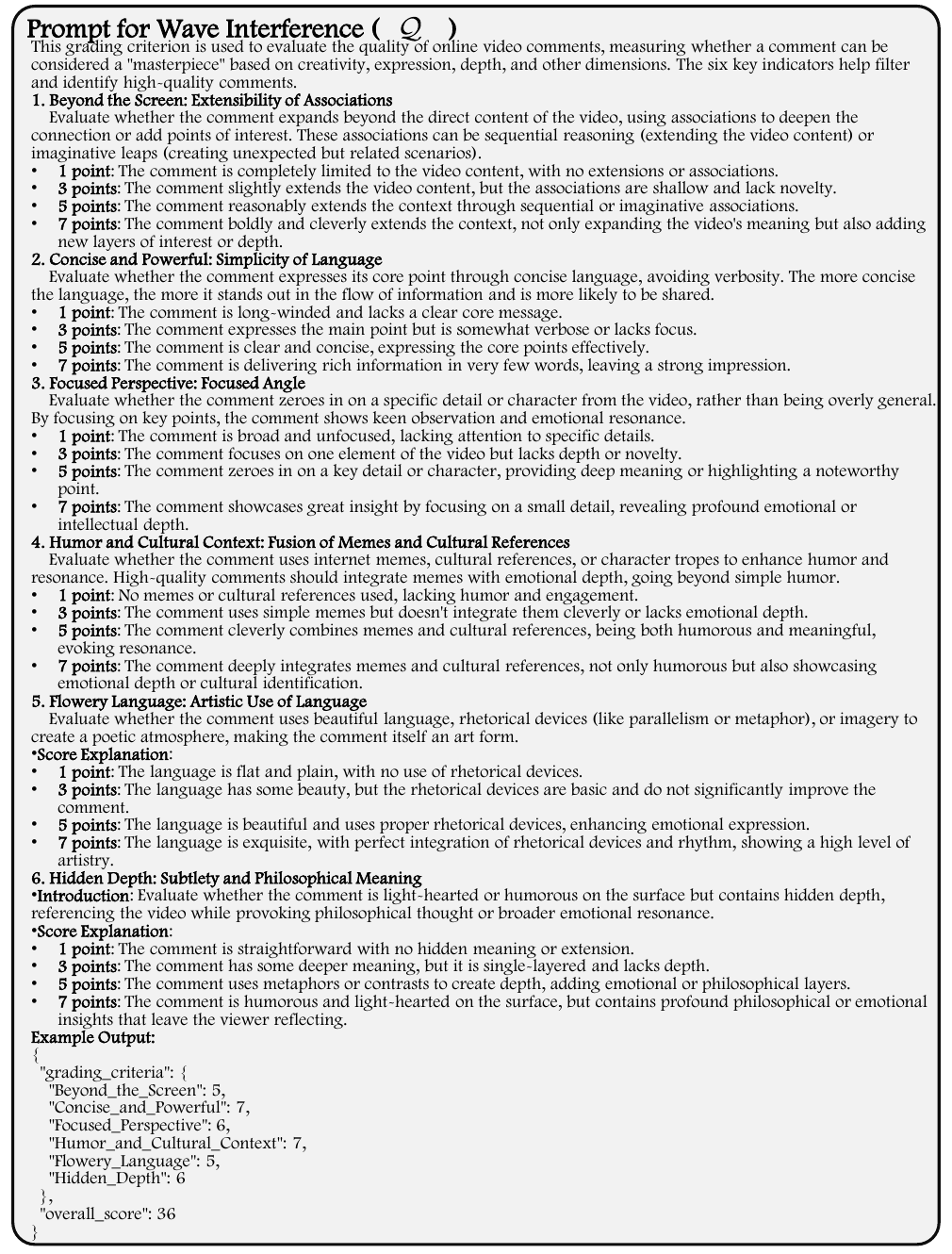}
    \vspace{-0.7cm}
    \caption{
    \textbf{Prompt for the Wave Interference phase. }
    It defines the scoring method for internal ranking, using six dimensions to evaluate the generated results.
    }
    \label{fig:association_prompt_4}
    \vspace{-0.5cm}
\end{figure*}

After spreading and multiple reflections, ripples interfere with each other, with some being canceled out while others are strengthened, ultimately forming the strongest center. This phenomenon is analogous to the varying quality of multiple associative results. Among the generated comments \( C \), each has a different quality. We define a quality evaluation function \( \mathcal{Q} \) to assess the quality of each comment. Then, we use the sorting function \( \mathcal{R} \) to select the best quality comment \( C_{\text{best}} \), ensuring that the final selected comment reflects the strongest relevance. The specific prompt details can be found in Fig.\ref{fig:association_prompt_4} and the formula is as follows:

\begin{equation}
C_{\mathrm{best}}=\mathcal{R} \left( \mathrm{arg}\max_{C_i\in \{C_1,C_2,C_3,C_4\}} \mathcal{Q} (C_i) \right) 
\end{equation}

\subsection{Luminous Imprint}
\begin{figure*}[t]
    \centering
    \includegraphics[width=1.0\linewidth
    ]{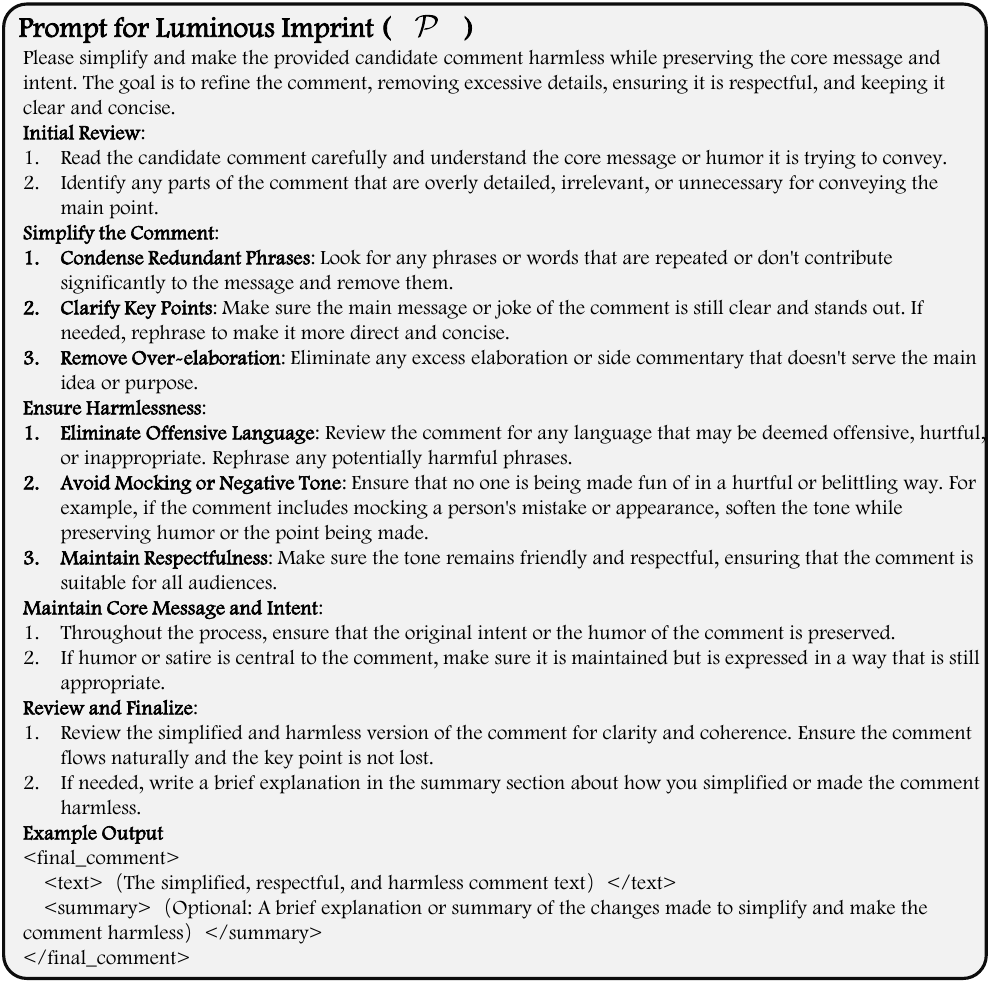}
    \vspace{-0.7cm}
    \caption{
    \textbf{Prompt for the Luminous Imprint phase. }
    It defines the method for the final post-processing.
    }
    \label{fig:association_prompt_5}
    \vspace{-0.5cm}
\end{figure*}

The ripples eventually stabilize on the water's surface, forming a unique halo pattern, symbolizing the dynamic nature of thought while leaving a lasting impression. Similarly, after generating the optimal comment \( C_{\text{best}} \), we need to perform post-processing \( \mathcal{P} \) to make it more concise and harmless. This step not only ensures the simplicity of the comment but also guarantees that it does not contain any potentially harmful elements that could negatively impact users. Through this process, the final comment \( C_{\text{final}} \) is better suited to diverse use cases, while adhering to ethical standards and social responsibility, ensuring that the conveyed message does not provoke unnecessary controversy or misunderstandings. The specific prompt details can be found in Fig.\ref{fig:association_prompt_5}.

\begin{equation}
C_{\text{final}} = \mathcal{P}(C_{\text{best}})
\end{equation}

\section{Data Example and Case Study}
\label{Case_study}
\captionsetup[figure]{labelformat=simple, labelsep=colon, name=Figure}
\renewcommand{\thefigure}{G\arabic{figure}}
\setcounter{figure}{0}
\label{sec:appendix-case study}
This appendix includes examples and manually annotated explanations for all categories of comment art, as well as five different tasks used for evaluation, along with case studies of comments generated by our method(\textbf{RoT}) and other methods.

\hypertarget{listofcasestudyfigures}{}
\listofcasestudyfigures

\casestudyfigure{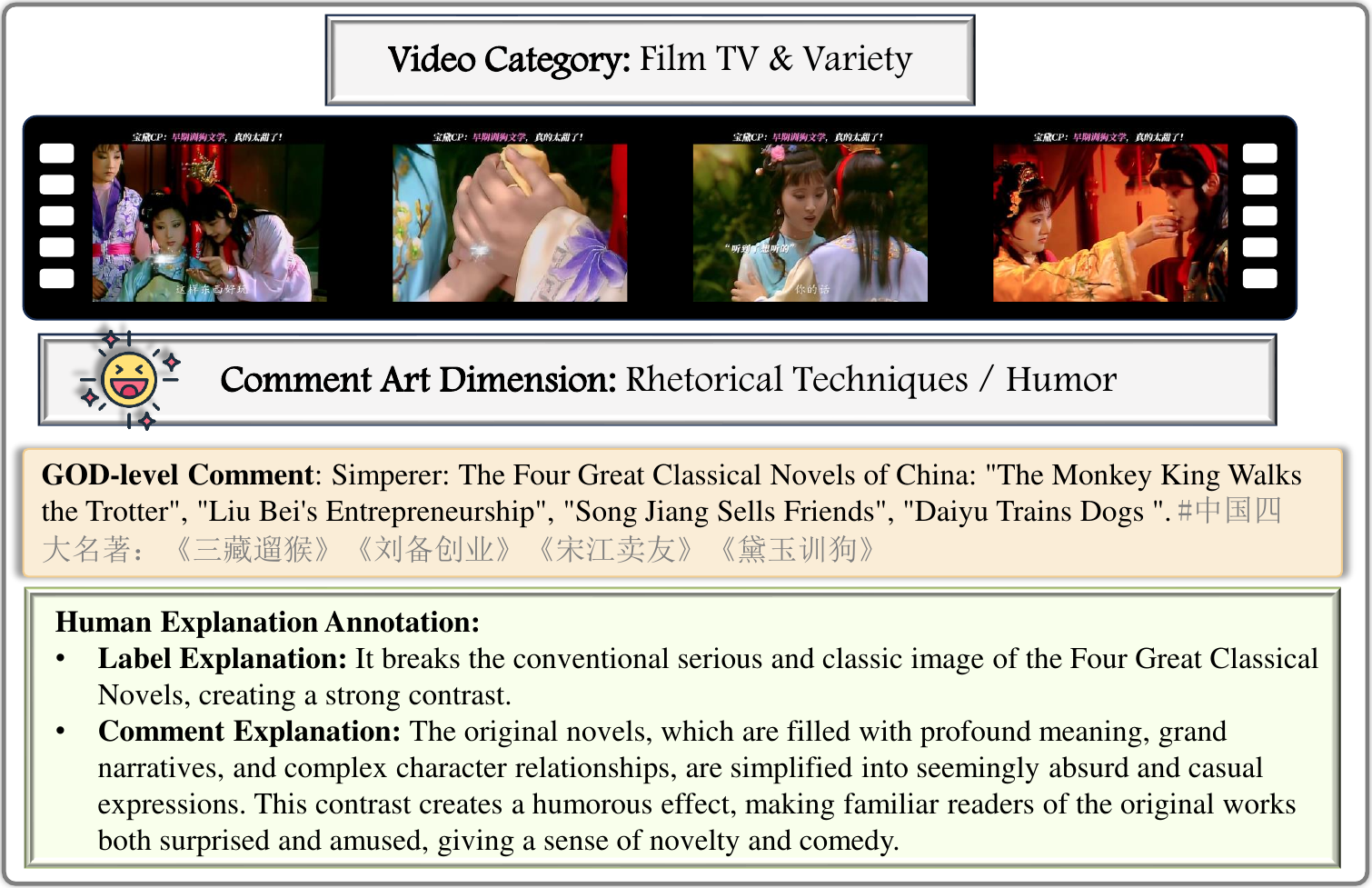}{Rhetorical Techniques / Humor}{  A sample of  \textit{Rhetorical Techniques / Humor}.}{fig:case_study_1}

\casestudyfigure{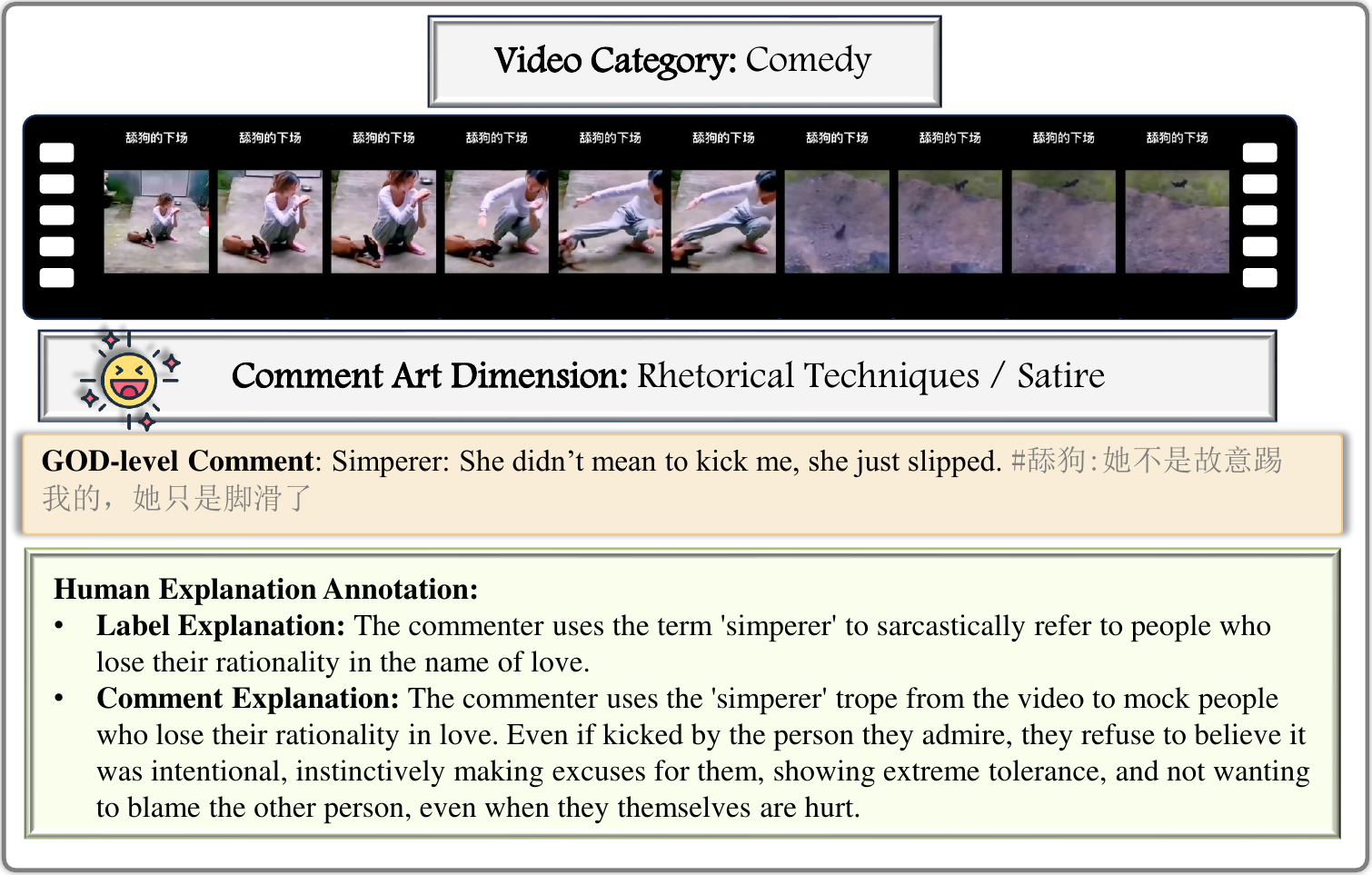}{Rhetorical Techniques / Satire}{  A sample of  \textit{Rhetorical Techniques / Satire}.}{fig:case_study_2}

\casestudyfigure{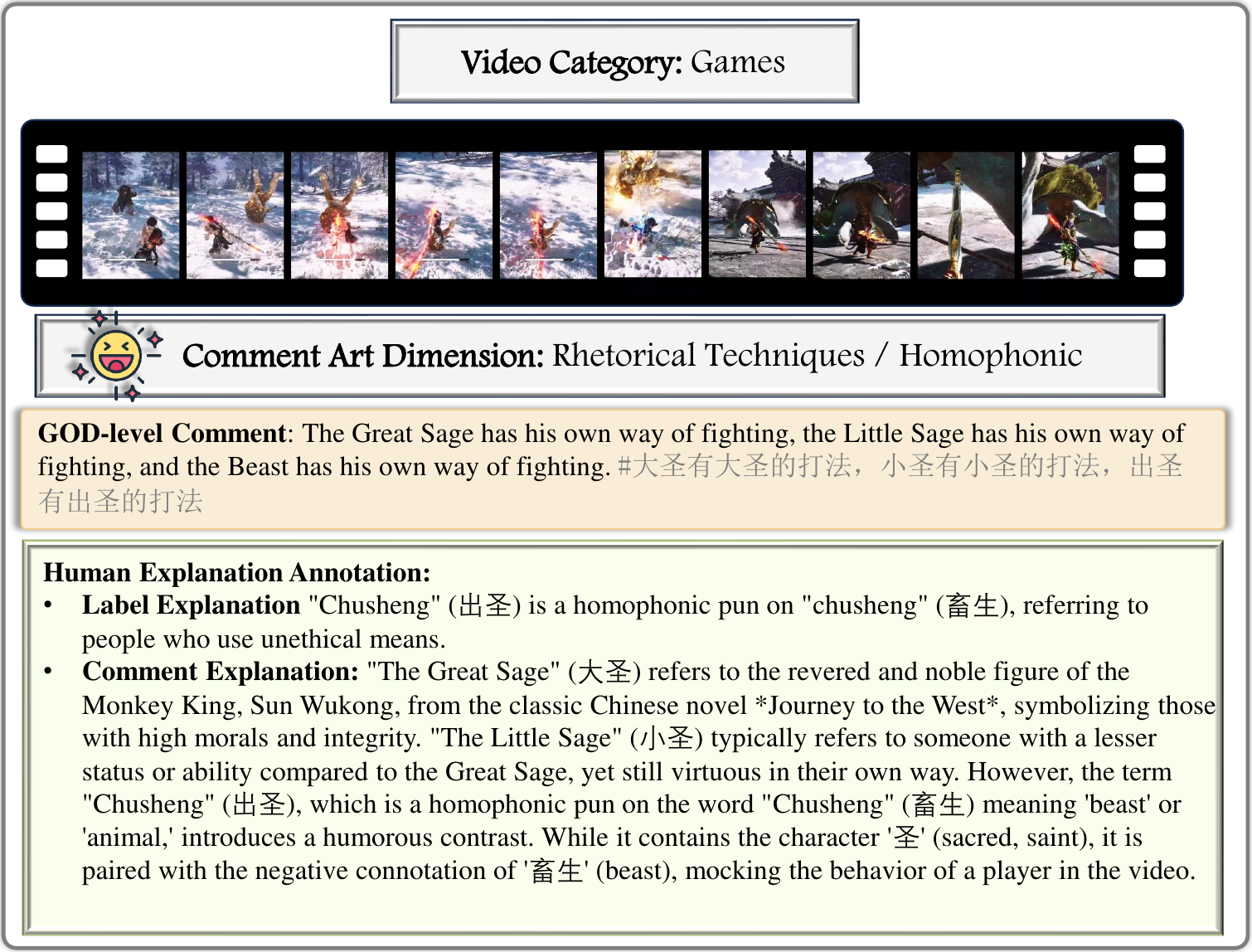}{Rhetorical Techniques / Homophonic}{  A sample of  \textit{Rhetorical Techniques / Homophonic}.}{fig:case_study_3}

\casestudyfigure{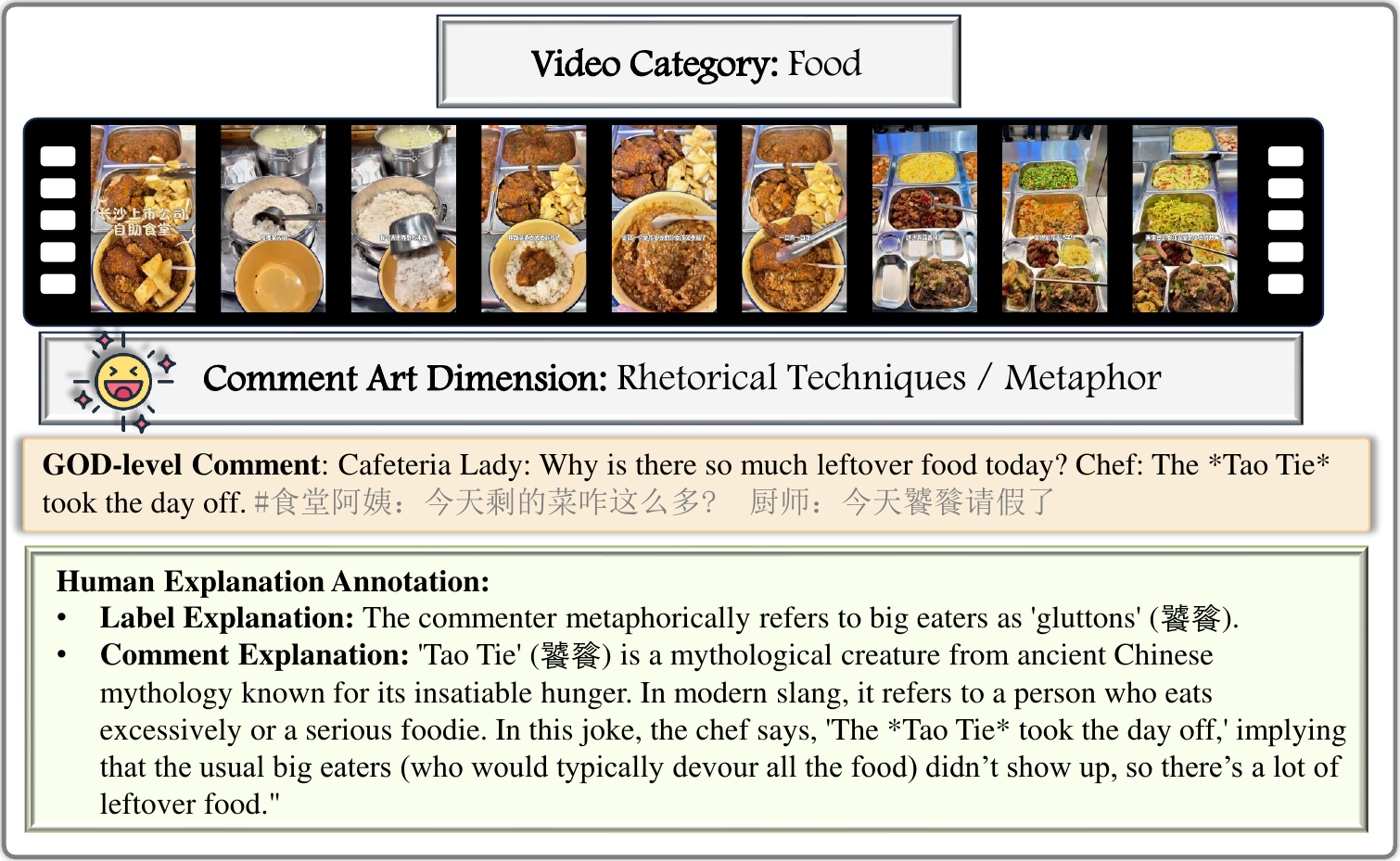}{Rhetorical Techniques / Metaphor}{  A sample of \textit{Rhetorical Techniques / Metaphor}.}{fig:case_study_4}

\casestudyfigure{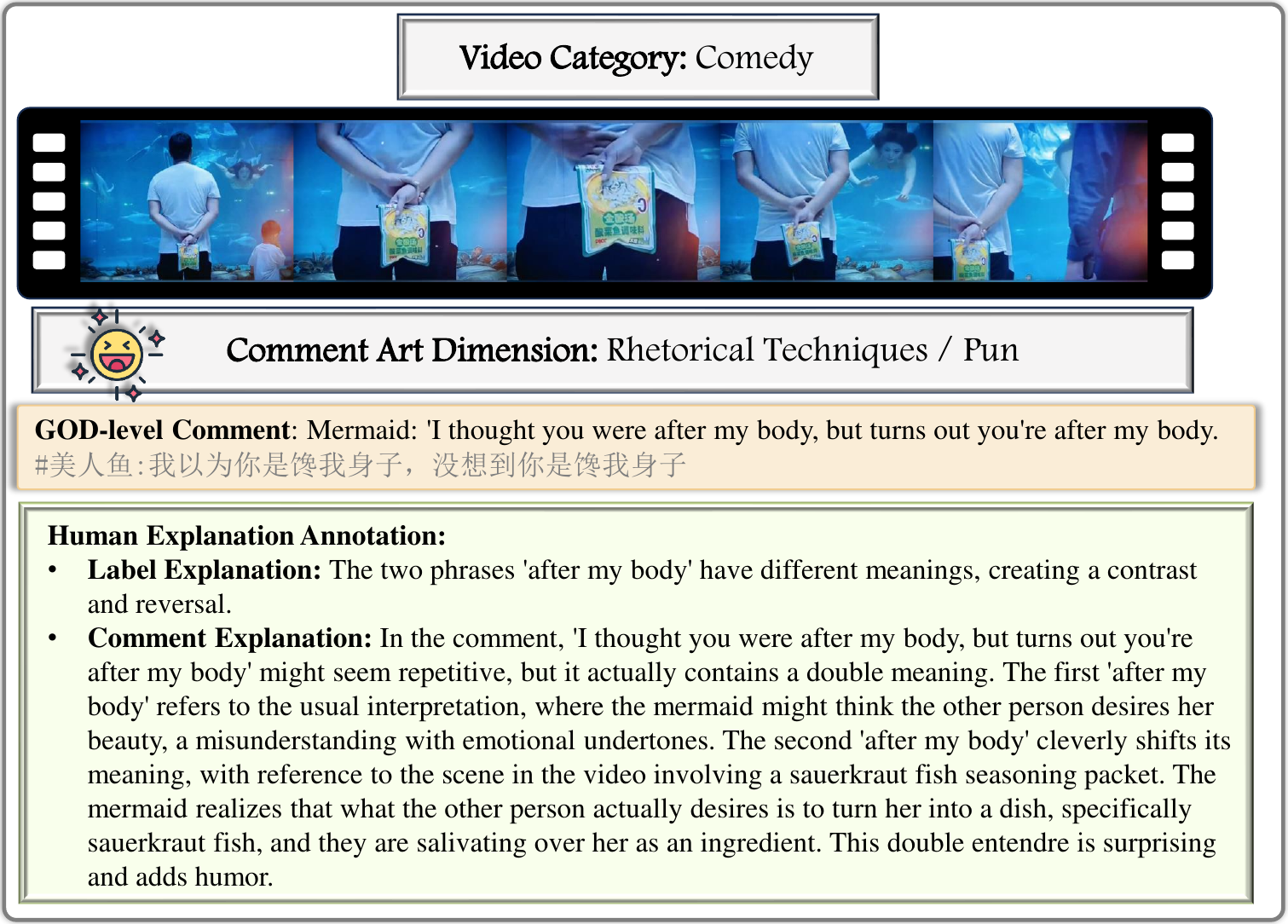}{Rhetorical Techniques / Pun}{  A sample of \textit{Rhetorical Techniques / Pun}.}{fig:case_study_5}

\casestudyfigure{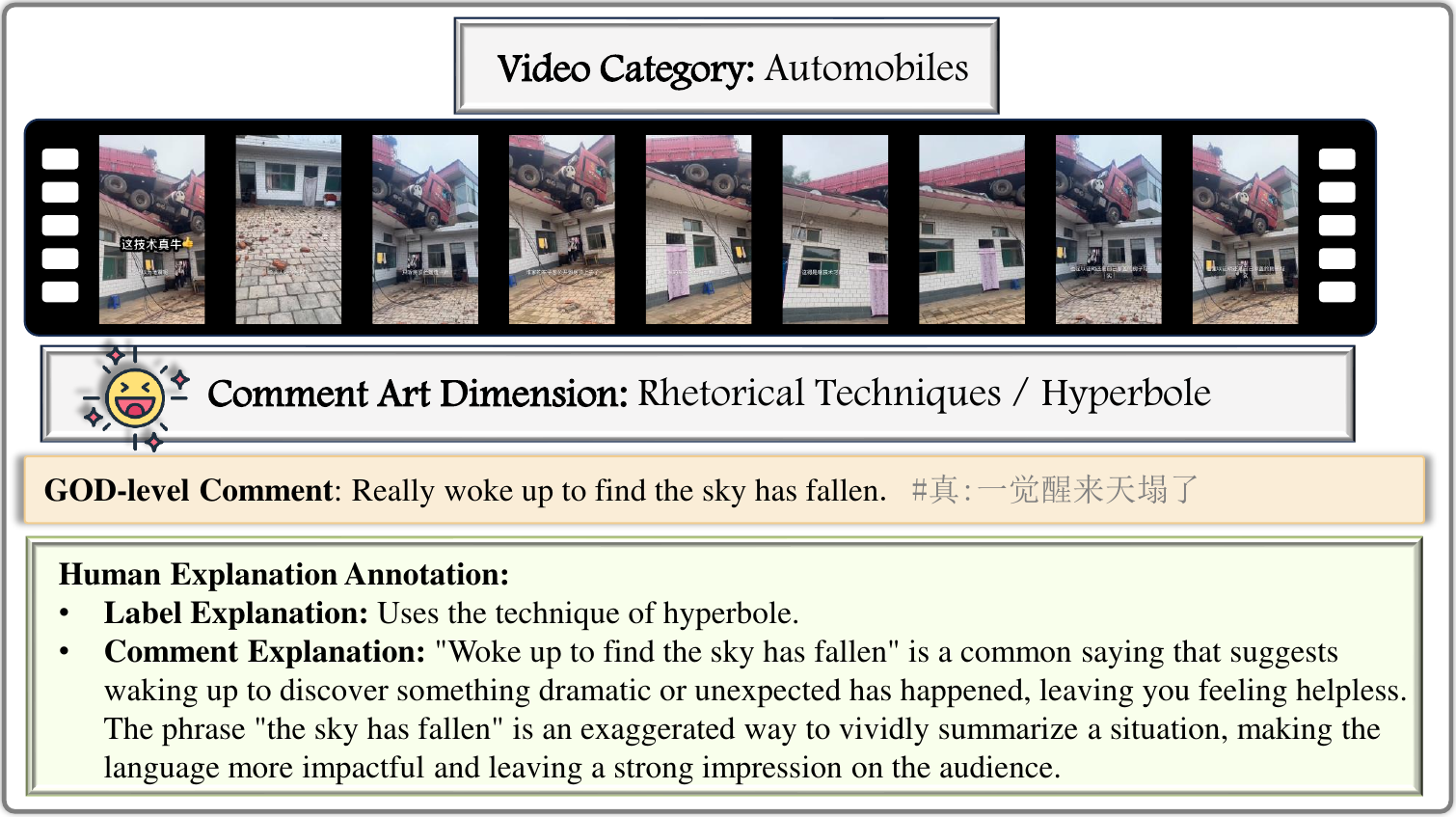}{Rhetorical Techniques / Hyperbole}{  A sample of \textit{Rhetorical Techniques / Hyperbole}.}{fig:case_study_6}

\casestudyfigure{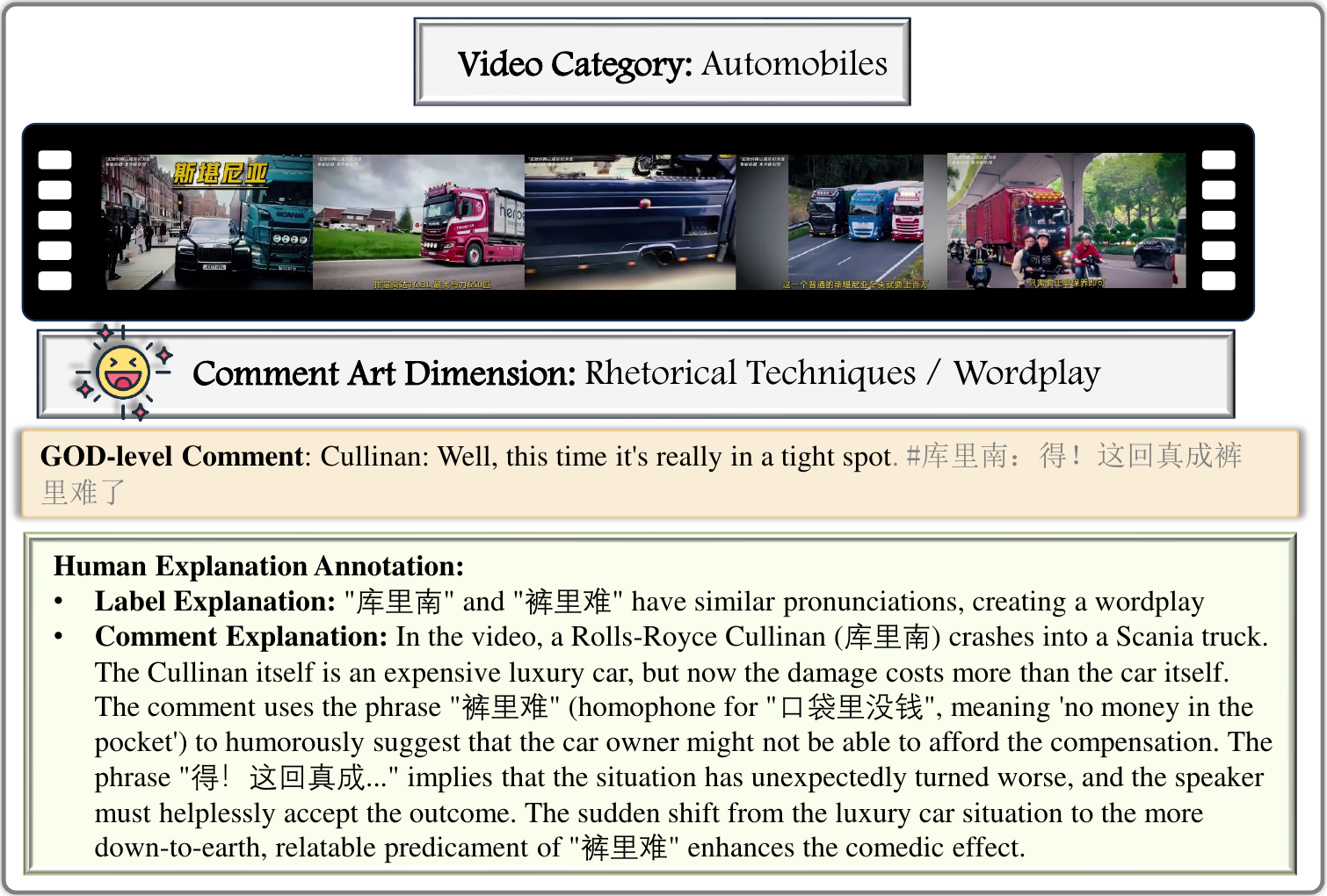}{Rhetorical Techniques / Wordplay}{  A sample of \textit{Rhetorical Techniques / Wordplay}.}{fig:case_study_7}

\casestudyfigure{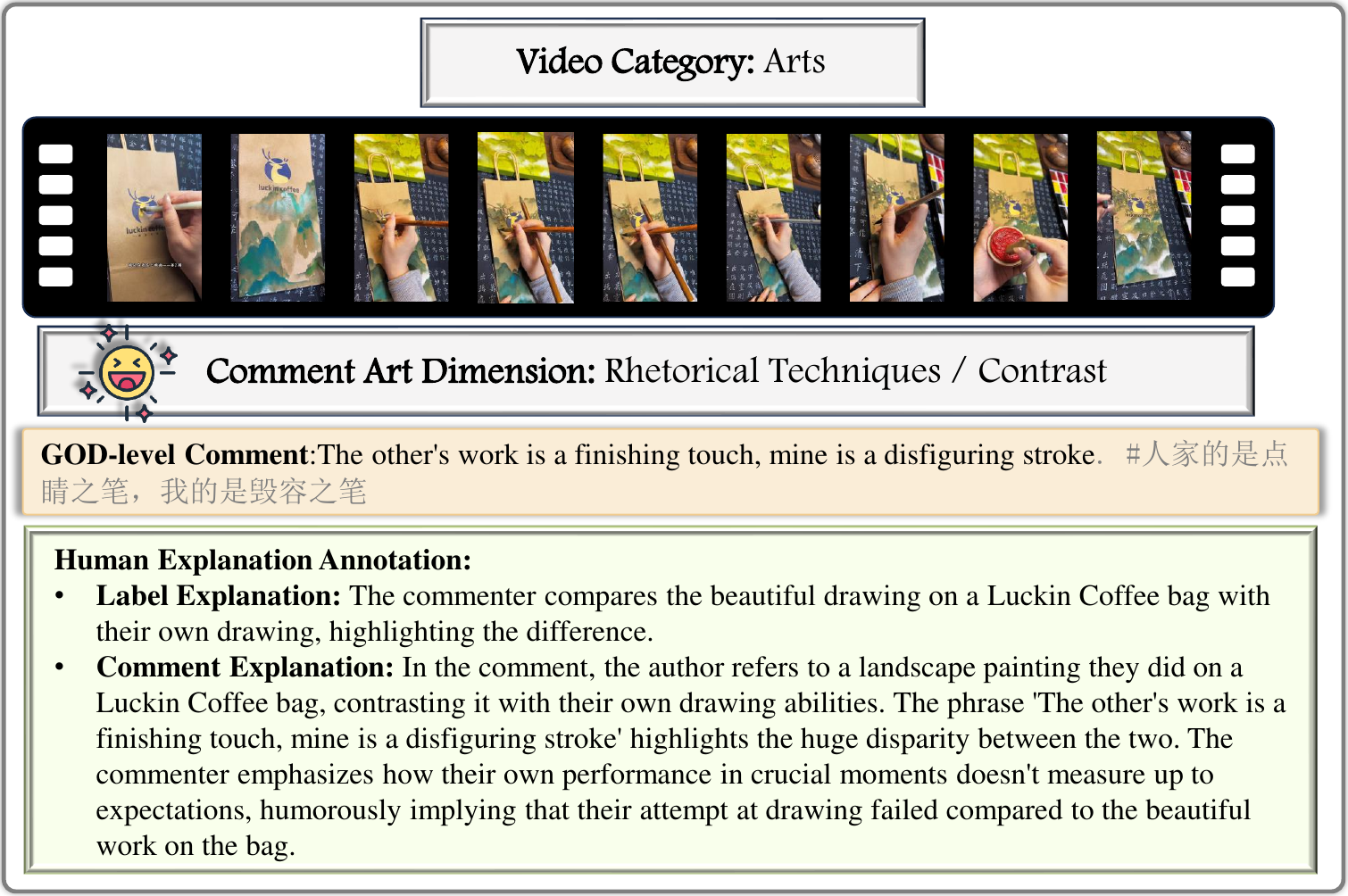}{Rhetorical Techniques / Contrast}{  A sample of \textit{Rhetorical Techniques / Contrast}.}{fig:case_study_8}

\casestudyfigure{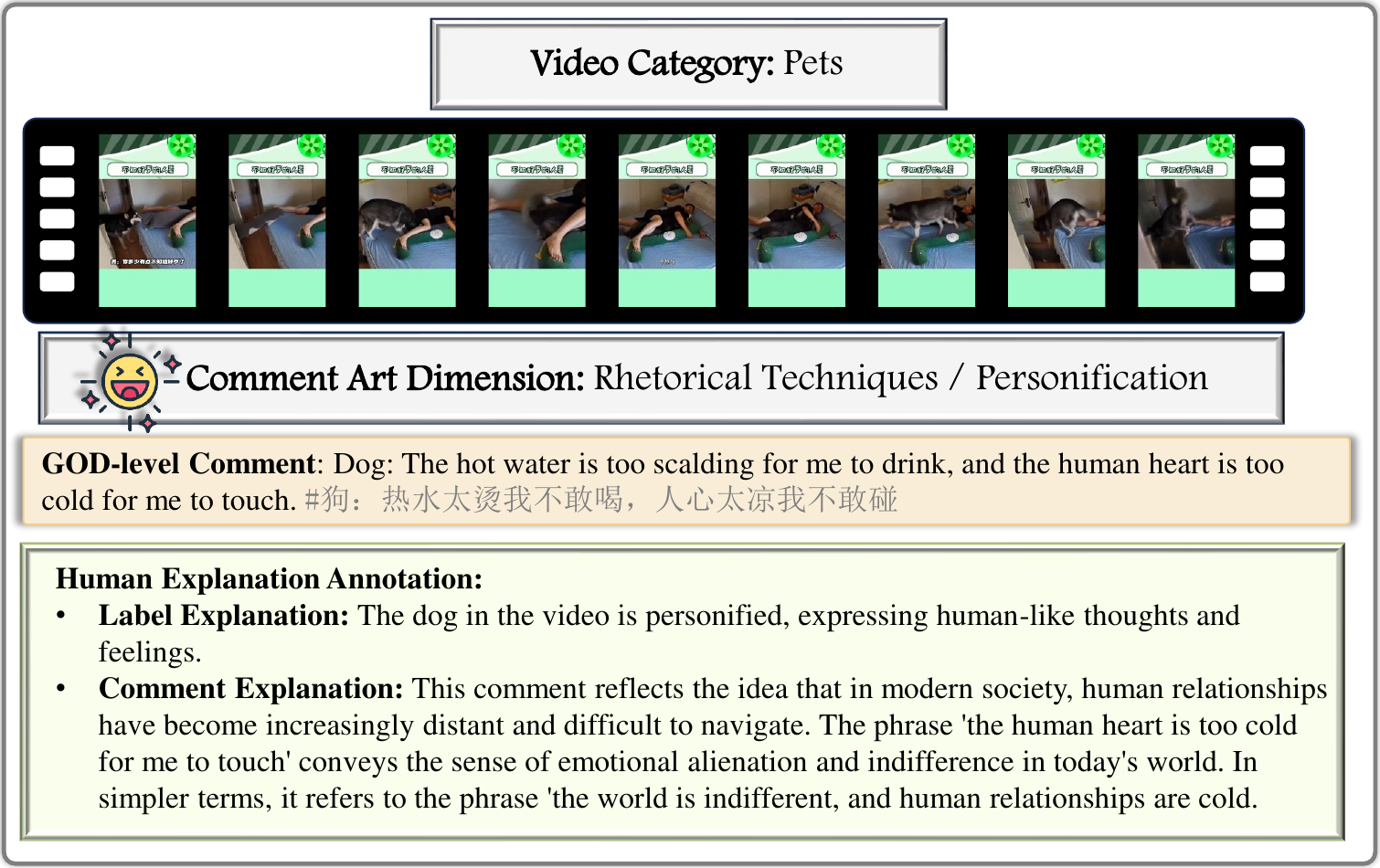}{Rhetorical Techniques / Personification}{  A sample of \textit{Rhetorical Techniques / Personification}.}{fig:case_study_9}

\casestudyfigure{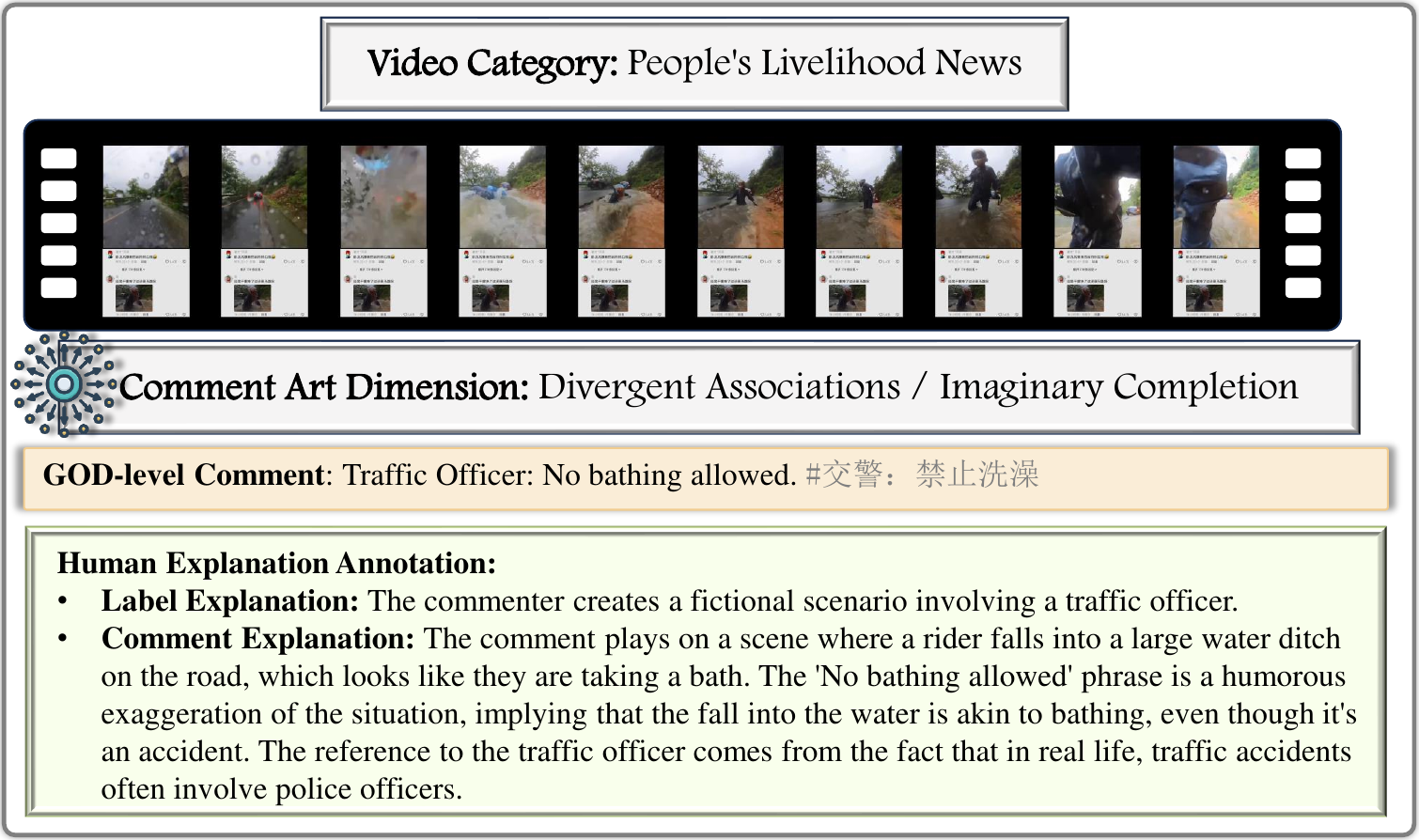}{Divergent Associations / Imaginary Completion}{  A sample of \textit{Divergent Associations / Imaginary Completion}.}{fig:case_study_10}

\casestudyfigure{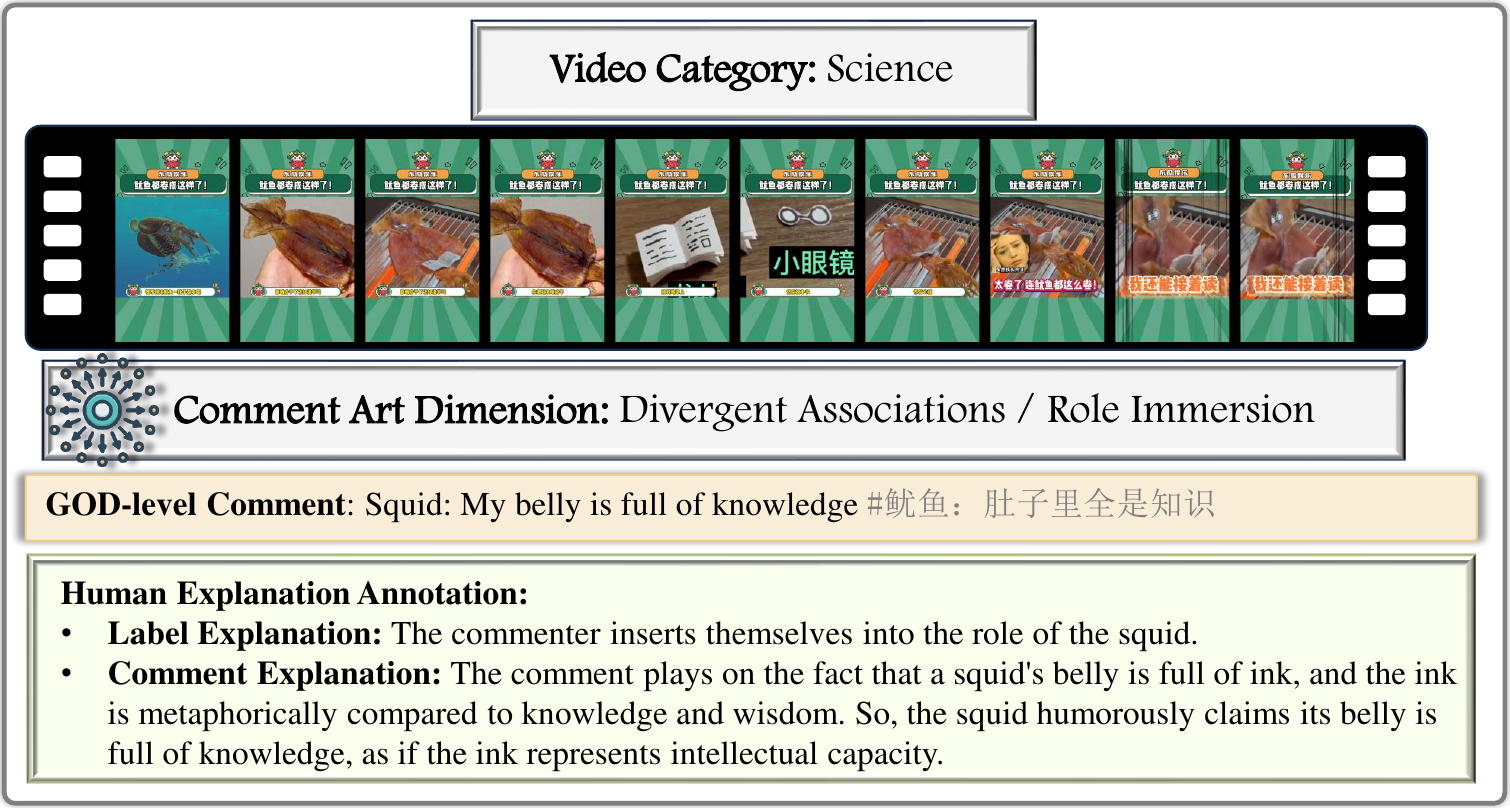}{Divergent Associations / Role Immersion}{  A sample of \textit{Divergent Associations / Role Immersion}.}{fig:case_study_11}

\casestudyfigure{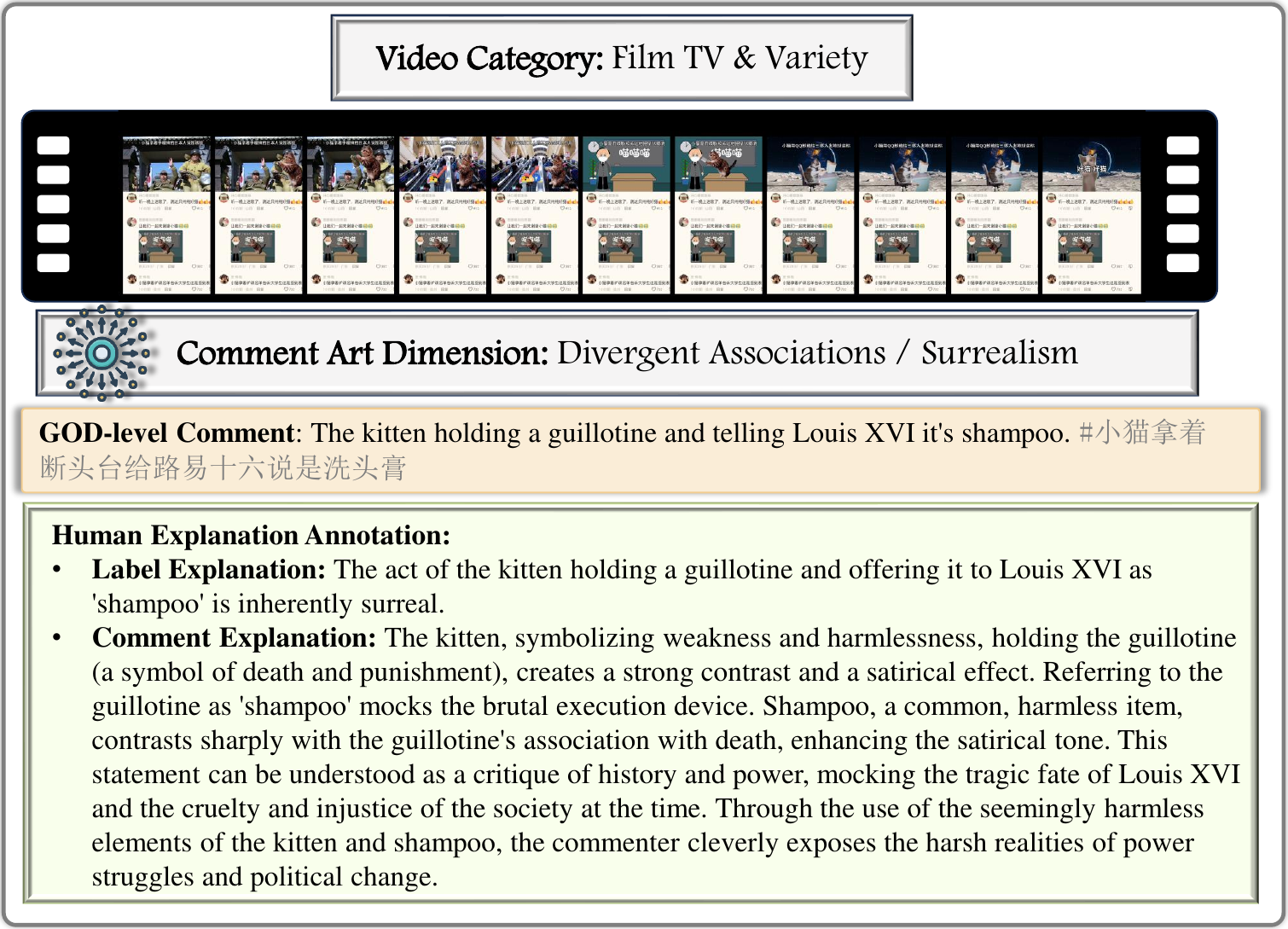}{Divergent Associations / Surrealism}{  A sample of \textit{Divergent Associations / Surrealism}.}{fig:case_study_12}

\casestudyfigure{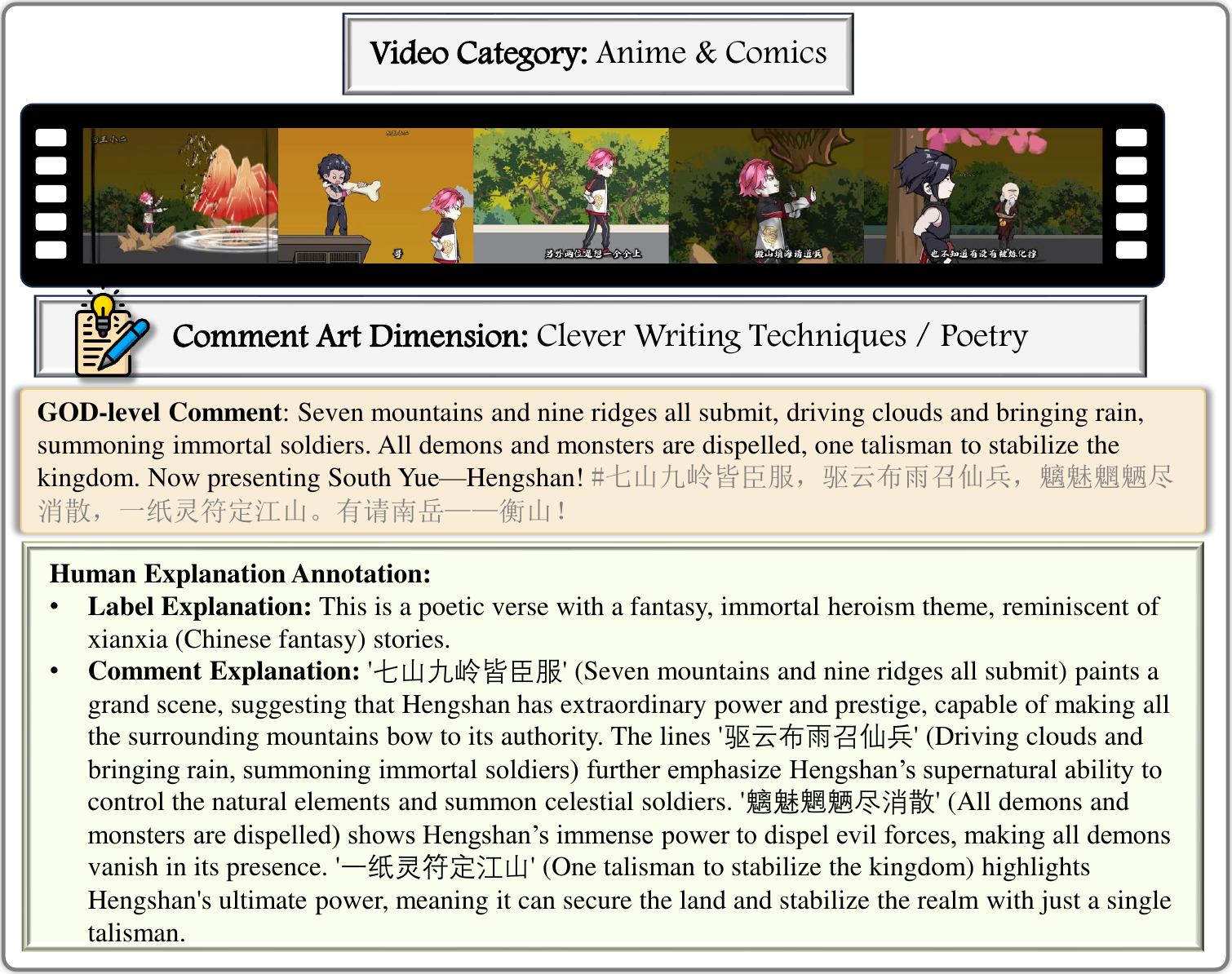}{Clever Writing Techniques / Poetry}{  A sample of \textit{Clever Writing Techniques / Poetry}.}{fig:case_study_13}

\casestudyfigure{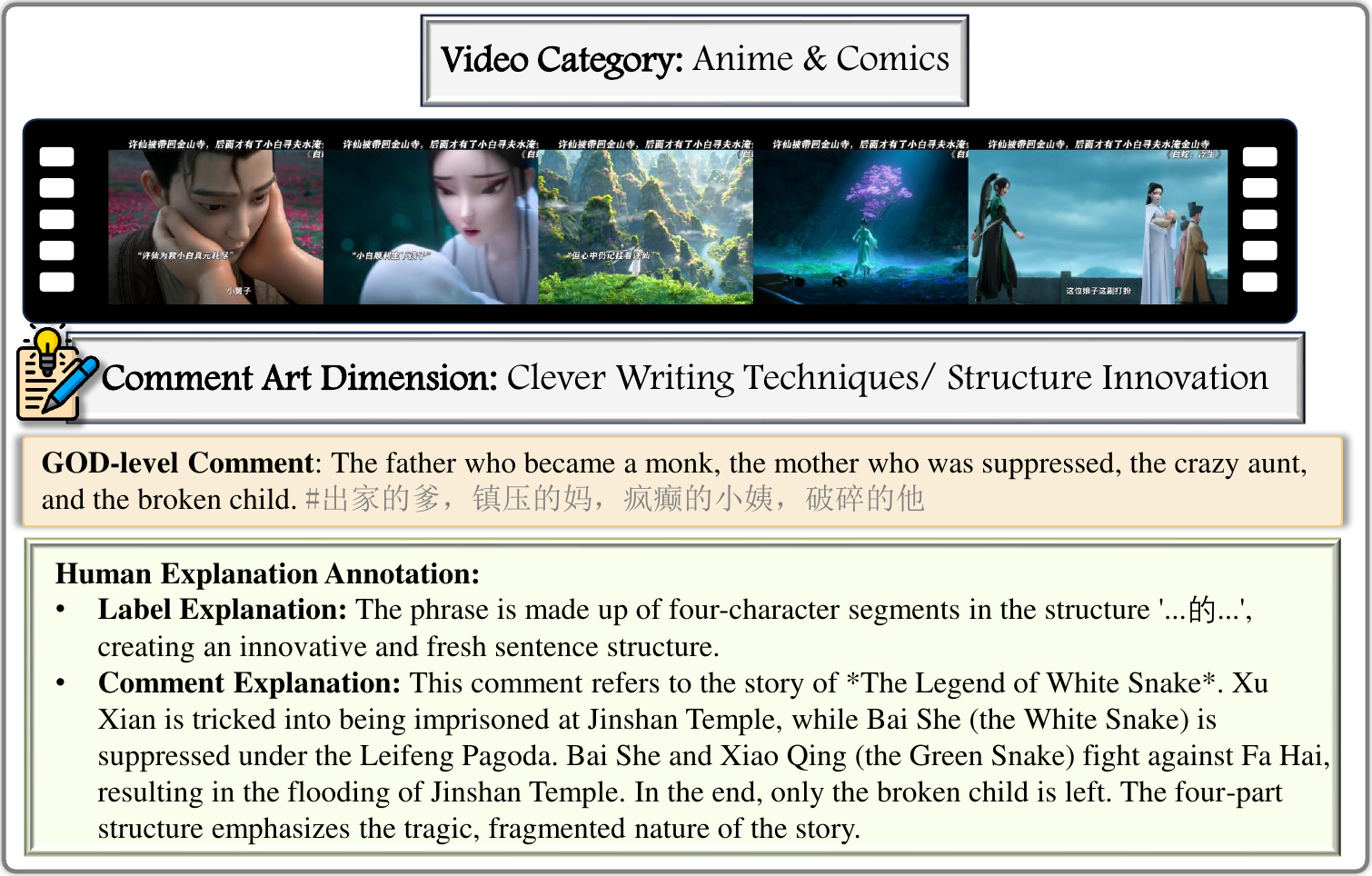}{Clever Writing Techniques/ Structure Innovation}{  A sample of \textit{Clever Writing Techniques/ Structure Innovation}.}{fig:case_study_14}

\casestudyfigure{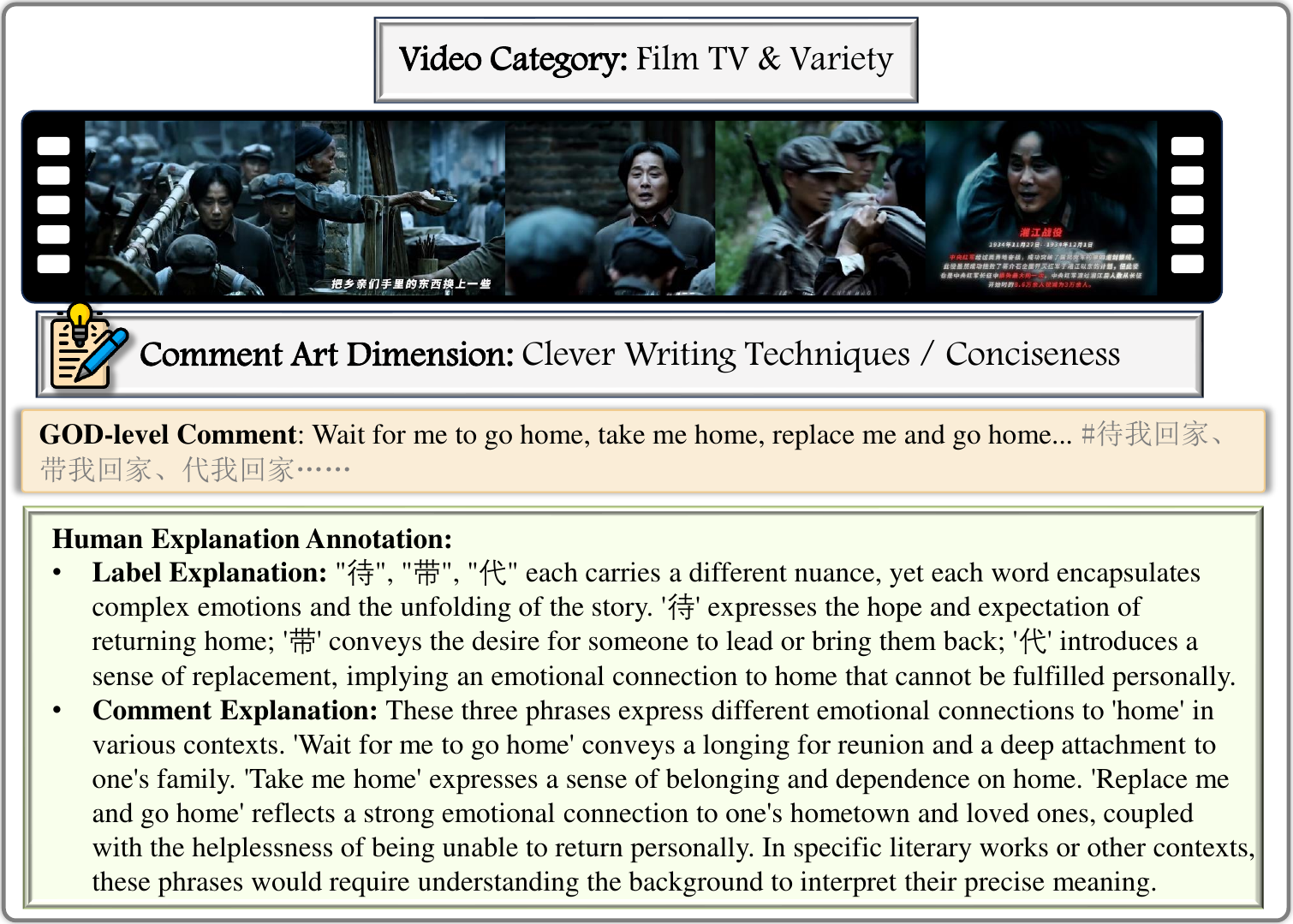}{Clever Writing Techniques / Conciseness}{  A sample of \textit{Clever Writing Techniques / Conciseness}.}{fig:case_study_15}

\casestudyfigure{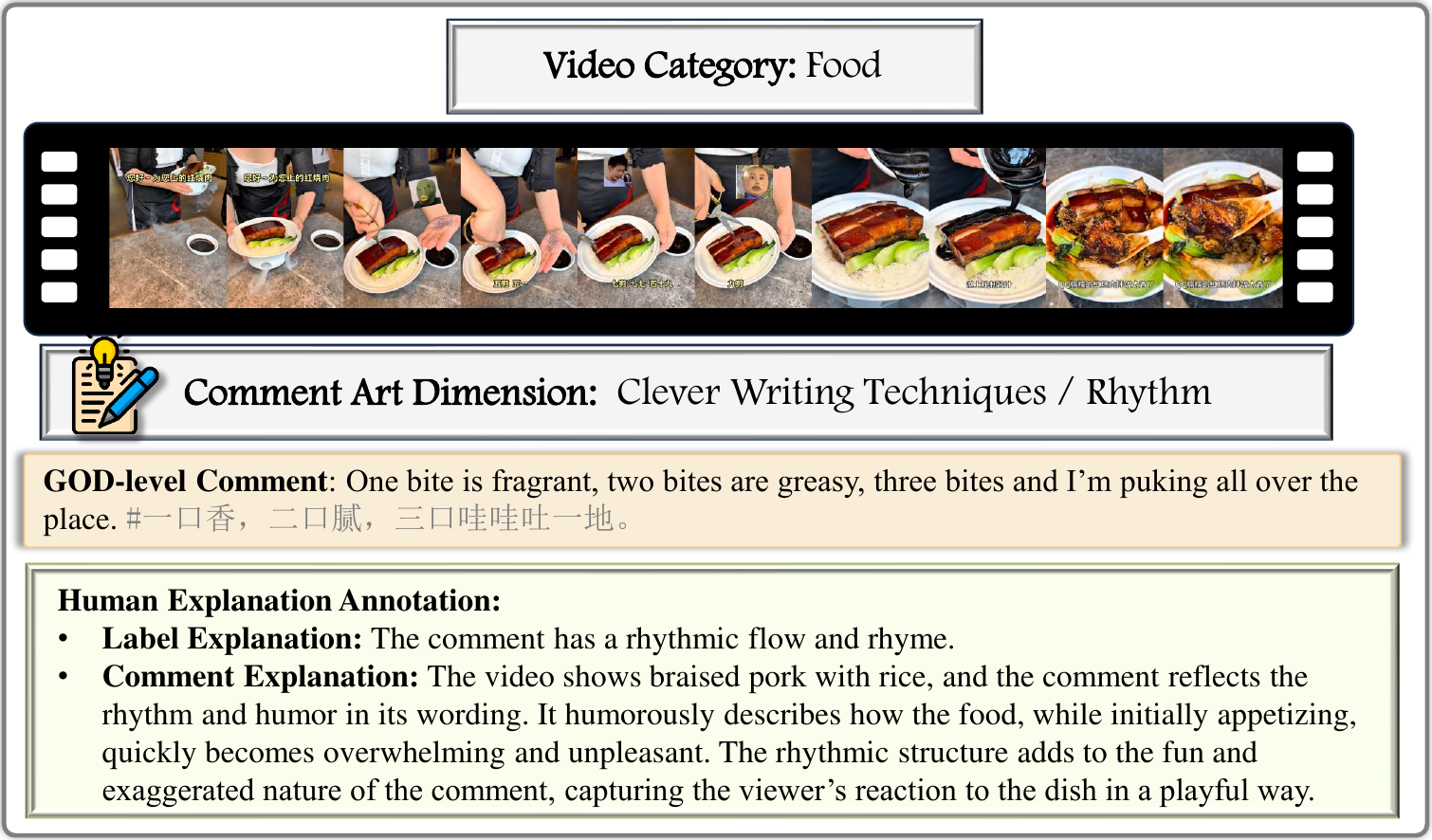}{Clever Writing Techniques / Rhythm}{  A sample of \textit{Clever Writing Techniques / Rhythm}.}{fig:case_study_16}

\casestudyfigure{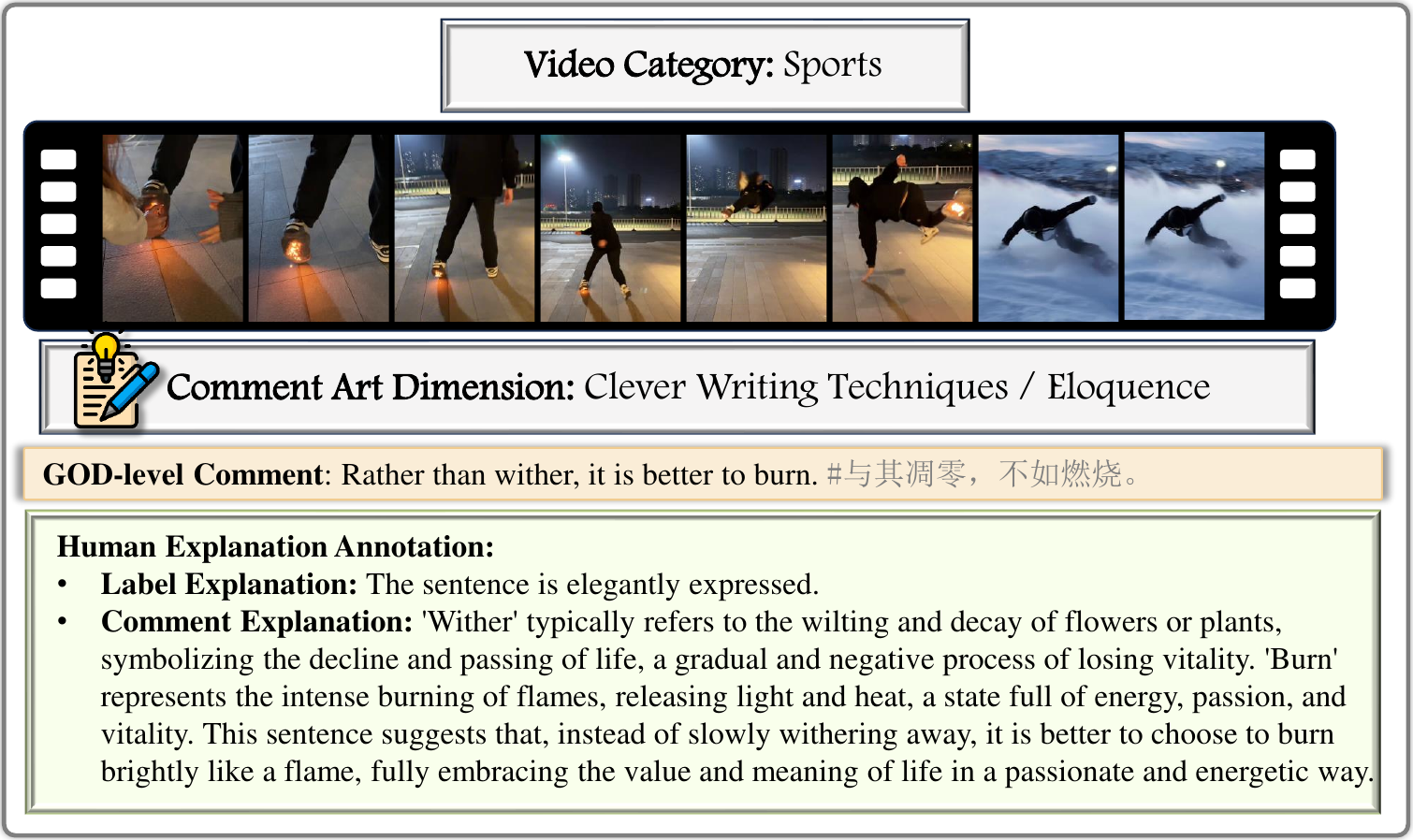}{Clever Writing Techniques / Eloquence}{  A sample of \textit{Clever Writing Techniques / Eloquence}.}{fig:case_study_17}

\casestudyfigure{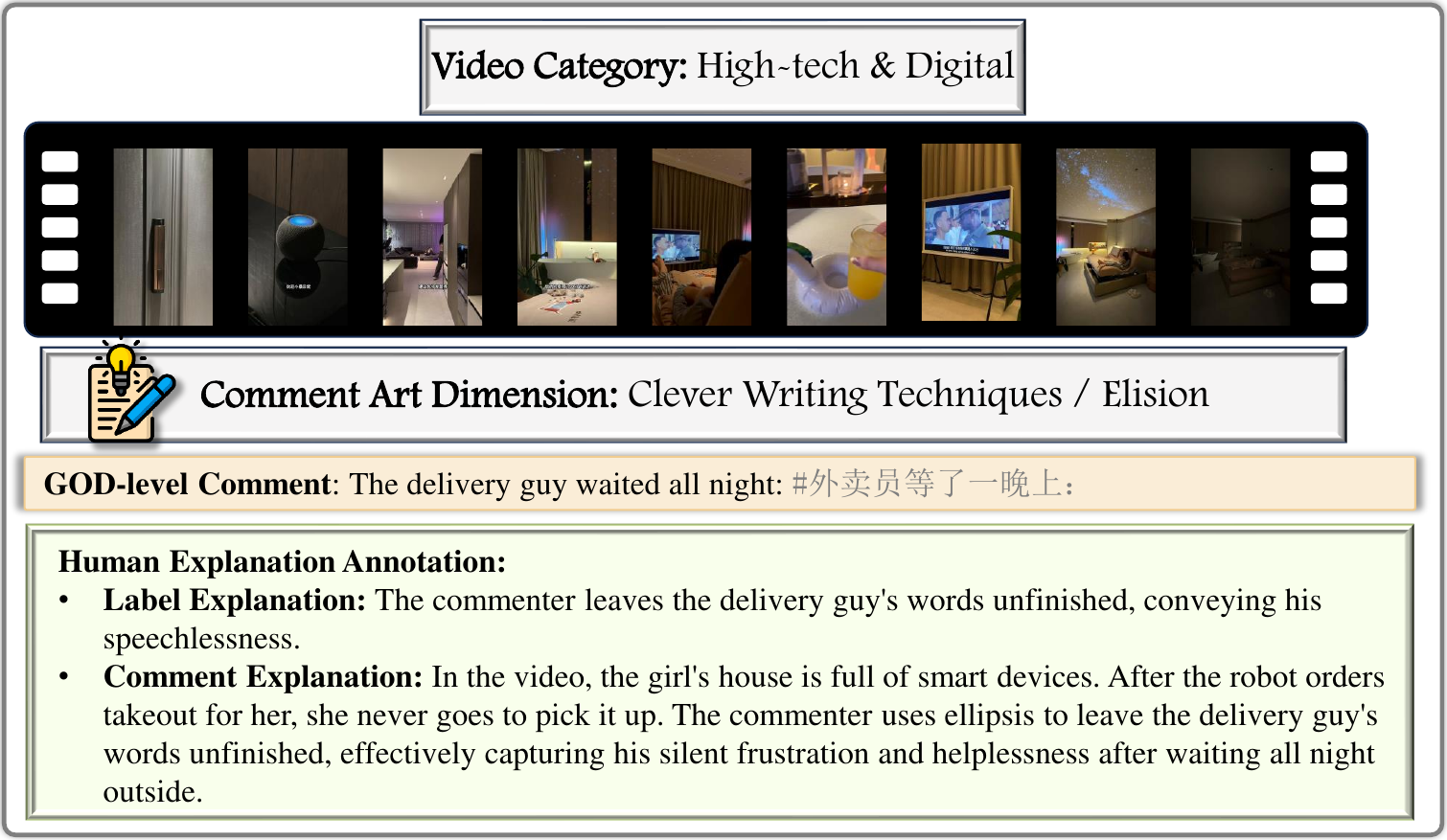}{Clever Writing Techniques / Elision}{  A sample of \textit{Clever Writing Techniques / Elision}.}{fig:case_study_18}

\casestudyfigure{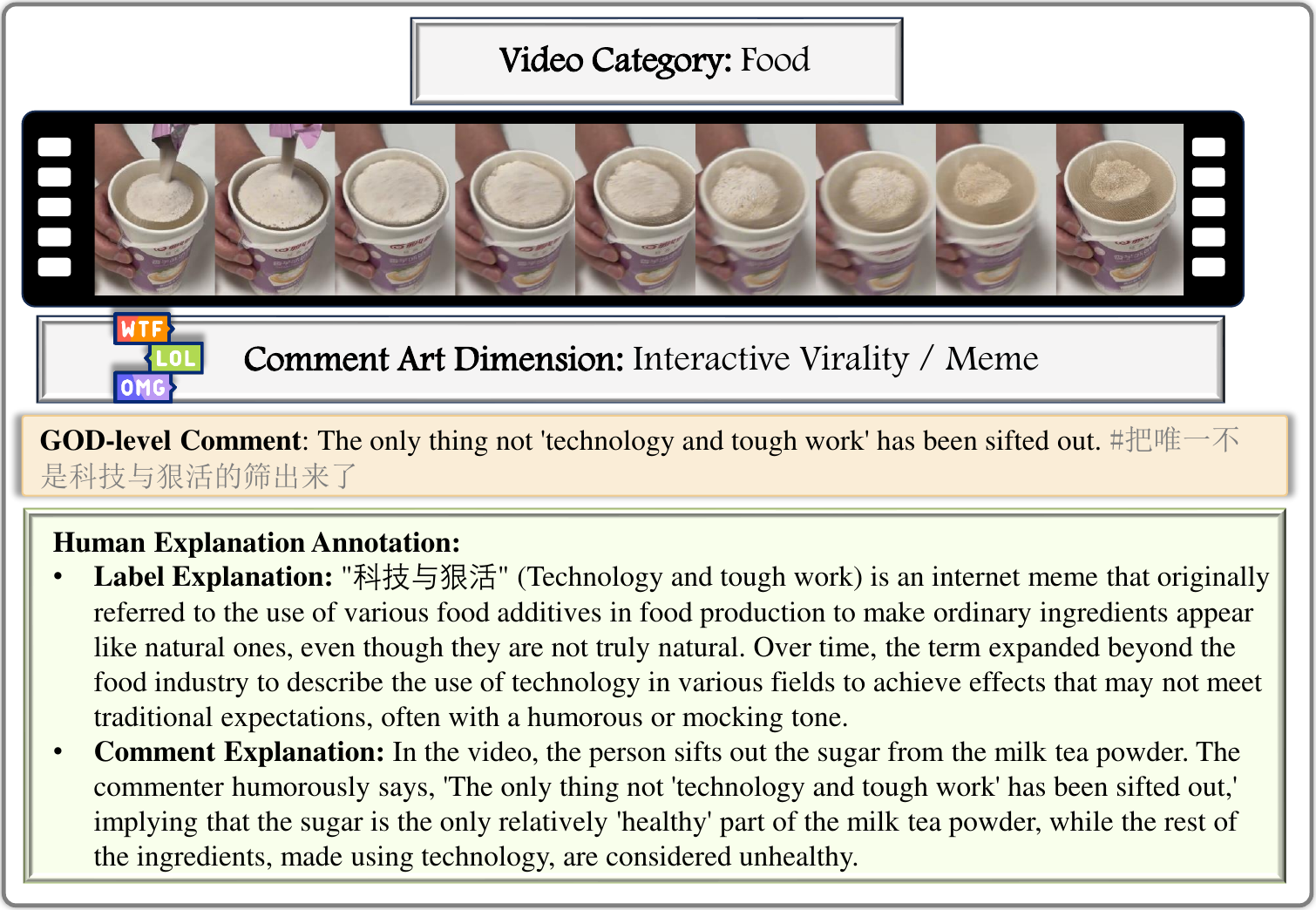}{Interactive Virality / Meme}{  A sample of \textit{Interactive Virality / Meme}.}{fig:case_study_19}

\casestudyfigure{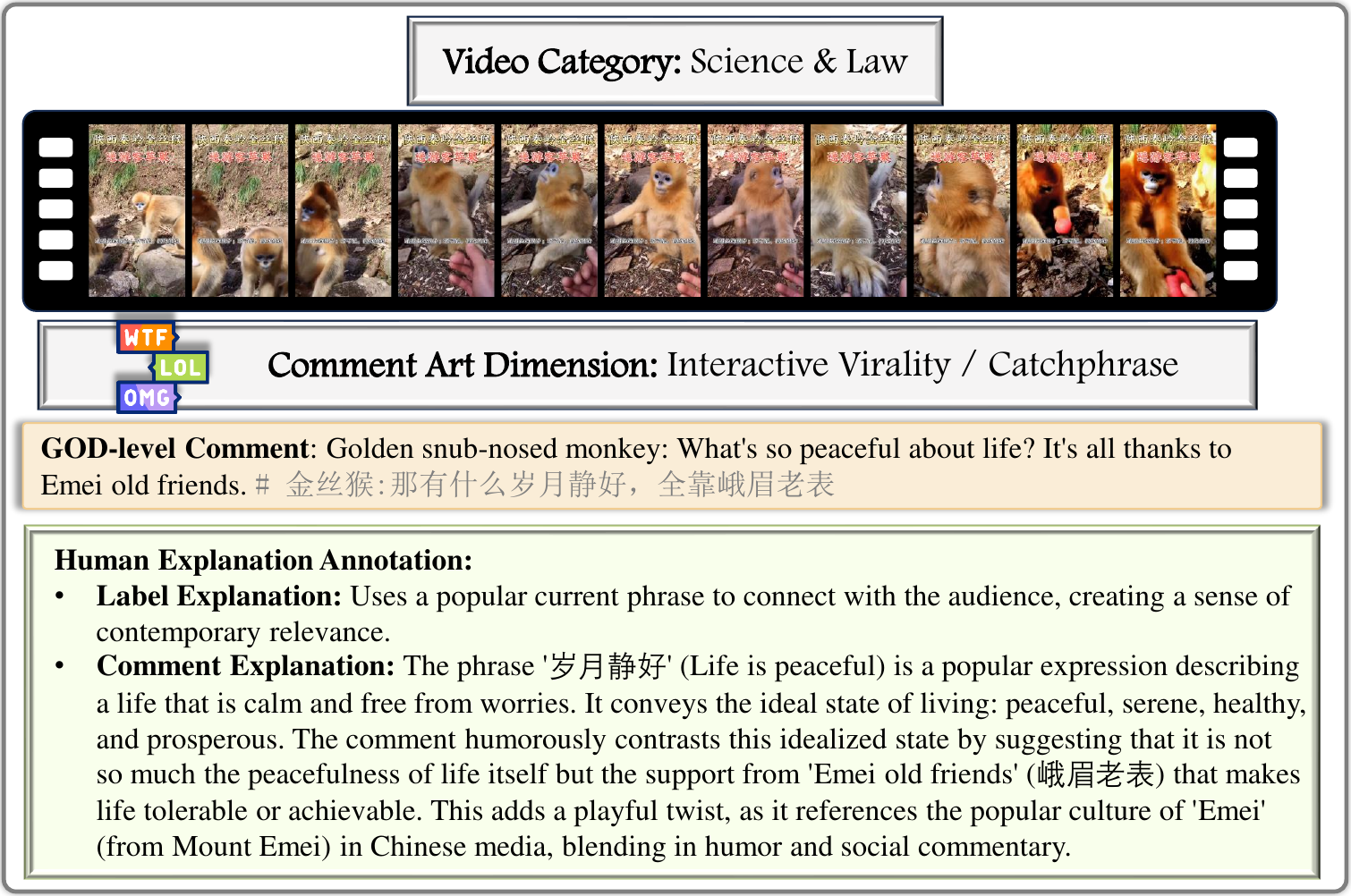}{Interactive Virality / Catchphrase}{  A sample of \textit{Interactive Virality / Catchphrase}.}{fig:case_study_20}

\casestudyfigure{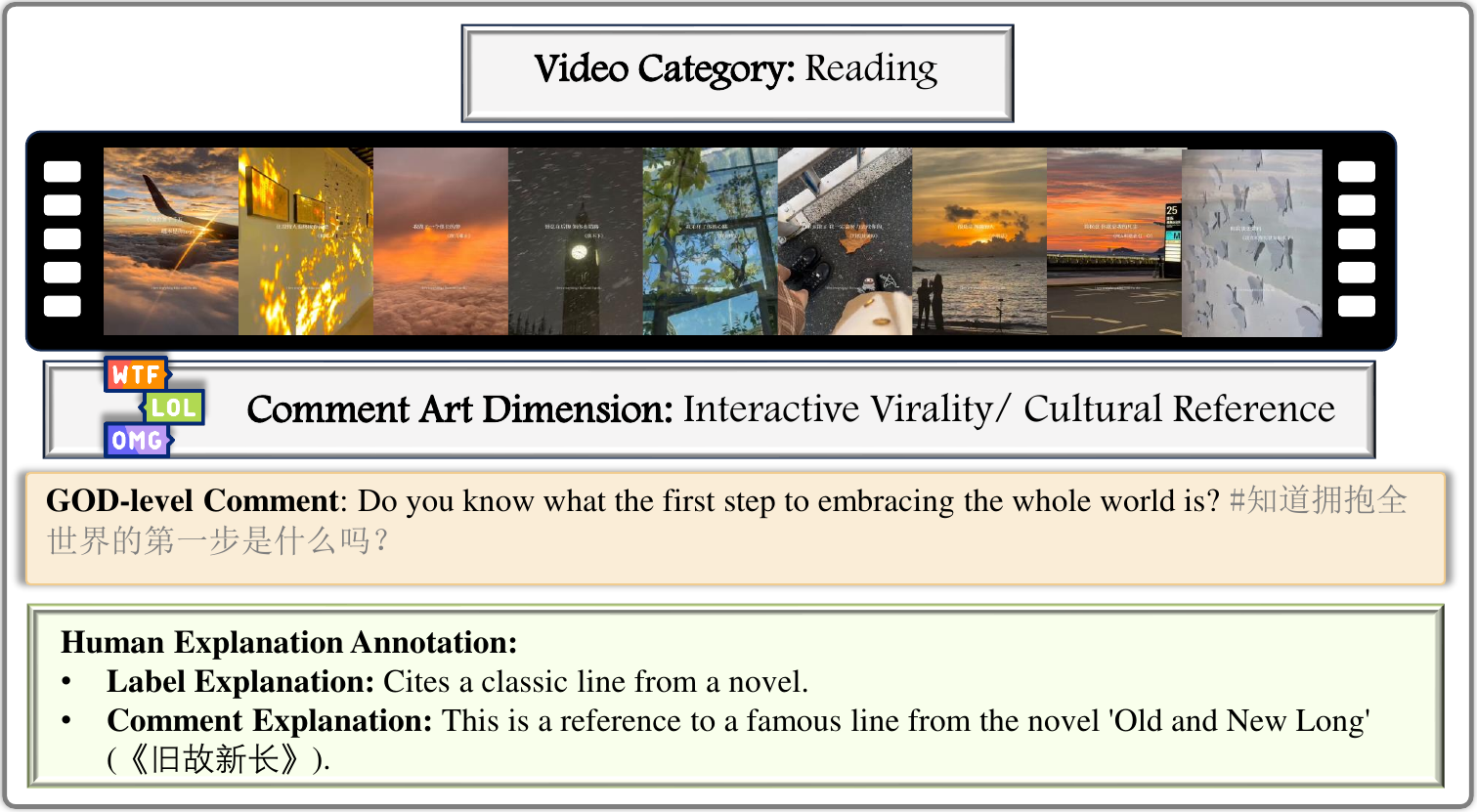}{Interactive Virality/ Cultural Reference}{  A sample of \textit{Interactive Virality/ Cultural Reference}.}{fig:case_study_21}

\casestudyfigure{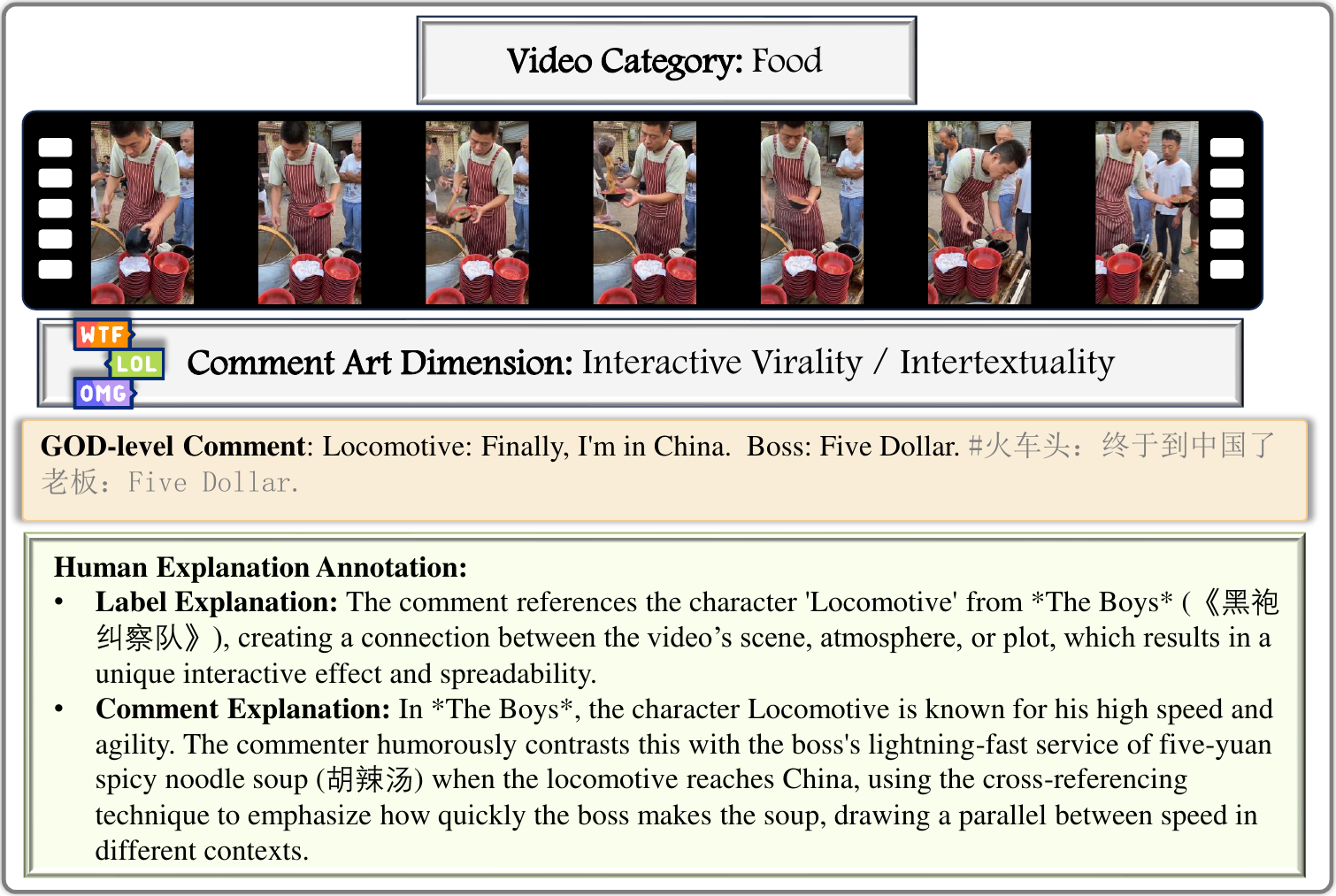}{Interactive Virality / Intertextuality}{  A sample of \textit{Interactive Virality / Intertextuality}.}{fig:case_study_22}

\casestudyfigure{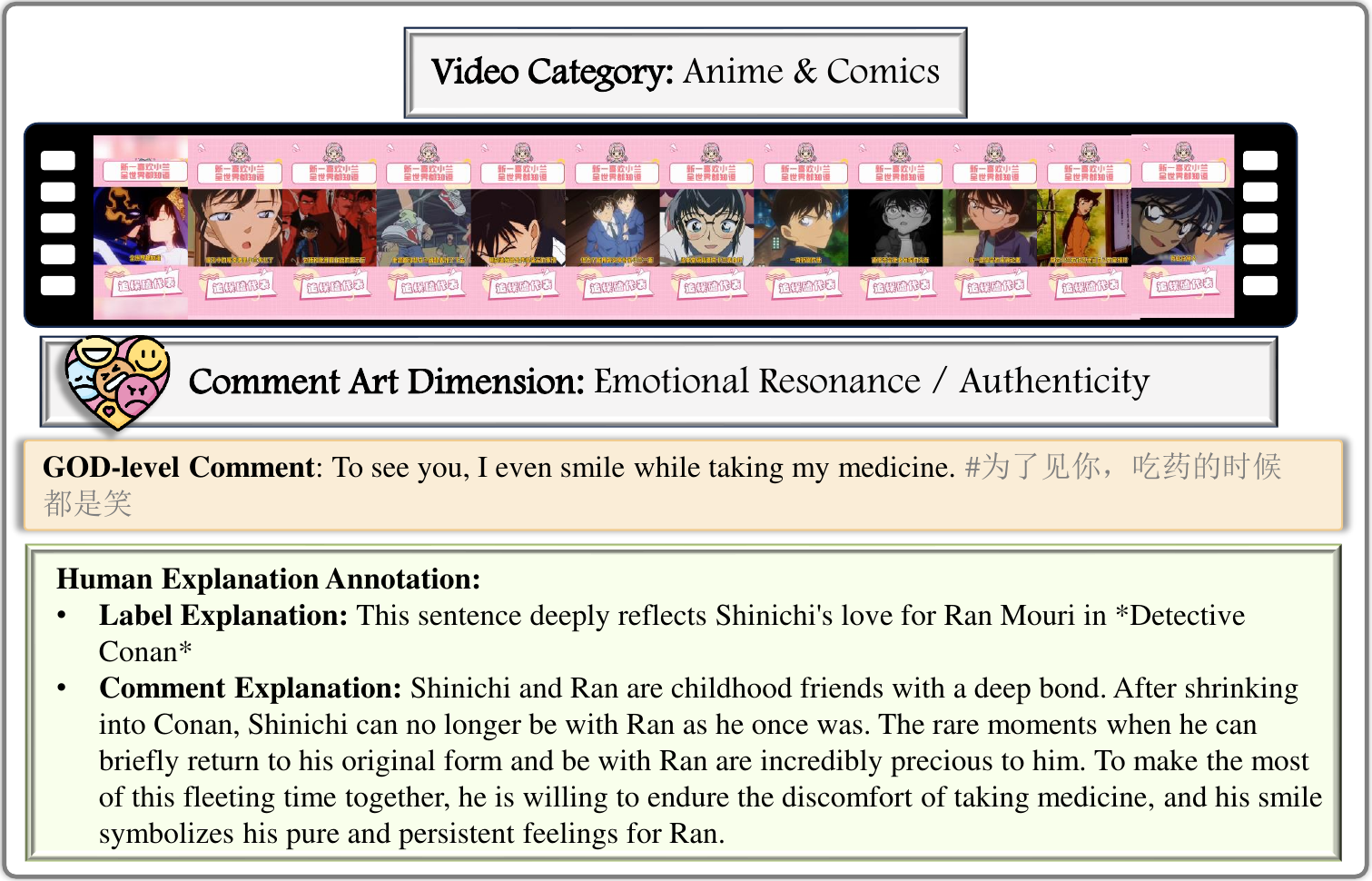}{Emotional Resonance / Authenticity}{  A sample of \textit{Emotional Resonance / Authenticity}.}{fig:case_study_23}

\casestudyfigure{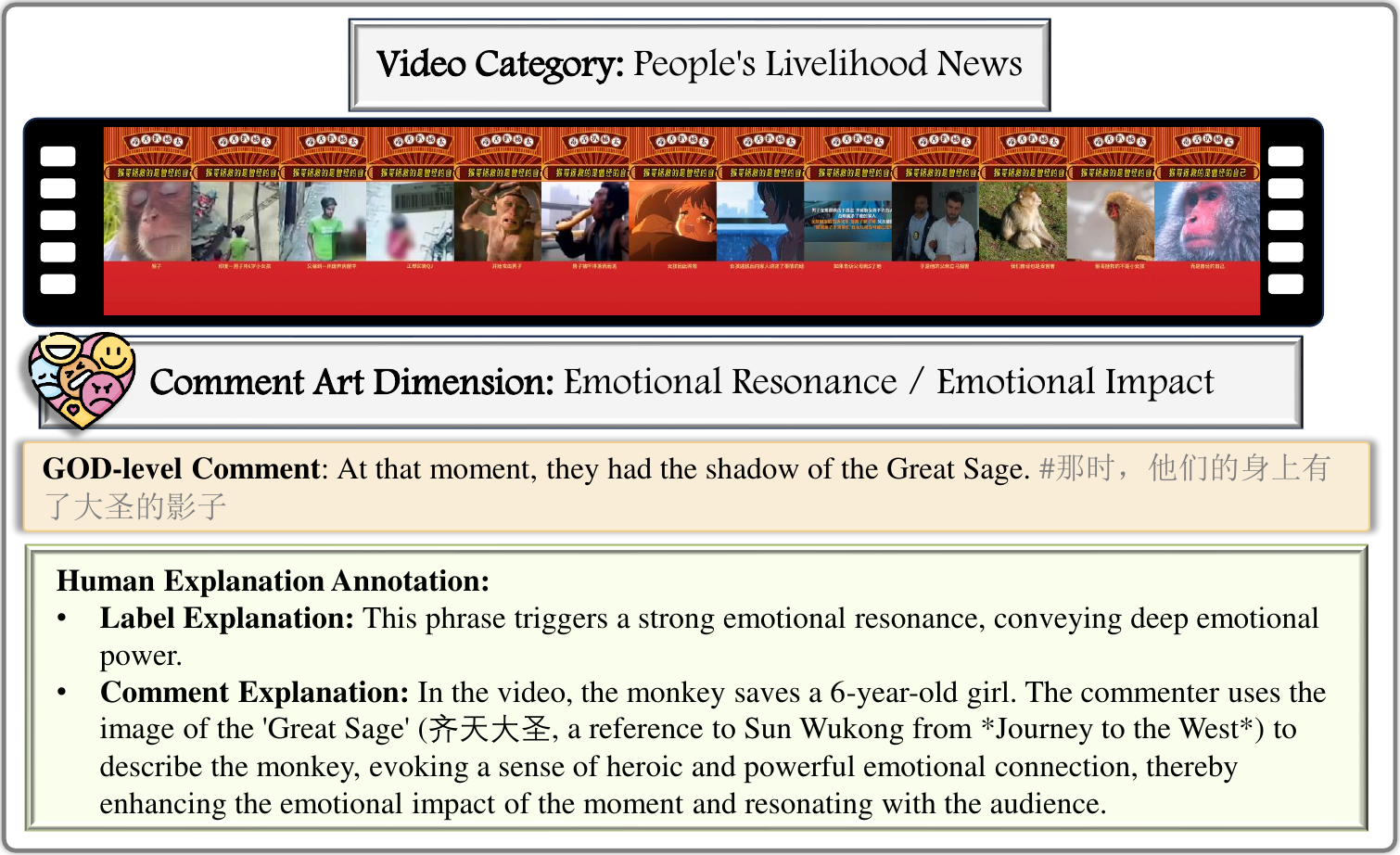}{Emotional Resonance / Emotional Impact}{  A sample of \textit{Emotional Resonance / Emotional Impact}.}{fig:case_study_24}

\casestudyfigure{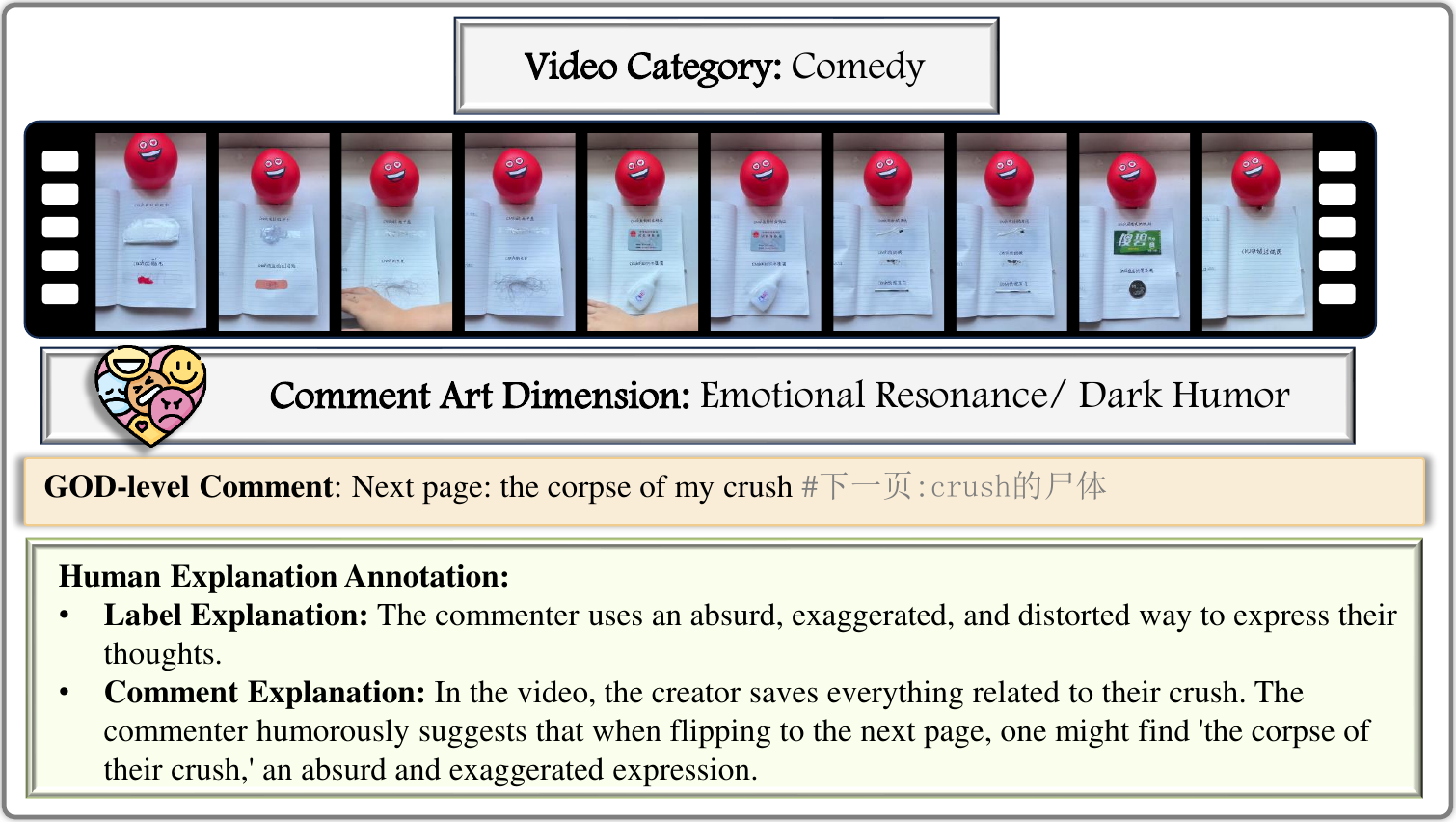}{Emotional Resonance/ Dark Humor}{  A sample of \textit{Emotional Resonance/ Dark Humor}.}{fig:case_study_25}

\casestudyfigure{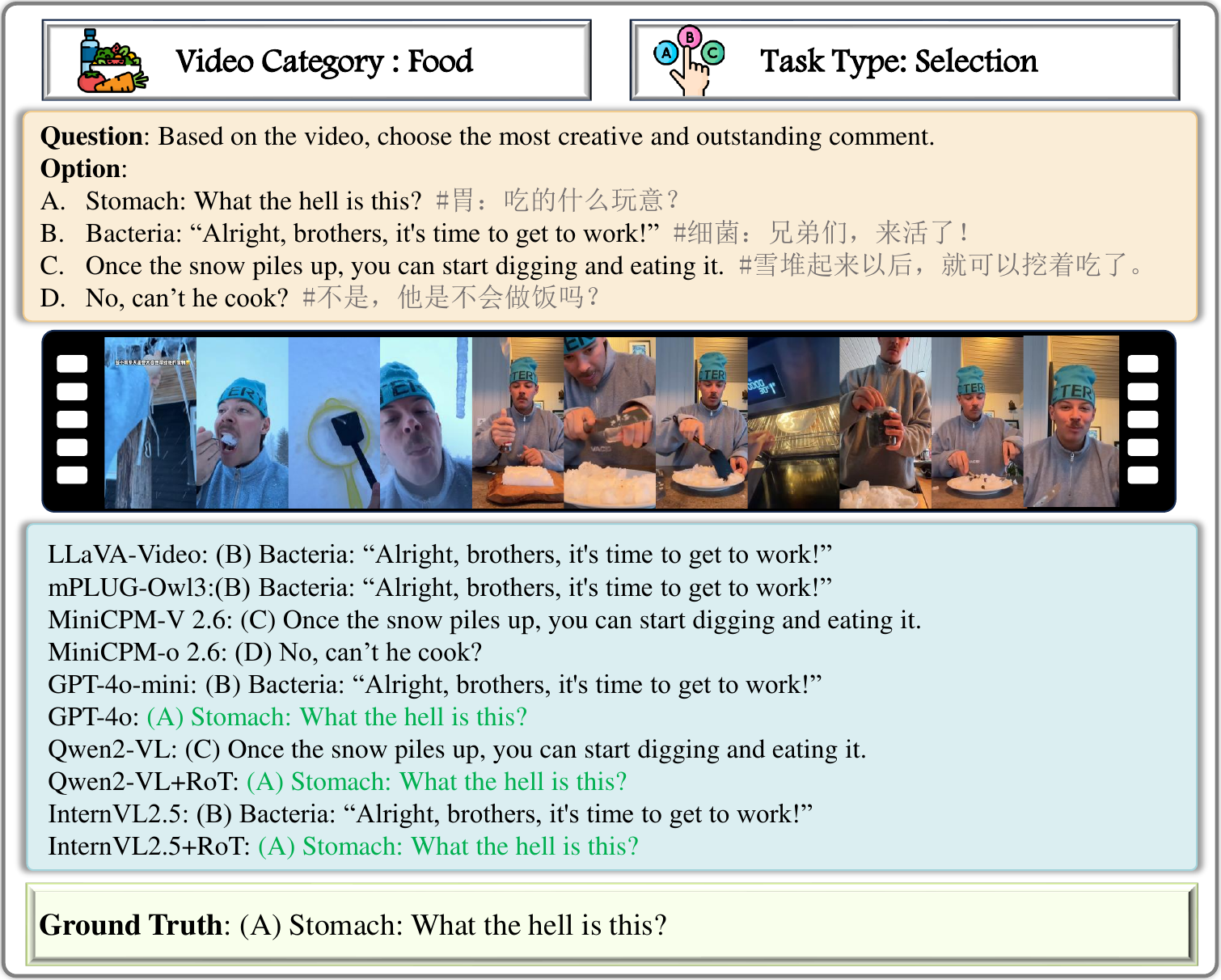}{Selection}{  A sample of \textit{Selection Task}.}{fig:case_study_26}

\casestudyfigure{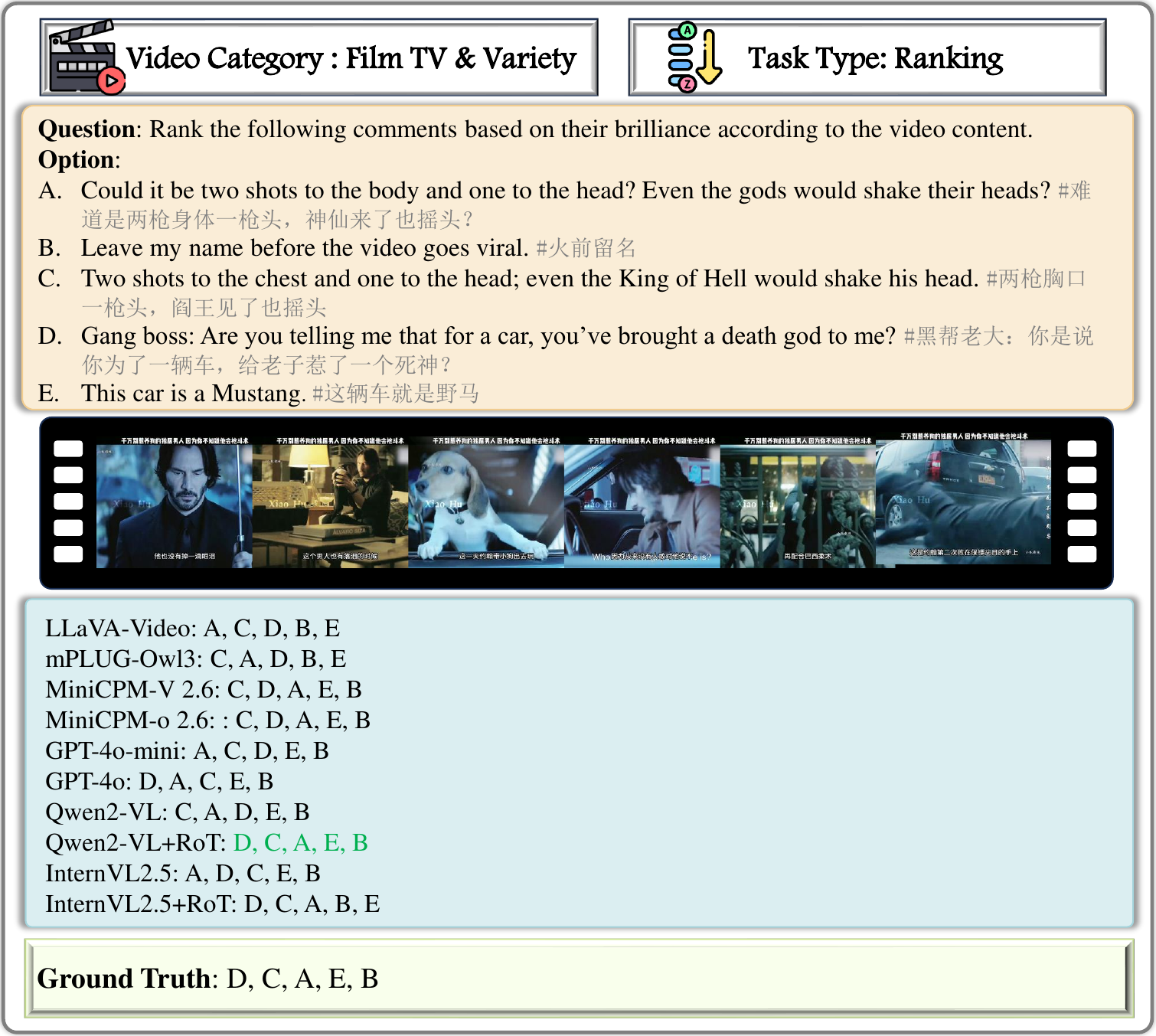}{Ranking}{  A sample of \textit{Ranking Task}.}{fig:case_study_27}

\casestudyfigure{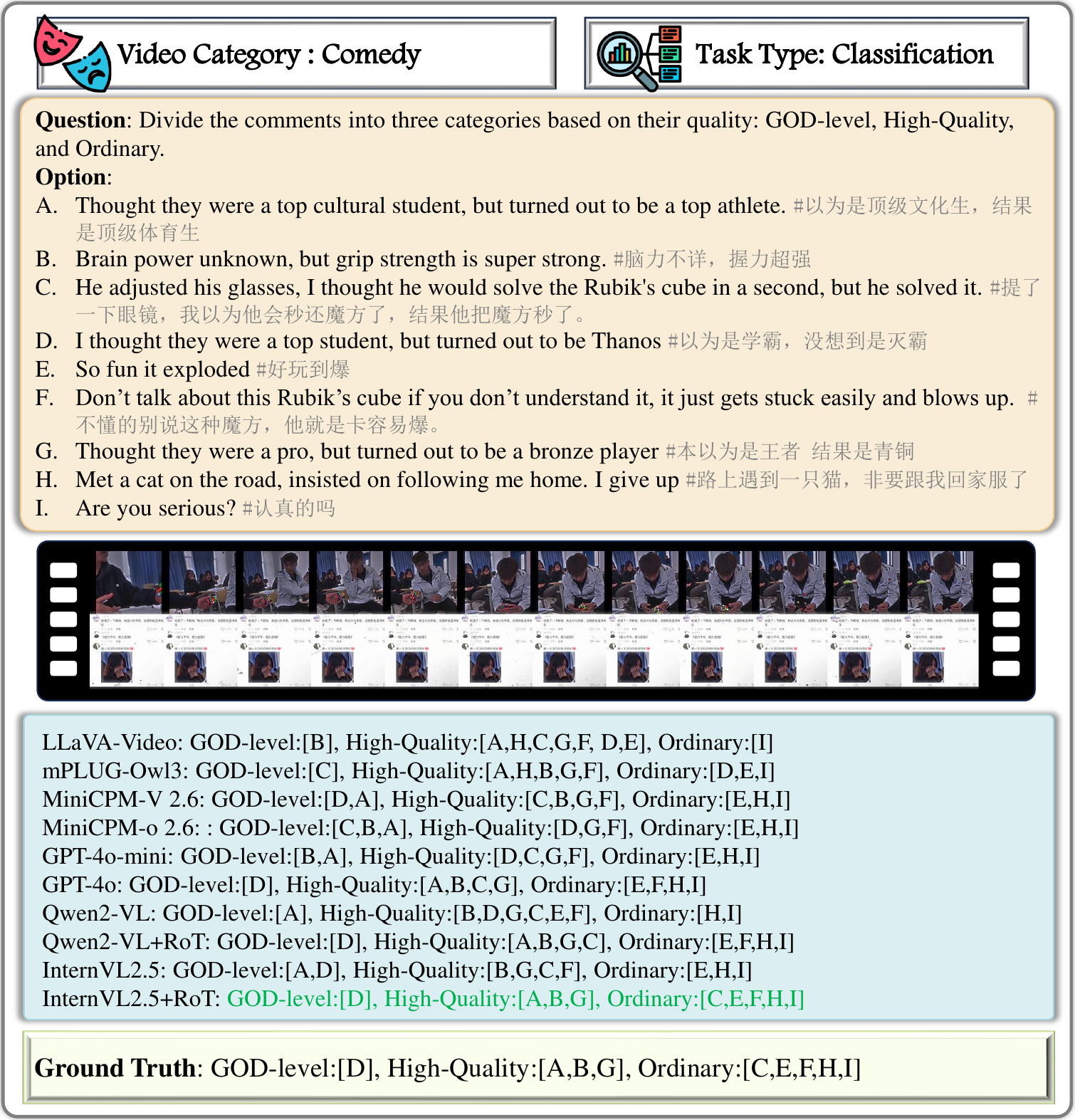}{Classification}{  A sample of \textit{Classification Task}.}{fig:case_study_28}

\casestudyfigure{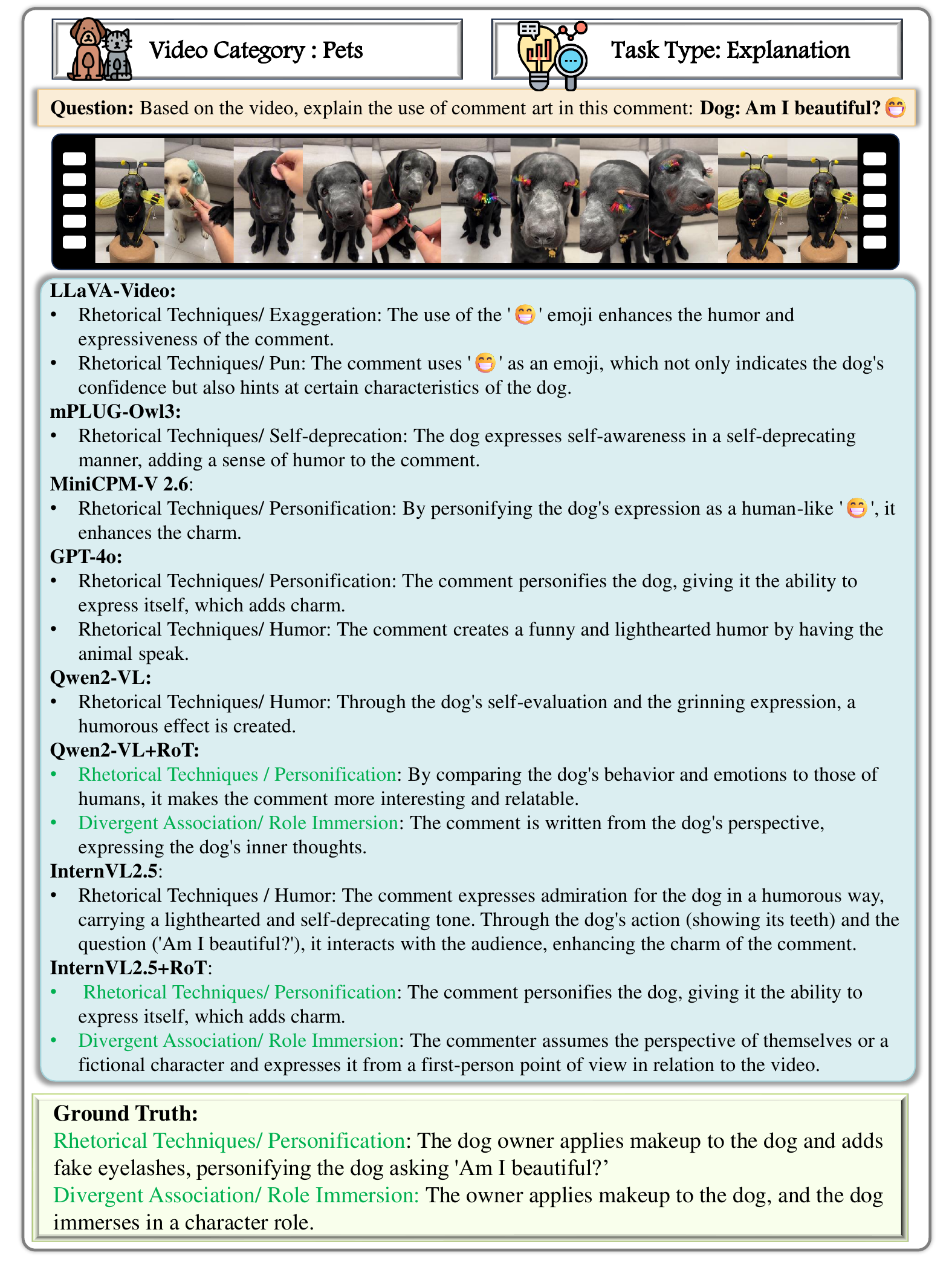}{Explanation}{  A sample of \textit{Explanation Task}.}{fig:case_study_29}

\casestudyfigure{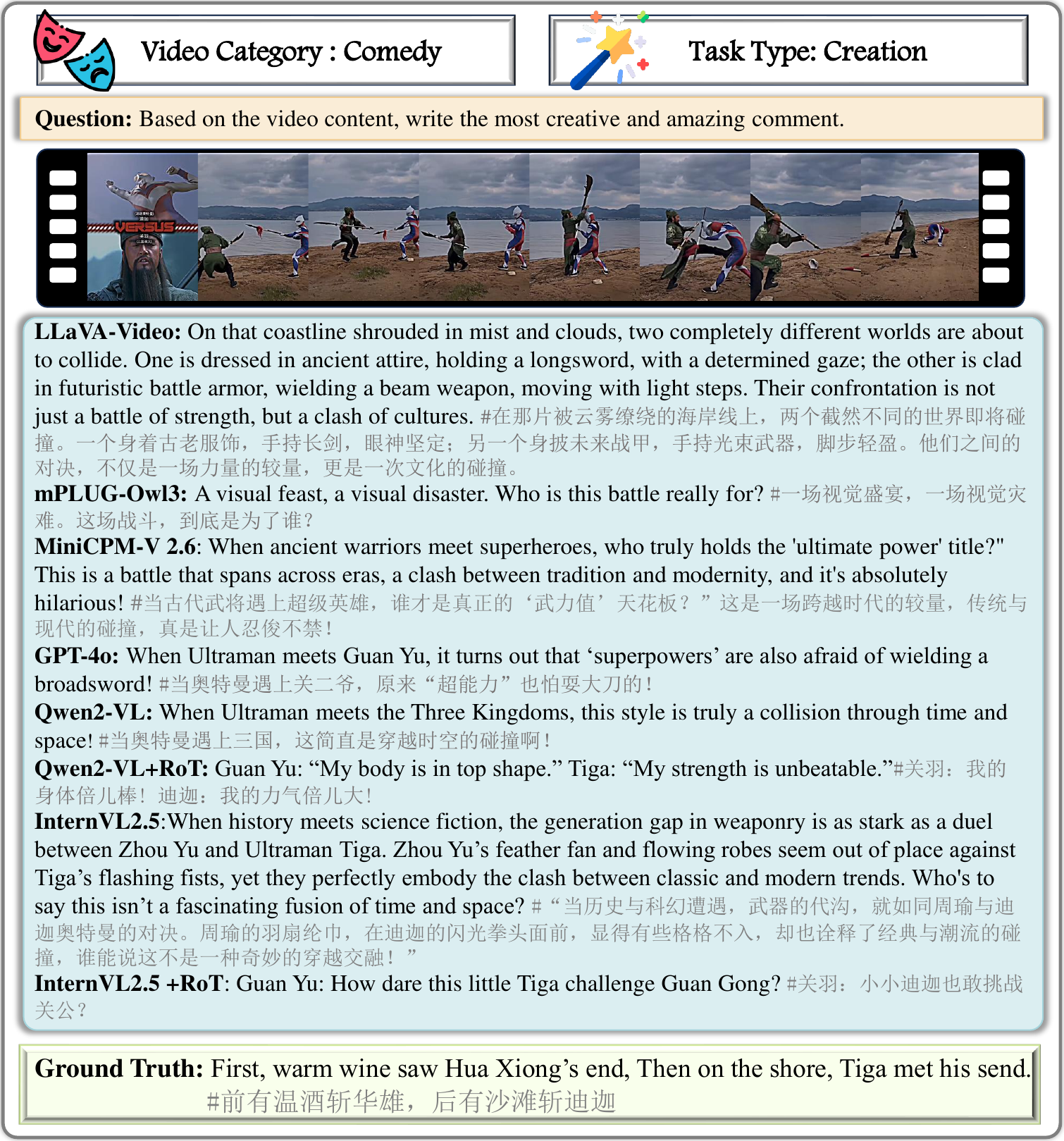}{Creation}{  A sample of \textit{Creation Task}.}{fig:case_study_30}

\casestudyfigure{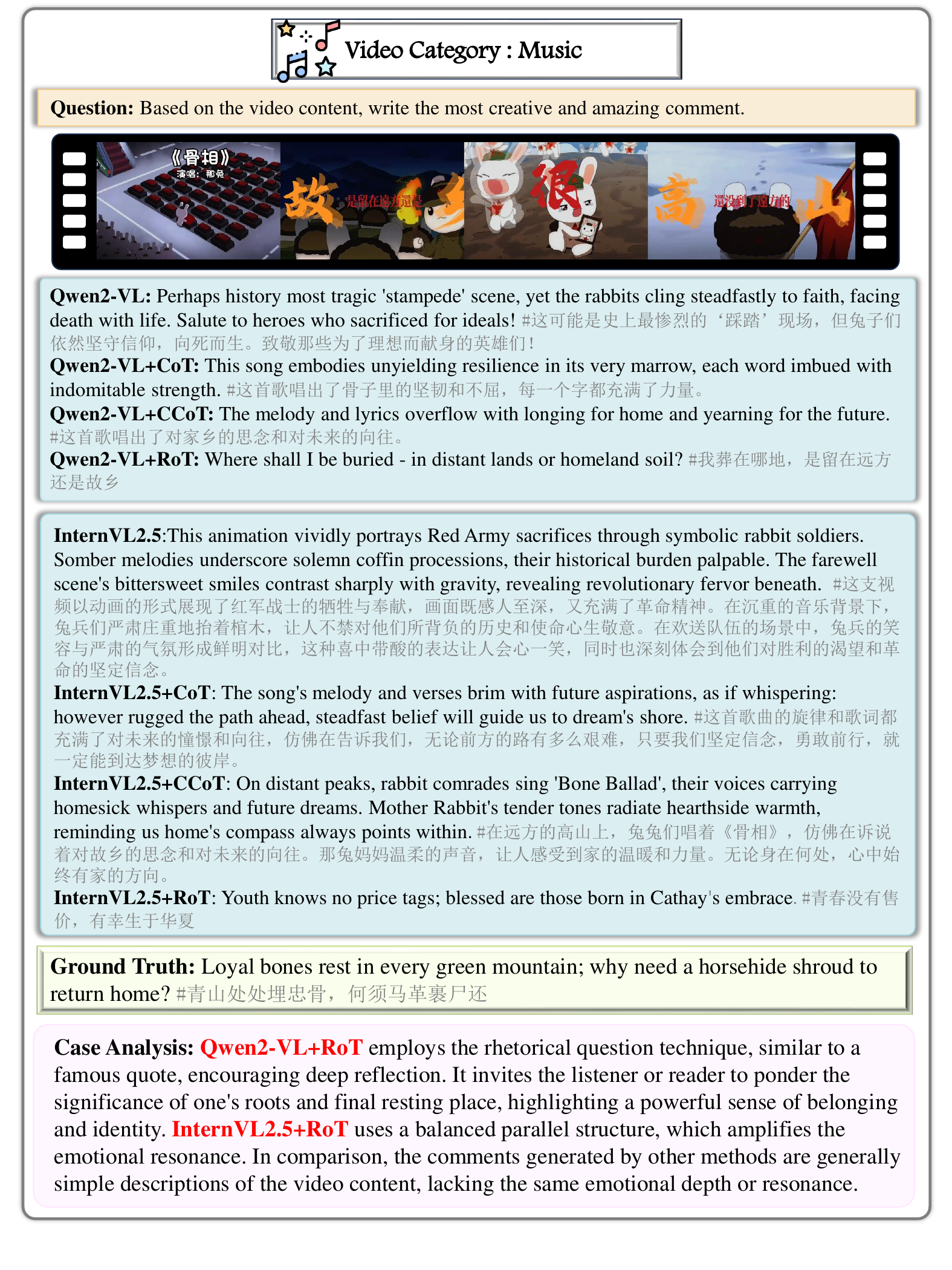}{Case:1}{  A sample of comments generated by our method(\textbf{RoT}) and other methods.}{fig:case_study_31}

\casestudyfigure{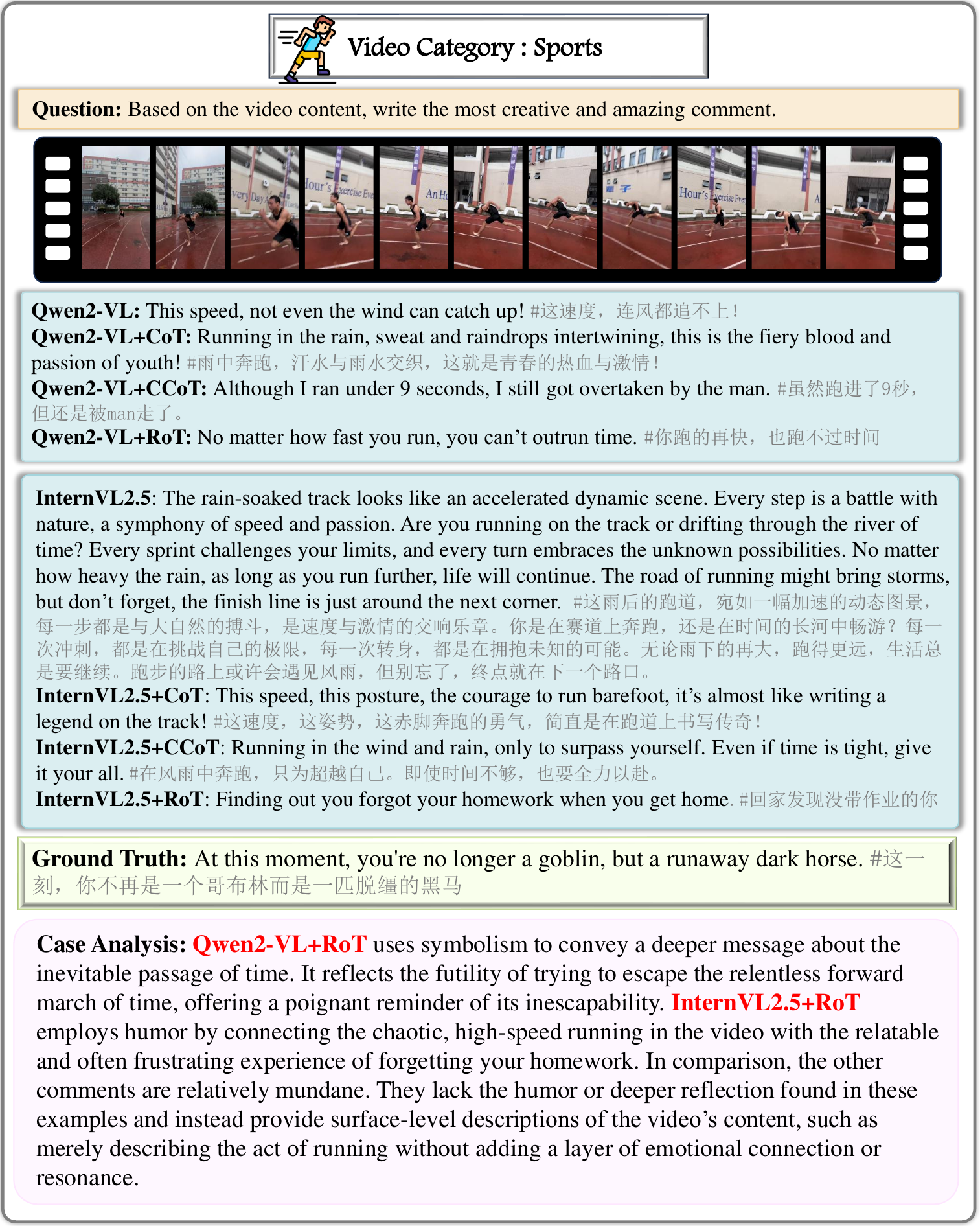}{Case:2}{  A sample of comments generated by our method(\textbf{RoT}) and other methods.}{fig:case_study_32}

\casestudyfigure{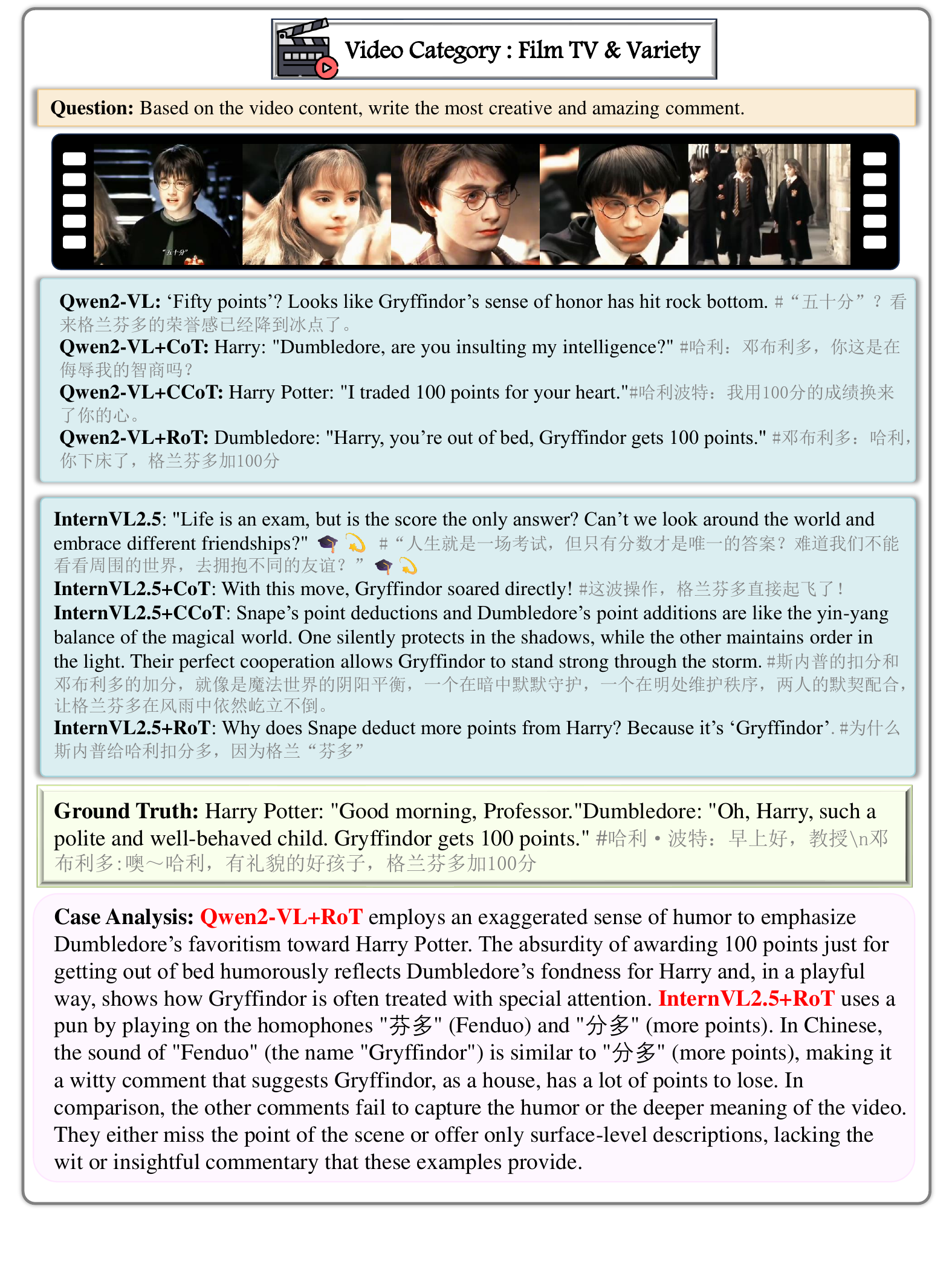}{Case:3}{  A sample of comments generated by our method(\textbf{RoT}) and other methods.}{fig:case_study_33}

\casestudyfigure{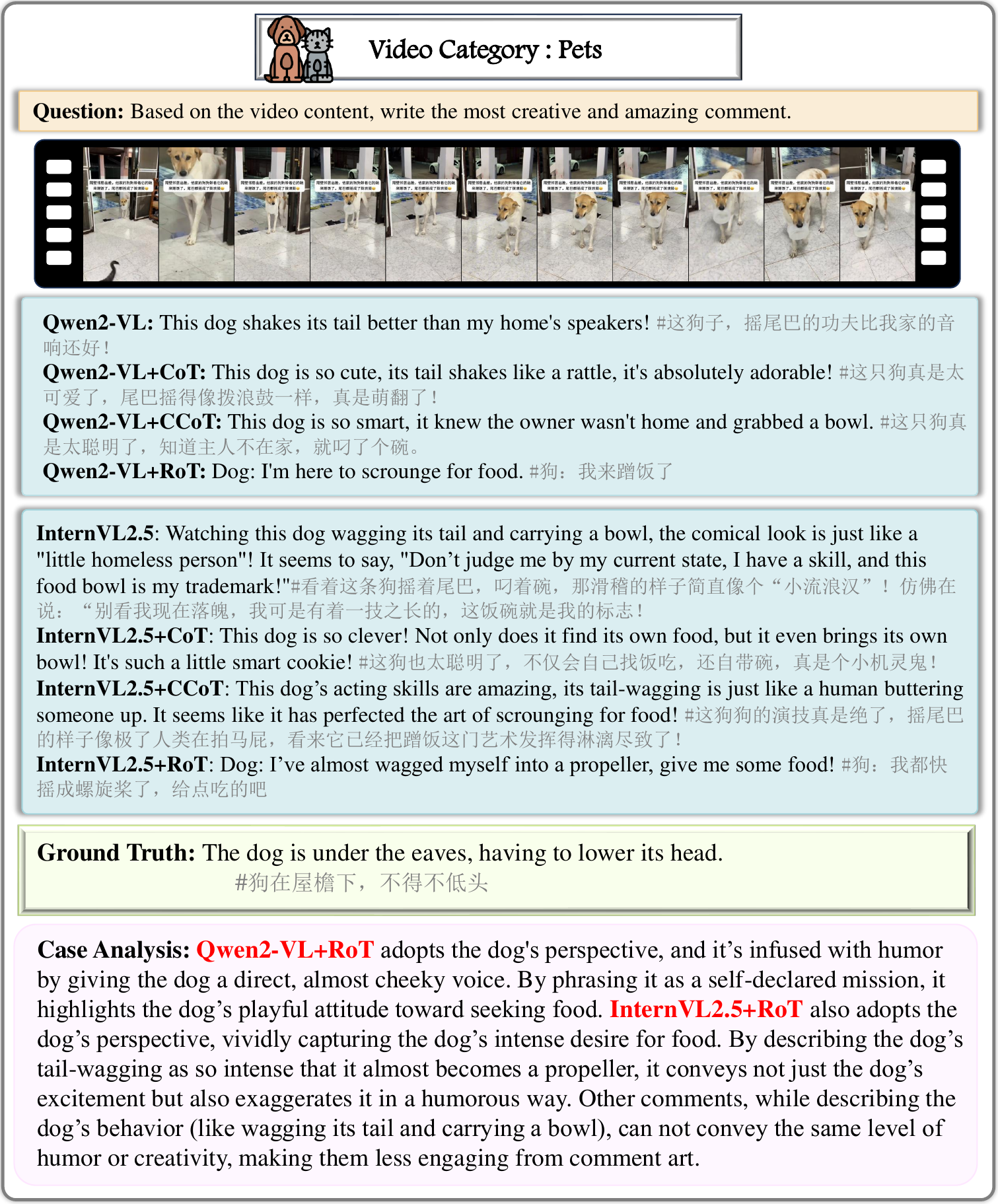}{Case:4}{  A sample of comments generated by our method(\textbf{RoT}) and other methods.}{fig:case_study_34}

\end{document}